\documentclass[journal]{IEEEtran}
\usepackage{makecell}
\usepackage{multirow} 
\usepackage{cite}
\usepackage{fixltx2e}
\usepackage{float}
\usepackage[dvips]{graphicx}
\usepackage[fleqn]{amsmath}
\usepackage{cases}
\usepackage{algorithmic}
\usepackage{algorithm}
\usepackage{mathrsfs}  
\usepackage[caption=false,font=normalsize,labelfont=sf,textfont=sf]{subfig}

\newcommand\NoDo{\renewcommand\algorithmicdo{}}

\usepackage{pdfpages}
\usepackage{bm}
\usepackage{booktabs}  
\usepackage{threeparttable} 
\usepackage{amsfonts,amssymb}
\usepackage{setspace}
\usepackage{color}
\usepackage{stmaryrd}
\usepackage[figuresright]{rotating}

\begin{document}
\title{Tensor Decompositions for Hyperspectral Data Processing in Remote Sensing: A Comprehensive Review}

\author{
Minghua Wang, \IEEEmembership{Member,~IEEE,} 
Danfeng Hong, \IEEEmembership{Senior Member,~IEEE,} 
Zhu Han, \IEEEmembership{Student Member,~IEEE,} 
Jiaxin Li, 
Jing Yao,
Lianru Gao, \IEEEmembership{Senior Member,~IEEE,} 
Bing Zhang, \IEEEmembership{Fellow,~IEEE,} 
Jocelyn Chanussot, \IEEEmembership{Fellow,~IEEE}
\thanks{Manuscript received XX, 2022; revised XX, 2022. This work was supported by the National Natural Science Foundation of China under Grant 62161160336 and Grant 42030111. This work was also supported by the MIAI@Grenoble Alpes (ANR-19-P3IA-0003) and the AXA Research Fund.}
\thanks{M.~Wang D.~Hong, J.~Li, J.~Yao, and L.~Gao are with the Key Laboratory of Computational Optical Imaging Technology, Aerospace Information Research Institute, Chinese Academy of Sciences, Beijing 10094, China (Email: wangmh@aircas.ac.cn; hongdf@aircas.ac.cn; yaojing@aircas.ac.cn; gaolr@aircas.ac.cn ).}
\thanks{J.~Li is with the Key Laboratory of Computational Optical Imaging Technology, Aerospace Information Research Institute, Chinese Academy of Sciences, Beijing 10094, China, and the College of Resources and Environment, University of Chinese Academy of Sciences, Beijing 100049, China (e-mail: lijiaxin203@mails.ucas.ac.cn).}
\thanks{Z.~Han is with the Key Laboratory of Digital Earth Science, Aerospace Information Research Institute, Chinese Academy of Sciences, Beijing 100094, China, and with the International Research Center of Big Data for Sustainable Development Goals, Beijing 100094, China, and also with the College of Resources and Environment, University of Chinese Academy of Sciences, Beijing 100049, China (e-mail: hanzhu19@mails.ucas.ac.cn).}
\thanks{B. Zhang is with the Aerospace Information Research Institute, Chinese Academy of Sciences, Beijing 100094, China, and also with the College of Resources and Environment, University of Chinese Academy of Sciences, Beijing 100049, China (e-mail: zb@radi.ac.cn).}
\thanks{J.~Chanussot is with the Univ. Grenoble Alpes, CNRS, Grenoble INP, GIPSA-Lab, Grenoble, 38000, France, and also with the Aerospace Information Research Institute, Chinese Academy of Sciences, Beijing 10094, China (Email: jocelyn@hi.is).}
}
\markboth{Submission to IEEE GRSM 2022}%
{Shell \MakeLowercase{\textit{et al.}}: Tensor decompositions for hyperspectral data processing: A Comprehensive Review}
\maketitle

\begin{abstract}

Owing to the rapid development of sensor technology, hyperspectral (HS) remote sensing (RS) imaging has provided a significant amount of spatial and spectral information for the observation and analysis of the Earth's surface at a distance of data acquisition devices, such as aircraft, spacecraft, and satellite. The recent advancement and even revolution of the HS RS technique offer opportunities to realize the full potential of various applications, while confronting new challenges for efficiently processing and analyzing the enormous HS acquisition data. Due to the maintenance of the 3-D HS inherent structure, tensor decomposition has aroused widespread concern and research in HS data processing tasks over the past decades. In this article, we aim at presenting a comprehensive overview of tensor decomposition, specifically contextualizing the five broad topics in HS data processing, and they are HS restoration, compressed sensing, anomaly detection, super-resolution, and spectral unmixing. For each topic, we elaborate on the remarkable achievements of tensor decomposition models for HS RS with a pivotal description of the existing methodologies and a representative exhibition on the experimental results. As a result, the remaining challenges of the follow-up research directions are outlined and discussed from the perspective of the real HS RS practices and tensor decomposition merged with advanced priors and even with deep neural networks.  
This article summarizes different tensor decomposition-based HS data processing methods and categorizes them into different classes from simple adoptions to complex combinations with other priors for the algorithm beginners. We also expect this survey can provide new investigations and development trends for the experienced researchers who understand tensor decomposition and HS RS to some extent.

\end{abstract}

\IEEEpeerreviewmaketitle

\section{Introduction}
\label{sect:Intro}


HS RS imaging has gradually become one of the most vital achievements in the field of RS since the 1980s \cite{c173}. Varied from an initial single-band panchromatic image, a three-band color RGB image, and a several-band multispectral (MS) image, an HS image contains hundreds of narrow and continuous spectral bands, which is promoted by the development of spectral imaging equipment and the improvement of spectral resolutions. The broader portion of the HS spectrum can scan from the ultraviolet, extend into the visible spectrum, and eventually reach in the near-infrared or short-wave infrared \cite{9395693}. Each pixel of HS images corresponds to a spectral signature and reflects the electromagnetic properties of the observed object. This enables the identification and discrimination of underlying objects, especially some that have a similar property in single-band or several-band RS images (such as panchromatic, RGB, MS) in a more accurate manner. As a result, the wealthy spatial and spectral information of HS images has extremely improved the perceptual ability of Earth observation, which makes the HS RS technique play a crucial role in the fields like precision agriculture (e.g., monitoring the growth and health of crops), space exploration (e.g., searching for signs of life on other planets), pollution monitoring (e.g., detection of the ocean oil spill), and military applications (e.g., identification of military targets) \cite{c1,c172,9174822}. 

Over the past decade, massive efforts have been made to process and analyze HS RS data after the data acquisition. Initial HS data processing considers either a gray-level image for each band or the spectral signature of each pixel. From one side, each HS spectral band is regarded as a gray-level image, and the traditional 2-D image processing algorithms are directly introduced band by band \cite{c175,c176}. From another side, the spectral signatures that have similar visible properties (e.g., color, texture) can be used to identify the materials \cite{c174}. Furthermore, extensive low-rank (LR) matrix-based methods are employed to explore the high correlation of spectral channels with the assumption that the unfolding HS matrix has a low rank \cite{c20,c177,c178}. Given an HS image of size h×v×z, the recovery of an unfolding HS matrix (hv×z) usually requires the singular value decomposition (SVD), which leads to the computational cost of $O(h^2 v^2 z+ z^3)$ \cite{c30,c31,c32}. In some typical tensor decomposition-based methods, the complexity of the tensor singular value decomposition (t-SVD) is about $O(hvzlogz+ hv^2z)$ \cite{c28,c34,c61}. Compared to matrix forms, tensor decompositions achieve excellent performances with a tolerable increment of computational complexity. However, these traditional LR models reshape each spectral band as a vector, leading to the destruction of the inherent spatial-spectral completeness of HS images. Correct interpretations of HS images and the appropriate choice of the intelligent models should be determined to reduce the gap between HS tasks and the advanced data processing technique. Both 2-D spatial information and 1-D spectral information are considered when an HS image is modeled as a three-order tensor.


\begin{figure*}[htp!]
	\begin{center}
    \includegraphics[width = 1\textwidth]{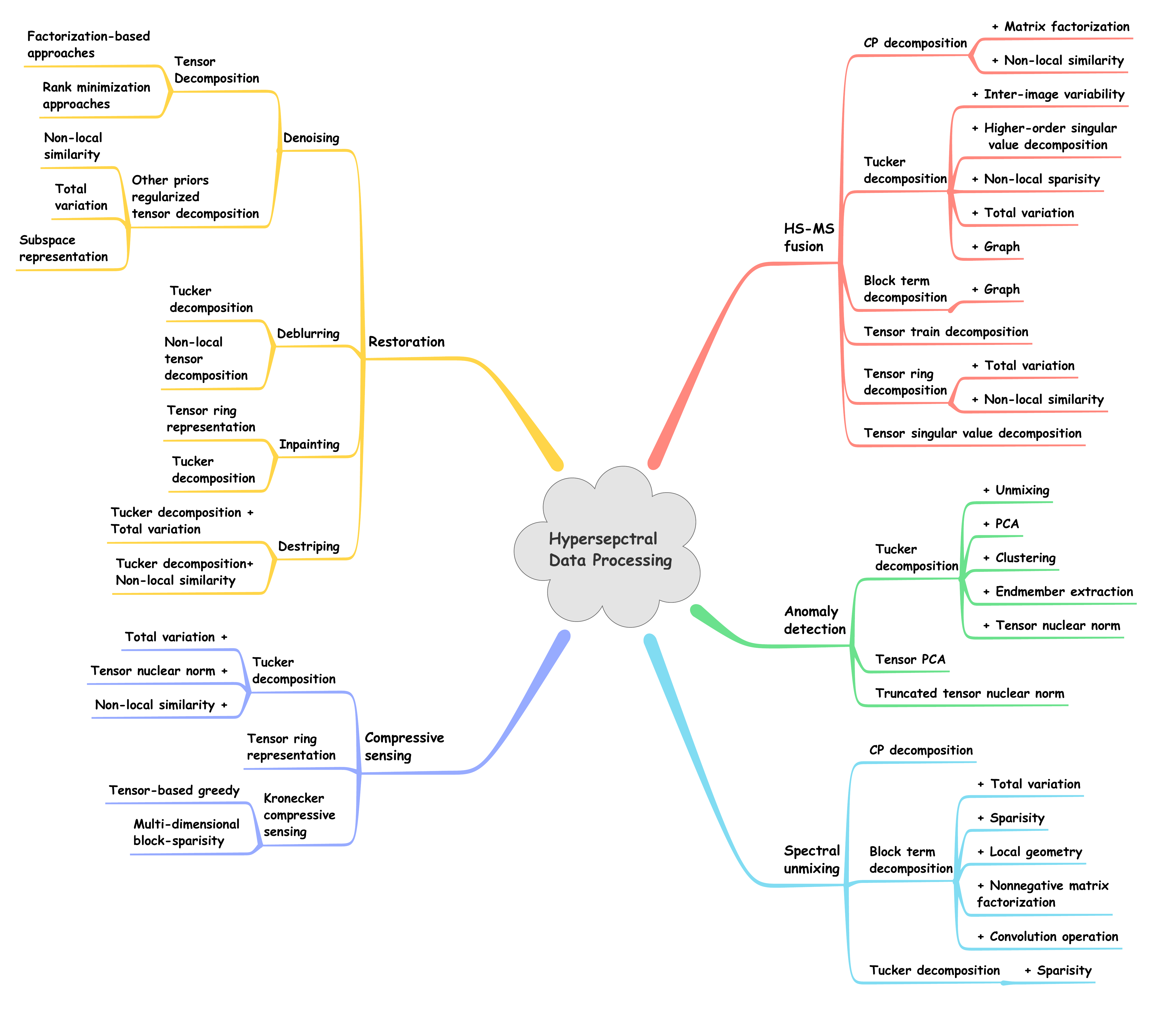}
	\end{center}
	\caption[houston]{A taxonomy of main tensor decomposition-based methods for HS data processing. }
	\label{fig:TLreference}
\end{figure*}

Tensor decomposition, which originates from Hitchcock's works in 1927 \cite{c179}, touches upon numerous disciplines, but it has recently become prosperous in the fields of signal processing, machine learning, data mining and fusion over the last ten years \cite{c181,c182,c183}. The early overviews focus on two common decomposition ways: Tucker decomposition and CANDECOMP/PARAFAC (CP) decomposition. In 2008, these two decompositions were first introduced into HS restoration tasks to remove the Gaussian noise \cite{c25,c26}. The tensor decomposition-based mathematical models avoid converting the original dimensions, and also to some degree, enhance the interpretability and completeness for problem modeling. Different types of prior knowledge (e.g, non-local similarity in the spatial domain, spatial and spectral smoothness) in HS RS are considered and incorporated into the tensor decomposition frameworks. However, on the one hand, additional tensor decomposition methods have been proposed recently, such as block term (BT) decomposition, t-SVD \cite{c184}, tensor train (TT) decomposition \cite{c185}, and tensor ring (TR) decomposition \cite{c126}. On the other hand, as a versatile tool, tensor decomposition related to HS image processing has not been reviewed until. In this article, we mainly present a systematic overview from the perspective of the state-of-the-art tensor decomposition techniques for HS data processing in terms of the five burgeoning topics previously mentioned, as presented in Fig. \ref{fig:TLreference}. 


\begin{figure}[htp!]
	\begin{center}
    \includegraphics[width = 0.45\textwidth]{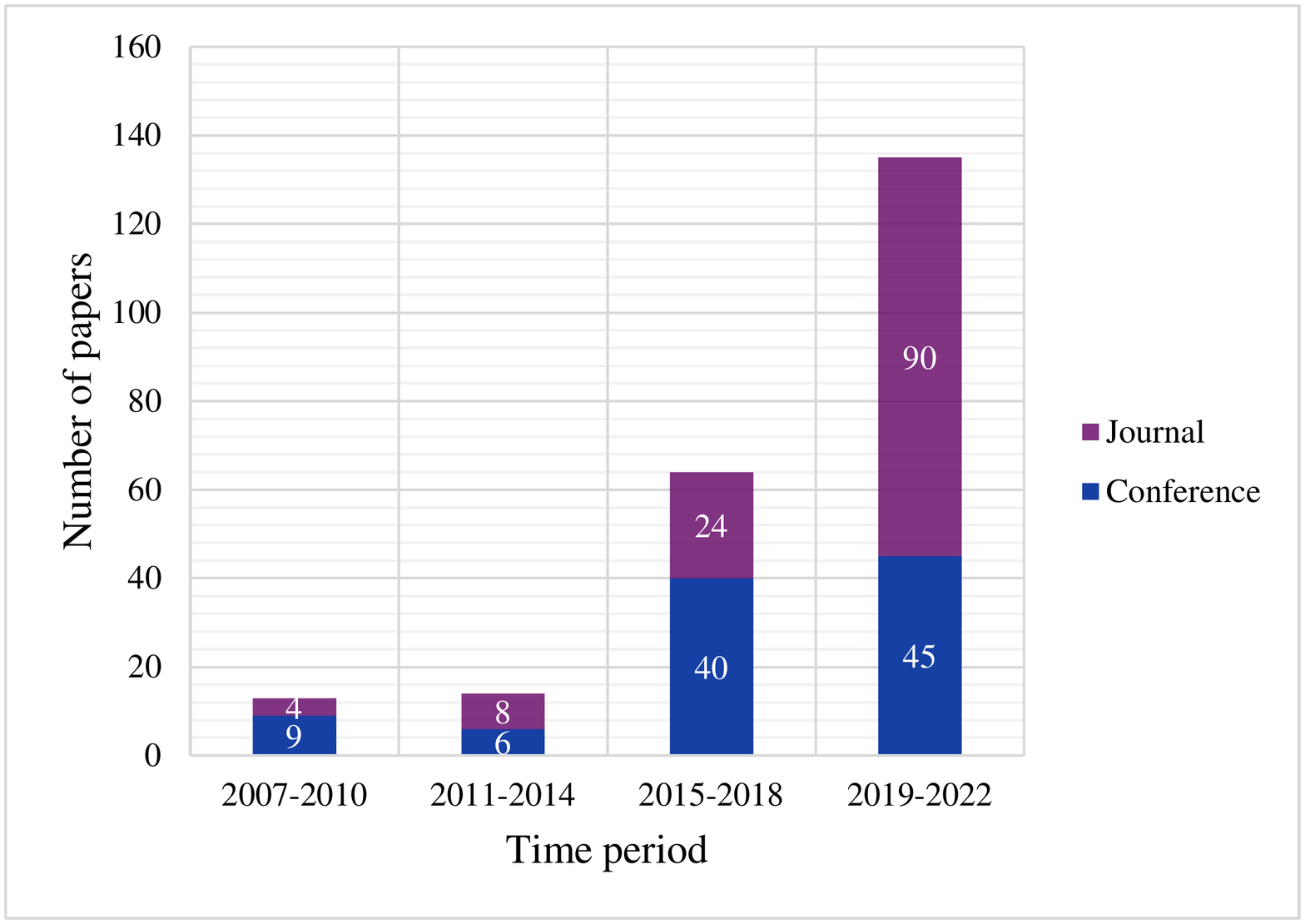}
	\end{center}
	\caption[houston]{The number of journal and conference papers that published in IEEE Xplore on the subject of "hyperspectral" and "tensor decomposition" within different time periods. }
	\label{fig:Visio-papernum}
\end{figure}
Fig. \ref{fig:Visio-papernum} displays the dynamics of tensor decompositions used for HS data processing in the HS community. The listed numbers contain both scientific journal and conference papers published in IEEE Xplore, which regards "hyperspectral" and "tensor decomposition" as the main keywords in abstracts. To highlight the increasing trend of number of publications, time period has been divided into four equal time slots (i.e., 2007-2010, 2011-2014, 2015-2018, 2019-2022(05 January)).
In this article, we mainly present a systematic overview from the perspective of the state-of-the-art tensor decomposition techniques for HS data processing in terms of the five burgeoning topics previously mentioned. 

\noindent
\hangafter=1
\setlength{\hangindent}{2em}
(1) To the best of our knowledge, this is the first time to provide a comprehensive survey of the state-of-the-art tensor decomposition techniques for processing and analyzing HS RS images. More than 100 publications in this field are reviewed and discussed, most of which were published during the last five years.

\noindent
\hangafter=1
\setlength{\hangindent}{2em}
(2) For each HS topic, major representative works are scrupulously presented in terms of the specific categories of tensor decomposition. We introduce and discuss the pure tensor decomposition-based methods and their variants with other HS priors in sequence. The experimental examples are performed for validating and evaluating theoretical methods, followed by a discussion of remaining challenges and further research directions.

\noindent
\hangafter=1
\setlength{\hangindent}{2em}
(3) This article makes a connection between tensor decomposition modeling and HS prior information. Tab. \ref{tab:tab1} summarizes with the publication years, brief description, and prior information. Either beginners or experiencers are expected to obtain certain harvest pertinent to the tensor decomposition-based frameworks for HS RS. The available codes are also displayed in Tab. \ref{tab:tab1} for the sake of repeatability and further studies.

\begin{table*}[!htbp]
\centering
\caption{Tensor decomposition-based approaches for HS RS.}
\resizebox{\textwidth}{!}{
\begin{tabular}{c|c|c|c|c|c}
\cline{1-6}
Category &Years & Methods & Brief Description & Prior Information & Code Links\\
\hline
Rstoration & 2008 & LRTA \cite{c25} & Tucker decomposition & Spectral correlation & \\
Denoising& 2008 & PARAFAC \cite{c26} & CP decomposition & Spectral correlation & \\
& 2013 & R1TD \cite{c59} & Rank-1 Tensor decomposition & Spectral correlation & \\
&2017& LRTR \cite{c28} & TNN & Spectral correlation & \\
&2019 & NTRM \cite{xue2019nonconvex} & Logarithmic TTN & Spectral correlation &\\
&2020 & 3DTNN / 3DLogTNN \cite{c61}& Three-directional TNN / Log-based TNN &Spectral correlation & https://yubangzheng.github.io/homepage/ \\
&2014& TDL & Tucker decomposition with dictionary learning & Spectral correlation + Non-local similarity & http://www.cs.cmu.edu/~deyum/ \\
&2018 & NSNTD \cite{c71} & Non-local similarity based nonnegative tucker decomposition &  Spectral correlation + Non-local similarity & \\
&2019 & GNWTTN \cite{c67} & Global and non-local weighted TTN &Spectral correlation + Non-local similarity & \\
&2015& NLTA-LSM \cite{c65} & Tensor decomposition with laplacian scale mixture & Spectral correlation + Non-local similarity & \\
& 2016  &ITS\cite{c63} & CP + Tucker decomposition &Spectral correlation + Non-local similarity & https://gr.xjtu.edu.cn/web/dymeng/3 \\
&2019 & NLR-CPTD \cite{xue2019nonlocal} & CP + Tucker decomposition &Spectral correlation + Non-local similarity  &\\
&2017& LLRT \cite{chang2017hyper} & hyper-Laplacian prior + Unidirectional LR tensor &  Non-local similarity + Spectral smoothness &https://owuchangyuo.github.io/publications/LLRT \\
&2019 & NGmeet \cite{he2019non}& Spectral subspace-based unidirectional LR tensor & Spectral correlation + Non-local similarity &https://prowdiy.github.io/weihe.github.io/publication.html\\
&2020& WLRTR \cite{c116} & Weighted Tucker decomposition & Spectral correlation + Non-local similarity & https://owuchangyuo.github.io/publications/WLRTR \\
& 2020 & NLTR \cite{c64} & Nonlocal TR decomposition & Spectral correlation + Non-local similarity  &https://chenyong1993.github.io/yongchen.github.io/\\
& 2018 & TLR-TV \cite{c77} & TNN + 2DTV / 3DTV & Spectral correlation + Spatial $\&$ Spectral smoothness & \\
& 2018 & SSTV-LRTF \cite{c34} & TNN + SSTV & Spectral correlation + Spatial-Spectral smoothness & \\
& 2021 & MLR-SSTV \cite{8920965} & Multi-directional weighted TNN + SSTV  & Spectral correlation + Spatial-Spectral smoothness &\\
&2018& LRTDTV \cite{c36} & 3DwTV + Tucker decomposition & Spectral correlation + Spatial-Spectral smoothness & https://github.com/zhaoxile\\
&2021 &TLR-$L_{1\--2}{\rm SSTV}$ \cite{c84} & $L_{1\--2}{\rm SSTV}$ + Local-patch TNN&Spectral correlation + Spatial-Spectral smoothness& \\
&2019 &LRTDGS \cite{c78}& Weighted group sparsity-regularized TV + Tucker decomposition& Spectral correlation + Spatial-Spectral smoothness &https://chenyong1993.github.io/yongchen.github.io/ \\
&2019 & LRTF$L_0$ \cite{c81}& $l_0$ gradient constraint + LR BT decomposition & Spectral correlation + Spatial-Spectral smoothness & http://www.xiongfuli.com/cv/ \\
&2021&TLR-${l_0}\text{TV}$ \cite{c82}&${l_0}\text{TV}$ + LR tensor & Spectral correlation + Spatial-Spectral smoothness &https://github.com/minghuawang666/TLR-L0TV\\
& 2019  & SNLRSF \cite{c72}&  Subspace-based non-local LR and sparse factorization & Spectral correlation + Non-local tensor subspace  & https://github.com/AlgnersYJW/\\
&2020 & LRTF-DFR \cite{c86}& double-factor-regularized LR tensor factorization & Subspace spectral correlation + spatial \& spectral constraints & https://yubangzheng.github.io/homepage/\\
&2021 &DNTSLR \cite{c88}&  Difference continuity + Non-local tensor subspace  &  Spectral correlation + Non-local tensor subspace & \\
\cline{2-6}
Deblurring&2020& WLRTR \cite{c116} & Weighted Tucker decomposition & Spectral correlation + Non-local similarity & https://owuchangyuo.github.io/publications/WLRTR \\
&2021& OLRT \cite{c117} & Joint spectral and non-local LR tensor & Non-local similarity +Spectral smoothness & https://owuchangyuo.github.io/publications/OLRT \\
\cline{2-6}
Inpaninting & 2015  &  TMac \cite{c118}& LR TC by parallel matrix factorization (TMac)  & Spectral correlation&   \\
& 2015 & TNCP \cite{c119} & TNN + CP decomposition             & Spectral correlation  & \\
& 2017 & AWTC \cite{c121} & HaLRTC with well-designed weights  & Spectral correlation & \\
& 2019 & LRRTC \cite{c130}& logarithm of the determinant + TTN & Spectral correlation  & \\
& 2020 & LRTC \cite{c123,c124} & t-SVD             & Spectral correlation   & \\
& 2019 & TRTV \cite{c128} &  TR decomposition + spatial TV     & Spectral correlation + Spatial smoothness  & \\
&2020& WLRTR \cite{c116} & Weighted Tucker decomposition & Spectral correlation + Non-local similarity & https://owuchangyuo.github.io/publications/WLRTR \\
& 2021 & TVWTR \cite{c129}& Weighted TR decomposition + 3DTV   & Spectral correlation + Spatial Spectral smoothness& https://github.com/minghuawang666/TVWTR \\
\cline{2-6}
Destriping& 2018 & LRTD \cite{c132}& Tucker decomposition + Spatial \& Spectral TV & Spectral correlation + Spatial Spectral smoothness &  https://github.com/zhaoxile?tab=repositories \\
& 2018 & LRNLTV \cite{c135} & Matrix nuclear norm + Non-local TV & Spectral correlation + Non-local similarity &\\
& 2020 & GLTSA \cite{c133}  & Global and local tensor sparse approximation & Sparisity + Spatial \& Spectral smoothness& \\
&2020& WLRTR \cite{c116} & Weighted Tucker decomposition & Spectral correlation + Non-local similarity & https://owuchangyuo.github.io/publications/WLRTR \\
\hline
CS  & 2017 & JTenRe3-DTV \cite{c97} & Tucker decomposition + weighted 3-D TV & Spectral correlation + Spatial \& Spectral smoothness & https://github.com/andrew-pengjj/Enhanced-3DTV\\
& 2017 & PLTD \cite{c101}       & Non-local Tucker decomposition             & Spectral correlation + Non-local similarity      & \\
& 2019 & NTSRLR \cite{c99}      & TNN + Tucker decomposition                 & Spectral correlation + Non-local similarity      &  \\
& 2020 & SNLTR \cite{c90}       & TR decomposition + Subspace representation & Non-local similarity + +Spectral smoothness  & \\
& 2015 & 3D-KCHSI \cite{c106}   & KCS with independent sampling dimensions   & Spectral correlation &   \\
& 2015 & T-NCS \cite{c95}       & Tucker decomposition                       & Spectral correlation & \\
& 2013 & NBOMP \cite{6797642}   & KCS with a tensor-based greedy algorithm   & Spectral correlation & \\
& 2016 & BOSE \cite{7544443}    & KCS with beamformed mode-based sparse estimator & Spectral correlation &\\
& 2020 & TBR \cite{c107}        & KCS with multi-dimensional block-sparsity  & Spectral correlation &\\
\hline
AD & 2015 & LTDD \cite{c110} & Tucker decomposition + Umixing & Spectral correlation & \\
& 2016 & TenB \cite{c111} & Tucker decomposition + PCA & Spectral correlation & \\
& 2019 & TDCW \cite{c139} & Tucker decomposition + Clustering & Spectral correlation & \\
& 2020 & TEELRD \cite{c113} & Tucker decomposition + Endmember extraction & Spectral correlation + Subspace Learning &\\
& 2019 & LRASTD \cite{c115} & Tucker decomposition +TNN & Spectral correlation + Subspace Learning & \\
& 2018 & TPCA \cite{c112} &  TPCA + Fourier transform & Spectral correlation  & \\ 
& 2020 & PTA \cite{c114}  & TRNN + Spatial TV & Spectral correlation + Spatial smoothness & https://github.com/l7170/PTA-HAD \\
& 2022 & PCA-TLRSR \cite{minghuaTC} & weighted TNN + Multi-subspace & Spectral correlation + Subspace & https://github.com/minghuawang666/ \\
\hline
SR & 2018 & STEREO \cite{c145} & CP decomposition & Spectral correlation & https://github.com/marhar19/HSR\_via\_tensor\_decomposition \\
& 2020 & NCTCP \cite{c152} & Nonlocal coupled CP decomposition & Spectral correlation + Non-local similarity & \\
& 2018 & SCUBA \cite{c151} & CP decomposition with matrix factorization &Spectral correlation & \\
& 2018 & CSTF \cite{c146} & Tucker decomposition & Spectral correlation & https://github.com/renweidian/CSTF \\
& 2021 & CT/CB-STAR \cite{c155} & Tucker decomposition with inter-image variability & Spectral correlation + Spatial Spectral Variability & https://github.com/ricardoborsoi
\\
& 2021 & CNTD \cite{c170} & Nonnegative Tucker decomposition & Spectral correlation &\\
& 2018 & CSTF-$l_2$ \cite{c147} & Tucker decomposition & Spectral correlation &\\
& 2020 & SCOTT \cite{c153} & Tucker decomposition + HOSVD & Spectral correlation & https://github.com/cprevost4/HSR$\_$Software\\
& 2020 & NNSTF \cite{c154} & Tucker decomposition + HOSVD & Spectral correlation + Non-local similarity& \\
& 2020 & WLRTR \cite{c116} & Weighted Tucker decomposition & Spectral correlation + Non-local similarity & https://owuchangyuo.github.io/publications/WLRTR \\
& 2017 & NLSTF \cite{c156} & Non-local sparse Tucker decomposition &Spectral correlation + Non-local similarity & https://github.com/renweidian/NLSTF \\
& 2020 & NLSTF-SMBF \cite{c157} & Non-local sparse Tucker decomposition &Spectral correlation + Non-local similarity & https://github.com/renweidian/NLSTF \\
& 2020 & UTVTD \cite{c160} & Tucker decomposition + Unidirectional TV &Spectral correlation + Spatial-Spectral smoothness & https://liangjiandeng.github.io/ \\
& 2020 & NLRTD-SU \cite{c158} & Non-local Tucker decomposition + SU +3-DTV & Spectral correlation + Non-local similarity + Spatial-Spectral smoothness \\
& 2018 & SSGLRTD \cite{c159} & Spatial–spectral-graph Tucker decomposition & Spectral correlation + Local geometry & \\
& 2021 & gLGCTD \cite{c161} & Graph Laplacian-guided Tucker decomposition & Spectral correlation + Local geometry & \\
& 2019 & NN-CBTD \cite{c163} & BT decomposition & Spectral correlation & \\
& 2021 & BSC-LL1 \cite{c138} & BT decomposition &Spectral correlation & https://github.com/MengDing56 \\
& 2021 & GLCBTD \cite{c164} & Graph Laplacian-guided BT decomposition & Spectral correlation + Local Geometry \\
& 2019 & LTTR \cite{c148} & Non-local TT decomposition &  Spectral correlation + Non-local similarity &https://github.com/renweidian/LTTR \\
& 2021 & NLRSR \cite{c162} & Non-local TT decomposition & Spatial-Spectral correlation + Non-local similarity\\
& 2022 & CTRF \cite{c150} & Coupled TR decomposition & Spectral correlation & \\
& 2020 & HCTR \cite{c149} & High-Order Coupled TR decomposition & Spectral correlation + Local Geometry & \\
& 2021 & FSTRD \cite{c169} & TR decomposition + TV &Spectral correlation + Spatial-Spectral smoothness & \\
& 2021 & LRTRTNN \cite{c171} & Non-local TR decomposition + TNN & Spectral correlation + Non-local similarity&\\
& 2019 & LTMR \cite{c165} & Subspace based LR multi-Rank  &  Spectral correlation + Non-local similarity &https://github.com/renweidian/LTMR \\
& 2021 & FLTMR \cite{c166} & LTMR with a Truncation Concept &  Spectral correlation + Non-local similarity& \\
& 2019 & NPTSR \cite{c8} & Non-local tensor sparse representation & Spectral correlation + Non-local similarity& \\
& 2019 & TV-TLMR \cite{c168} & Tucker decomposition + TV &  Spectral correlation + Spatial-Spectral smoothness & \\
& 2021 & LRTA-SR \cite{c167} & TTN & Spectral correlation & \\
\hline 
SU& 2007 & NTF-SU \cite{c190,c191} & CP decomposition & Spectral correlation & \\
& 2020 & ULTRA-V \cite{c212} & CP decomposition & Spectral correlation & https://github.com/talesimbiriba/ULTRA-V\\
& 2017 & MVNTF \cite{c192} & BT decomposition & Spectral correlation & https://gitSUb.com/bearshng/mvntf \\
& 2019 & NTF-TV \cite{c193} & TV + BT decomposition&	Spectral correlation + Spatial-Spectral smoothness &	http://www.xiongfuli.com/cv/ \\
& 2021 & SPLRTF \cite{c194}	&LR + sparsity + BT decomposition &	Spectral correlation & \\	
& 2019 & svr-MVNTF \cite{c195} &  BT decomposition	& Spectral correlation + Local Geometry \\
& 2020 & SCNMTF \cite{c196}	& BT decomposition + NMF &	Spectral correlations& \\
& 2021 & NLTR \cite{c197} & TV + Non-local LR &	Spectral correlations+ Nonlocal similarity+ Spatial-Spectral smoothness	& \\
& 2021 & BUTTDL1 \cite{c208} & sparsity + Tucker decomposition & Spectral correlations \\
& 2021 & SeCoDe \cite{c198} & Convolution operation + BT decomposition &	Spectral correlations +  Spatial-Spectral smoothness &	https://gitSUb.com/danfenghong/IEEE\_TGRS\_SeCoDe \\
& 2020 & WNLTDSU \cite{c210} & Weighted non-local LR + TV & Spectral correlation + Sparsity + Spatial smoothness & https://github.com/sunlecncom/WNLTDSU\\
& 2021 & NL-TSUn \cite{c210} &  Non-local LR + Joint sparsity & Spectral correlation + Sparsity & \\
& 2021 & LRNTF\cite{c204} &	BT decomposition &	Spectral correlations &	https://gitSUb.com/LinaZhuang/HSI\_nonlinear\_unmixing\_LR\-NTF \\
\hline
\end{tabular}}
\label{tab:tab1}
\end{table*}

\section{Notations and Preliminaries}
\label{sect:Notations}
In this section, we introduce some notations and preliminaries. For clear description, the notations are list in Table \ref{tab:tab0}. The main abbreviations used in this article are given in Table \ref{tab:abbreviation}.

\begin{table}[!htbp]
\centering
\caption{The notations used in the paper}
\begin{spacing}{1.3}
\scalebox{0.75}{
\begin{tabular}{c|c}
\hline
\cline{1-2}
\cline{1-2}
Notation&Description\\
\hline

\cline{1-2}
$x$ &scalars\\
$\textbf{x}$ & vectors\\
$\textbf{X}$ & matrices\\
$vec(\textbf{X})$ & $vec(\textbf{X}$) stacks the columns of $\textbf{X}$\\
$\mathcal{X} \in \mathbb{R}^{h \times v \times z}$ & tensors with 3-modes\\
$\mathcal{X}_{h_i,v_i,z_i}$ & the ($h_i,v_i,z_i$)-element of $\mathcal{X}$\\
$\mathcal{X}(i,:,:)$, $\mathcal{X}(:,i,:)$ and $\mathcal{X}(:,:,i)$ & the $i^{th}$ horizontal, lateral and frontal slices \\
 $||\mathcal{X}||_1=\sum_{h_i,v_i,z_i}{|\mathcal{X}_{h_i,v_i,z_i}|}$ & $l_1$ norm\\
 $||\mathcal{X}||_F= \sqrt{\sum_{h_i,v_i,z_i}{|\mathcal{X}_{h_i,v_i,z_i}|^2}}$  & Frobenius norm\\
 ${\sigma}_i(\textbf{X})$ & the singular values of matrix $\textbf{X}$ \\
   $||\textbf{X}||_* = \sum_i {\sigma}_i(\textbf{X})$ & nuclear norm\\   
 $||\textbf{x}||_2 = \sqrt{\sum_i {|\textbf{x}_i|^2}}$  & $l_2$ norm \\
  $ \hat{\mathcal{X}}=$fft$(\mathcal{X},[],3)$ &  Fourier transformation of $\mathcal{X}$ along mode-3\\
\hline
\end{tabular}
}
\end{spacing}
\label{tab:tab0}
\end{table}

\begin{table}[!htbp]
\centering
\caption{Main abbreviations used in the paper}
\begin{spacing}{1.3}
\scalebox{0.75}{
\begin{tabular}{c|c}
\hline
\cline{1-2}
\cline{1-2}
Abbreviation &Full name\\
\hline
AD  & Anomaly detection\\
CP  & CANDECOMP/PARAFAC\\
CS  & Compressive sensing\\
BT  & Block term\\
HS  & Hyperspectral\\
LMM & Linear mixing model\\
LR  & Low-rank\\
NMF & Nonnegative matrix factorization\\
RS  & Remote sensing\\
SNN & The sum of the nuclear norm\\
SU  & Spectral unmixing\\
TNN & Tensor nuclear norm\\
TV  & Total variation\\
TT & Tensor train\\
TTN & Tensor trace norm\\
TR & Tensor ring\\
1-D & One dimensional\\
2-D & Two dimensional\\
3-D & Three dimensional\\
4-D & Four dimensional\\
\hline
\end{tabular}
}
\end{spacing}
\label{tab:abbreviation}
\end{table}

$\textbf{Definition 1}$ (T-product \cite{c43}): The T-product of two three-order tensors $\mathcal{A} \in \mathbb{R}^{n_1 \times n_2 \times n_3}$ and $\mathcal{B} \in \mathbb{R}^{n_2 \times n_4 \times n_3}$ is denoted by $\mathcal{C} \in \mathbb{R}^{n_1 \times n_4 \times n_3}$:
\begin{equation}
\label{eq:product}
\mathcal{C}(i,k,:)= \sum_{j=1}^{n_2} \mathcal{A}(i,j,:) \star \mathcal{B}(j,k,:)
\end{equation}
where $\star$ represents the circular convolution between two tubes.

$\textbf{Definition 2}$ (Tensor $n$-mode product \cite{c181}): The $n$-mode product of a tensor $\mathcal{A} \in \mathbb{R}^{r_1  \times r_2  \times ...   \times r_N }$ and a matrix $\mathbf{B} \in \mathbb{R}^{ B \times r_n}$ is the tensor $\mathcal{X} \in \mathbb{R}^{r_1  \times r_2  \times ... r_{n-1} \times B \times r_{n+1} ...   \times r_N } $ defined by
\begin{equation}
\begin{aligned}
\label{eq:sec-product}
\mathcal{X} = \mathcal{A} \times_n \mathbf{B}
\end{aligned}
\end{equation}
The unfolding matrix form of Eq.(\ref{eq:sec-product}) is
\begin{equation}
\begin{aligned}
\label{eq:sec-product2}
\mathbf{X}_{(n)} = \mathbf{B}  \times \mathbf{A}_{(n)}  
\end{aligned}
\end{equation}

$\textbf{Definition 3}$ (Four Special tensors \cite{c27}):

Conjugate transpose: The conjugate transpose of a three-order tensor $ \mathcal{X}\in \mathbb{R}^{h \times v \times z}$ is the tensor $ {\rm conj}(\mathcal{X})=\mathcal{X}^*  \in \mathbb{R}^{v \times h \times z}$, which can be obtained by conjugately transposing each front slice and reversing the order of transposed frontal 2 through $z$.

Identity tensor: The identity tensor denoted by $\mathcal{I}\in \mathbb{R}^{h \times v \times z}$ is the tensor whose first frontal slice is an identity matrix and all other frontal slices are zero.

Orthogonal tensor: A three-order tensor $\mathcal{Q}$ is orthogonal if it satisfies $\mathcal{Q}^* * \mathcal{Q}= \mathcal{Q} * \mathcal{Q}^*=\mathcal{I}$.

F-diagonal tensor: A three-order tensor $\mathcal{S}$ is f-diagonal if all of its slices are diagonal matrices.

$\textbf{Definition 4}$ (First Mode-$k$ Unfolding/matricization \cite{c181}): This operator noted unfold$(\mathcal{X},k)$ converts a tensor $\mathcal{X} \in \mathbb{R}^{I_1 ... I_k \times I_{k+1} ...I_N }$ into a matrix $\textbf{X}_{(k)} \in \mathbb{R}^{I_k \times I_1..I_{k-1}I_{k+1}...I_N }$. Inversely, fold($\textbf{X}_{(k)}, k$) denotes the folding of the matrix into a tensor.

$\textbf{Definition 5}$ (Second Mode-$k$ Unfolding/matricization \cite{c126}): For a tensor $\mathcal{X} \in \mathbb{R}^{I_1 ... I_k \times I_{k+1} ...I_N }$, its second Mode-$k$ Unfolding matrix represented by $\textbf{X}_{<k>} \in \mathbb{R}^{I_k \times I_{k+1}...I_N I_1..I_{k-1} }$. The inverse operation is matrix folding (tensorization).

$\textbf{Definition 6}$ (Mode-$k$ permutation \cite{c29}):  
For a three tensor $\mathcal{X} \in \mathbb{R}^{ I_1 \times I_2 \times  ... \times I_N }$, this operator noted by $\mathcal{X}^k$=permutation($\mathcal{X}$, $k$) changes its permutation order with $k$ times and obtain a new tensor $\mathcal{X}^k \in \mathbb{R}^{ I_k \times ...  \times I_N \times I_1 \times  ... \times I_{k-1} } $. The inverse operator is defined as $\mathcal{X}$ = ipermutation($\mathcal{X}^k$, $k$). For example, three mode-$k$ permutation of an HS tensor $\mathcal{X}^1\in \mathbb{R}^{  h \times v \times z }$ can be written as
$\mathcal{X}^1\in \mathbb{R}^{v \times z \times h}$, $\mathcal{X}^2\in \mathbb{R}^{z \times h \times v}$, $\mathcal{X}^3\in \mathbb{R}^{h \times v \times z}$.

$\textbf{Definition 7}$ (Tensor Trace Norm (TTN) \cite{c44}) It is the sum of the nuclear norm (SNN) of the mode-$k$ unfolding matrix for a $3$-way HS tensor:
\begin{equation}
\label{eq:tracenorm}
||\mathcal{X}||_{\rm SNN}:=\sum_{k=1}^{3} \alpha _k ||\textbf{X}_{(k)}||_*
\end{equation}
where weights $\alpha_k$ satisfy $\alpha_k \geq 0 (k=1,2,3)$ and $\sum_{k=1}^{3} \alpha_k =1$. 

$\textbf{Definition 8}$ (Tucker decomposition \cite{c181,c186,c187}): The Tucker decomposition of an $N$-order tensor $\mathcal{X} \in \mathbb{R}^{I_1 \times I_2 \times  ... \times I_N}$ is defined as
\begin{equation}
\begin{aligned}
\label{eq:Tucker1}
\mathcal{X}= \mathcal{A} \times_1 \mathbf{B}_1 \times_2 \mathbf{B}_2 ... \times_N \mathbf{B}_N
\end{aligned}
\end{equation}
where $\mathcal{A} \in \mathbb{R}^{r_1 \times r_2 \times... \times r_N}$ stands for a core tensor and $\mathbf{B}_n \in \mathbb{R}^{I_n \times r_n}, n=1,2,..,N$ represent factor matrices. The Tucker ranks are represented by ${\rm rank}_{\rm Tucker}(\mathcal{X}) = [r_1, r_2,..., r_N] $.

$\textbf{Definition 9}$ (CP decomposition \cite{c180,c181,c188}):
The CP decomposition of an $N$-order tensor $\mathcal{X} \in \mathbb{R}^{I_1 \times I_2 \times  ... \times I_N} $ is defined as
\begin{equation}
\begin{aligned}
\label{eq:CP1}
 \mathcal { X }=\sum_{r=1}^{R} \tau_r \mathbf{b}^{(1)}_{r} \circ \mathbf{b}^{(2)}_{r} \circ ... \circ \mathbf{b}^{(N)}_{r} 
\end{aligned}
\end{equation}
where $\tau_r$ are non-zero weight parameters, and $\mathbf{b}^{(1)}_{r} \circ \mathbf{b}^{(2)}_{r} \circ ... \circ \mathbf{b}^{(N)}_{r} $ denotes a rank-one tensor with $\mathbf{b}^{(n)}_{r} \in \mathbb{R}^{I_n}$. The CP rank denoted by ${\rm rank}_{\rm CP}(\mathcal{X}) = R $ is the sum number of rank-one tensors.

$\textbf{Definition 10}$ ( BT decomposition \cite{c213}) 
The BT decomposition of an three-order tensor $\mathcal{X} \in \mathbb{R}^{h \times v \times z} $ is defined as
\begin{equation}
    \begin{aligned}
    \label{eq:BTD1}
    \mathcal{X} = \sum_{r=1}^{R} \mathcal{G}_r \times_1 \mathbf{A}_r \times_2 \mathbf{B}_r \times_3 \mathbf{C}_r
    \end{aligned}
\end{equation}
where $\mathcal{G} \in \mathbb{R}^{L_h \times L_v \times L_z}$, $\mathbf{A}_r \in \mathbb{R}^{h \times L_h}$, $\mathbf{B}_r \in \mathbb{R}^{v \times L_v}$, and $\mathbf{C}_r \in \mathbb{R}^{z \times L_z}$. Each of $R$ component tensors can be expressed by rank ($L_h$, $L_v$, $L_z$) Tucker decomposition. BT decomposition can be regarded as the combination of Tucker and CP decomposition. On the hand, Eq. (\ref{eq:BTD1}) becomes Tucker decomposition when $R=1$. On the other hand, when each component is represented by a rank ($L$, $L$, $1$) tensor, Eq. (\ref{eq:BTD1}) is written by
\begin{equation}
    \begin{aligned}
    \label{eq:BTD2}
    \mathcal{X}  =\sum_{r=1}^{R} \mathbf{A}_{r} \cdot \mathbf{B}_{r}^{T} \circ \mathbf{c}_{r} 
    \end{aligned}
\end{equation}
where the matrix $\mathbf{A}_{r} \in \mathbb{R}^{h \times L_r}$ and the matrix $\mathbf{B}_{r} \in \mathbb{R}^{v \times L_r}$ are also rank-$L$. If rank-$L$ $\mathbf{E}_{r} \in \mathbb{R}^{h \times v}$ is factorized as $\mathbf{A}_{r} \cdot \mathbf{B}_{r}^{T}$. Eq. (\ref{eq:BTD2}) can be rewritten as
\begin{equation}
    \begin{aligned}
    \label{eq:BTD3}
           \mathcal{X} =\sum_{r=1}^{R} \mathbf{E}_{r} \circ \mathbf{c}_{r} 
    \end{aligned}
\end{equation}

$\textbf{Definition 11}$ (Tensor Nuclear Norm (TNN) \cite{c123}) Let $\mathcal{X}=\mathcal{U} * \mathcal{S} * \mathcal{V}^{*}$ be the t-SVD of $\mathcal{X} \in \mathbb{R}^{h \times v \times z}$, TNN is the sum of singular values of $\mathcal{X}$, that is,
\begin{equation}
\label{eq:TNN}
||\mathcal{X}||_*:=\sum_{k=1}^{z} \mathcal{S}(k,k,1),
\end{equation}
and also can be expressed as the sum of nuclear norm of all the frontal slices of $\hat{\mathcal{X}}$ :
\begin{equation}
\label{eq:TNN2}
||\mathcal{X}||_*:=\sum_{k=1}^{z} ||\hat{\mathcal{X}}(:,:,k)||_*.
\end{equation}

\begin{figure*}[htb]
	\begin{center}
		\includegraphics[width = 1\textwidth]{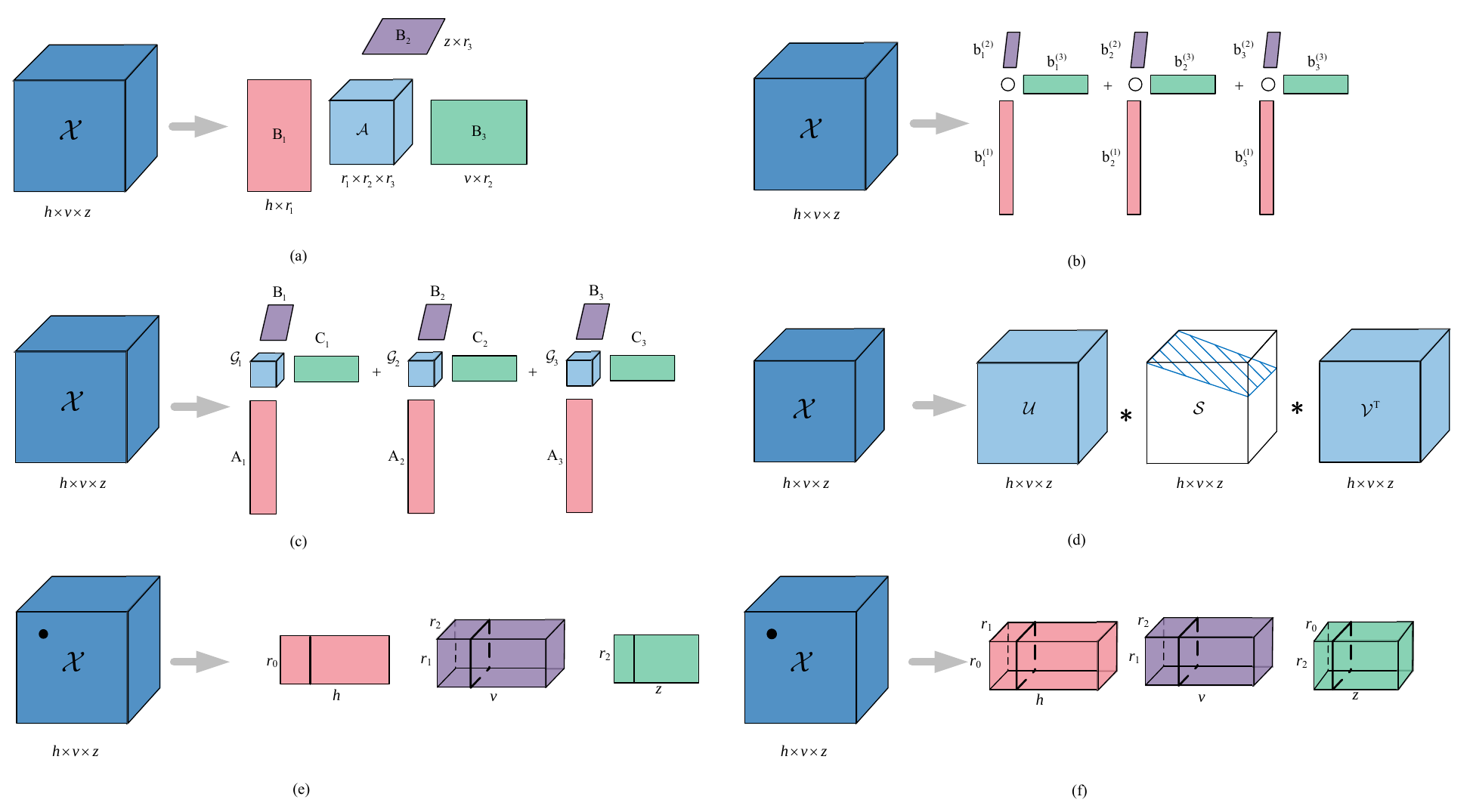}
	\end{center}
	\caption[restoration]{ Illustration to show six tensor decompositions of third-order tensor: (a) Tucker decomposition, (b) CP decomposition, (c) BT decomposition, (d) t-SVD, (e) TT decomposition, (f) TR decomposition. }
	\label{fig:td}
\end{figure*}

For more intuitively understanding the above mentioned tensor decompositions, examples for third-order tensor are shown in Fig. \ref{fig:td}, which benefits the consequent tensor decomposition-based researches of third-order HS data.

$\textbf{Definition 12}$ (t-SVD \cite{c123}) $\mathcal{X} \in \mathbb{R}^{h \times v \times z}$ can be factorized as
\begin{equation}
\label{eq:tsvd}
\mathcal{X}=\mathcal{U} * \mathcal{S} * \mathcal{V}^{*}
\end{equation}
where $\mathcal{U} \in \mathbb{R}^{h \times h \times z}$, $\mathcal{V} \in \mathbb{R}^{v \times v \times z}$ are orthogonal tensors and $\mathcal{S} \in \mathbb{R}^{h \times v \times z}$ is a f-diagonal tensor. The details of t-SVD are described in the Algorithm \ref{alg:tsvd1}.
\begin{algorithm}[htb]
	\caption{t-SVD} \label{alg:tsvd1}
	\begin{algorithmic}[1]
		\REQUIRE $\;  \mathcal{X}\in \mathbb{R}^{h \times v \times z}$\\
		\NoDo
        \STATE $ \hat{\mathcal{X}}=$fft$(\mathcal{X},[],3)$;
		\NoDo \FOR{$i=0,1,\dots, [\frac{z+1}{2}]$}
		\STATE $ [\hat{\mathcal{U}}(:,:,i), \hat{\mathcal{S}}(:,:,i), \hat{\mathcal{V}}(:,:,i)]=$SVD$(\hat{\mathcal{X}}(:,:,i))$;		
		\ENDFOR
		\NoDo \FOR{$i= [\frac{z+1}{2}+1],\dots,z$}
		\STATE $ \hat{\mathcal{U}}(:,:,i)=$conj($\hat{\mathcal{U}}(:,:,z-i+2)$);
	\STATE $ \hat{\mathcal{S}}(:,:,i)=(\hat{\mathcal{S}}(:,:,z-i+2)$)
	\STATE$ \hat{\mathcal{V}}(:,:,i)=$conj($\hat{\mathcal{V}}(:,:,z-i+2)$);		
		\ENDFOR
		\STATE $\mathcal{U} =$ifft$(\hat{\mathcal{U}},[],3)$, $\mathcal{S} =$Ifft$(\hat{\mathcal{S}},[],3)$,$ \mathcal{V}=$fft$(\hat{\mathcal{V}},[],3)$;
		\ENSURE $\mathcal{U},\mathcal{S},\mathcal{V}$.
	\end{algorithmic}
\end{algorithm}

$\textbf{Definition 13}$ ( TT decomposition \cite{c185}) The TT decomposition of an $N$-order $\mathcal{X} \in \mathbb{R}^{I_1 \times I_2 \times  ... \times I_N} $ is represented by cores $\mathcal{G} = \{ \mathcal{G}^{(1)},..., \mathcal{G}^{(N)} \}$, where $\mathcal{G}^{(n)} \in \mathbb{R}^{ r_{n-1} \times I_n \times r_n} $, $n=1,2,...,N$, $r_0 = r_N =1$. The rank of TT decomposition is defined as $ {\rm rank}_{\rm TT}(\mathcal{X}) =[r_0,r_1,...,r_{N}]$. Each entry of the tensor $\mathcal{X}$ is formulated as
\begin{equation}
\label{eq:ttd}
\mathcal{X}(i_1,...,1_N) = \mathcal{G}^{(1)}(:,i_1,:)\mathcal{G}^{(2)}(:,i_2,:)...\mathcal{G}^{(N)}(:,i_N,:).
\end{equation}

$\textbf{Definition 14} $ (TR decomposition \cite{c126}): The purpose of TR decomposition is to represent a high-order $\mathcal{X}$ by multi-linear products of a sequence of three-order tensors in circular form. Three-order tensors are named TR factors $\{\mathcal{G}^{(n)}\}^N_{n=1} = \{\mathcal{G}^{(1)}, \mathcal{G}^{(2)},...,\mathcal{G}^{(N)}\}$, where $\mathcal{G} ^{(n)}\in \mathbb{R}^{r_n \times I_n \times r_{n+1}}$, $n=1, 2, ..., N$, $r_0 =r_N$. In this case, the element-wise relationship of TR decomposition with factors $\mathcal{G}$ can be written as
\begin{equation}
\label{eq:TR1}
\begin{aligned}
\mathcal{X}(i_1, i_2,...,i_N) &= {\rm Tr}(\mathcal{G}^{(1)}(:,i_1,:)\mathcal{G}^{(2)}(:,i_2,:)...\mathcal{G}^{(N)}(:,i_N,:))\\
&= {\rm Tr}(\prod_{i=1}^{N} \mathcal{G}^{(n)}(:,i_n,:))
\end{aligned}
\end{equation}  
where ${\rm Tr}$ denotes the matrix trace operation.

$\textbf{Definition 15} $ (Multi-linear Product \cite{c16}): Given two TR factors $\mathcal{G}^{(n)}$ and $\mathcal{G}^{(n+1)}$, their multi-linear product $\mathcal{G}^{(n,n+1)} \in \mathbb{R}^{r_n \times I_n I_{n+1}\times r_{n+1}}$ is calculated as
\begin{equation}
\label{eq:multilinear}
\begin{aligned}
\mathcal{G}^{(n,n+1)}(:,I_n(i_k -1) +j_k,:)=\mathcal{G}^{(n)}(:,i_k,:)\mathcal{G}^{(n+1)}(:,j_k,:)
\end{aligned}
\end{equation}  
for $i_k = 1,2,...,I_n, j_k = 1, 2,...,I_n+1$.

From the above $\textbf{Definition 15} $, the multi-linear product of all the TR factors can be induce as $[\mathcal{G}] = \prod^N_{n=1}\mathcal{G}^{(n)} = \mathcal{G}^{(1,2,...,n)} = \{ \mathcal{G}^{(1)},\mathcal{G}^{(2)},...,\mathcal{G}^{(n)}\} \in \mathbb{R}^{r_1 \times I_1 I_2... I_n \times r_1}$. The TR decomposition can be rewritten as $\mathcal{X} = \Phi({\mathcal{G}})$, where $\Phi$ is a dimensional shifting operator $\Phi: \mathbb{R}^{r_1 \times I_1 I_2...I_n \times r_1} \rightarrow \mathbb{R}^{I_1 \times I_2 \times...\times I_n}$. 

$\textbf{Lemma 1} $ (Circular Dimensional Permutation Invarience \cite{c126}): If the TR decomposition of $\mathcal{X}$ is $\mathcal{X} = \Phi(\mathcal{G}^{(1)},\mathcal{G}^{(2)},...,\mathcal{G}^{(N)})$, ${\stackrel{\leftarrow}{\mathcal{X}}}^n \in \mathbb{R}^{I_n \times I_{n+1} \times ... \times I_1 \times ... \times I_{n-1}} $ is defined as circular shifting the dimensions of $\mathcal{X}$ by $n$, we obtain the following relation:
\begin{equation}
\label{eq:circluar}
\begin{aligned}
\stackrel{\leftarrow}{\mathcal{X}}^n = \Phi( \mathcal{G}^{(n)},\mathcal{G}^{(n+1)},...,\mathcal{G}^{(N)},\mathcal{G}^{(1)},\mathcal{G}^{(2)},...,\mathcal{G}^{(N)})
\end{aligned}
\end{equation}

\begin{figure*}[htb]
	\begin{center}
		\includegraphics[height=9cm]{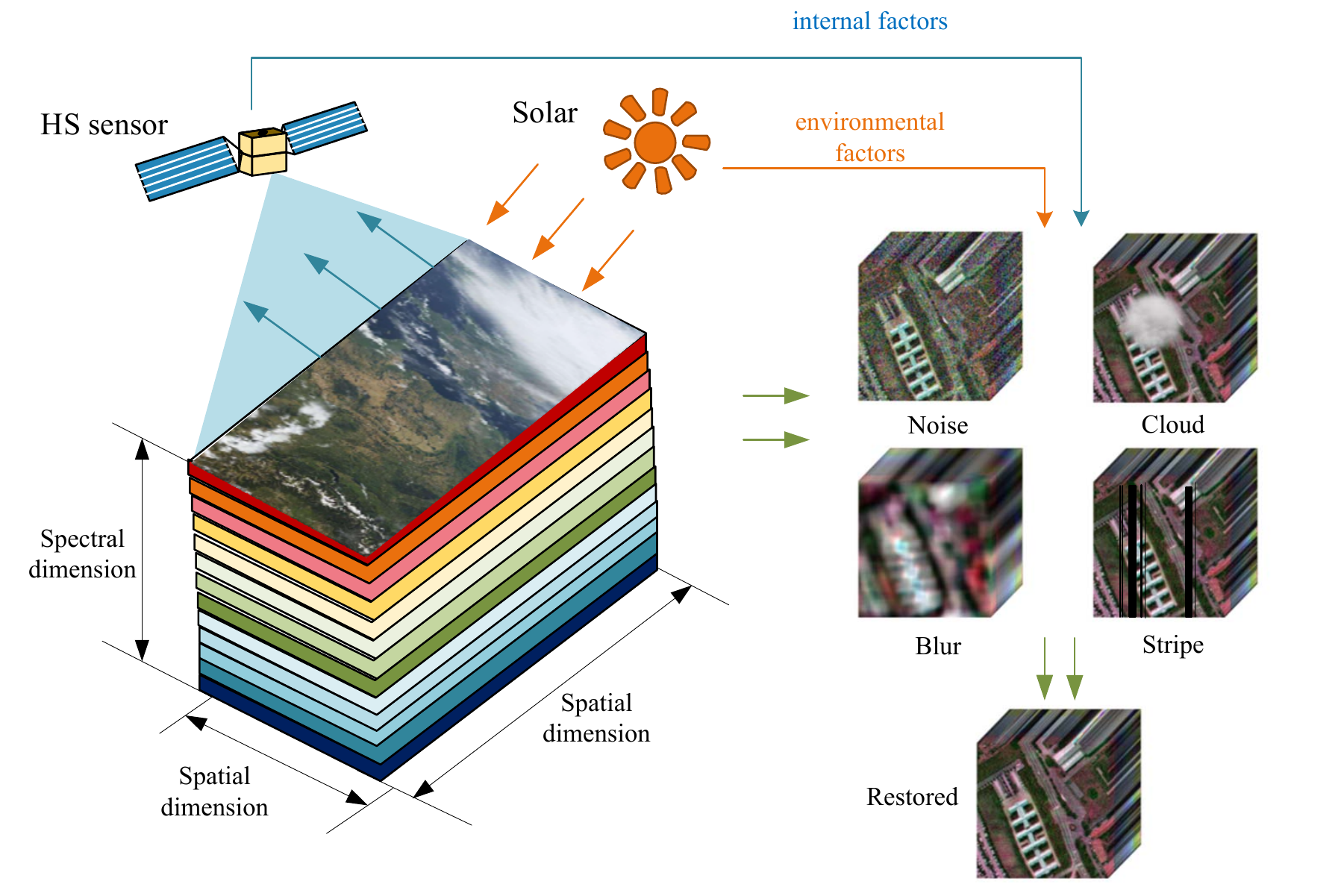}
	\end{center}
	\caption[restoration]{A schematic diagram of HS image restoration. }
	\label{fig:denoisingfr}
\end{figure*}

$\textbf{Definition 16} $ (Mixed $l_{1,0}$ pseudo-norm \cite{c41}): Given a vector $\textbf{y} \in \mathbb{R}^{m}$ and index sets $\theta_1,...,\theta_i,...,\theta_n (1 \leq n \leq m)$ that satisfies 
\begin{itemize}
\item Each $\theta_i$ is a subset of {1,...,m},
\item $\theta_i \cap \theta_l = \emptyset$ for any $i \neq l$,
\item $\cup^n_{i=1}\theta_i ={1,...,m}$,
\end{itemize}
the mixed $l_{1,0}$ pseudo-norm of $y$ is defined as:
\begin{equation}
||\textbf{y}||^{\theta}_{1,0} = ||(||\textbf{y}_{\theta_1}||_1,...,||\textbf{y}_{\theta_i}||_1,...||\textbf{y}_{\theta_n}||_1)||_0,
\end{equation}
where $\textbf{y}_{\theta_i}$ denotes a sub-vector of $\textbf{y}$ with its entries specified by $\theta_i$ and $|| \cdot ||_0$ calculates the number of the non-zero entries in ($\cdot$).


\section{HS Restoration}
\label{sect:restoration}

In the actual process of HS data acquisition and transformation, external environmental change and internal equipment conditions inevitably lead to noises, blurs, and missing data (including clouds and stripes) \cite{GRSM2022,li2021progressive} which degrade the visual quality of HS images and the efficiency of the subsequent HS data applications, such as a fine HS RS classification for crops and wetlands \cite{5779697,9598903} and the refinement of spectral information for target detection \cite{zhangTD2012,2009OE}. Fig. \ref{fig:denoisingfr} depicts the HS RS degradation and Restoration. Therefore, HS image restoration appears as a crucial pre-processing step for further applications. 

Mathematically, an observed degraded HS image can be formulated as follows
 \begin{equation}
	\begin{split}
		\label{eq:degrade}
\mathcal{T}=M(\mathcal{X}) + \mathcal{S} + \mathcal{N}
	\end{split}
\end{equation} 
where $\mathcal{T} \in \mathbb{R}^{h \times v \times z}$, $\mathcal{X} \in \mathbb{R}^{h \times v \times z}$, $\mathcal{S} \in \mathbb{R}^{h \times v \times z}$ and $\mathcal{N} \in \mathbb{R}^{h \times v \times z}$ represents an observed HS image, the restored HS image, the sparse error and additive noise, respectively, and $M(\cdot)$ denotes different linear degradation operators for different HS restoration problems: (a) when $M(\cdot)$ is a blur kernal also called as point spread function (PSF), Eq. (\ref{eq:degrade}) becomes HS deblurring problem; 
(b) when $M(\cdot)$ is a binary operation, i.e., 1 for original pixels, and 0 for missing data, Eq. (\ref{eq:degrade}) turns into the HS inpainting problem;
(c) when $M(\mathcal{X})$ keeps $\mathcal{X}$ constant, i.e., $M(\mathcal{X}) = \mathcal{X}$, Eq. (\ref{eq:degrade}) is reformulated as the HS destriping problem ($\mathcal{T}=\mathcal{X} + \mathcal{S}$) or HS denoising problem (only consider Gaussian noise $\mathcal{T}=\mathcal{X} + \mathcal{N}$ or consider mixed noise $\mathcal{T}=\mathcal{X} + \mathcal{S} + \mathcal{N}$). The HS restoration task is to estimate recovered HS images $\mathcal{X}$ from the given HS images $\mathcal{T}$. This ill-posed problem suggests that extra constraints on $\mathcal{X}$ need to be enforced for the optimal solution of $\mathcal{X}$. These additional constraints reveal the HS desired property and various types of HS prior information, such as non-local similarity, spatial and spectral smoothness, and subspace representation. The HS restoration problem can be summarized as 
\begin{equation}
    \begin{aligned}
    \label{eq:summ}
     \underset{\mathcal{X}}{ \min } \frac{1}{2} ||  \mathcal{T} - M(\mathcal{X}) - \mathcal{S}  ||^2_F + \tau f(\mathcal{X}) + \lambda g(\mathcal{S})
    \end{aligned}
\end{equation}
where $f(\mathcal{X})$ and $g(\mathcal{S})$ stand for the regularizations to explore the desired properties on the recovered $\mathcal{X}$ and sparse part $\mathcal{S}$, respectively. $\tau$ and $\lambda$ are regularization parameters.

\subsection{HS Denoising}
\label{sect:HS_Denoising}
\begin{figure}[htb]
	\begin{center}
		\includegraphics[height=3.2cm]{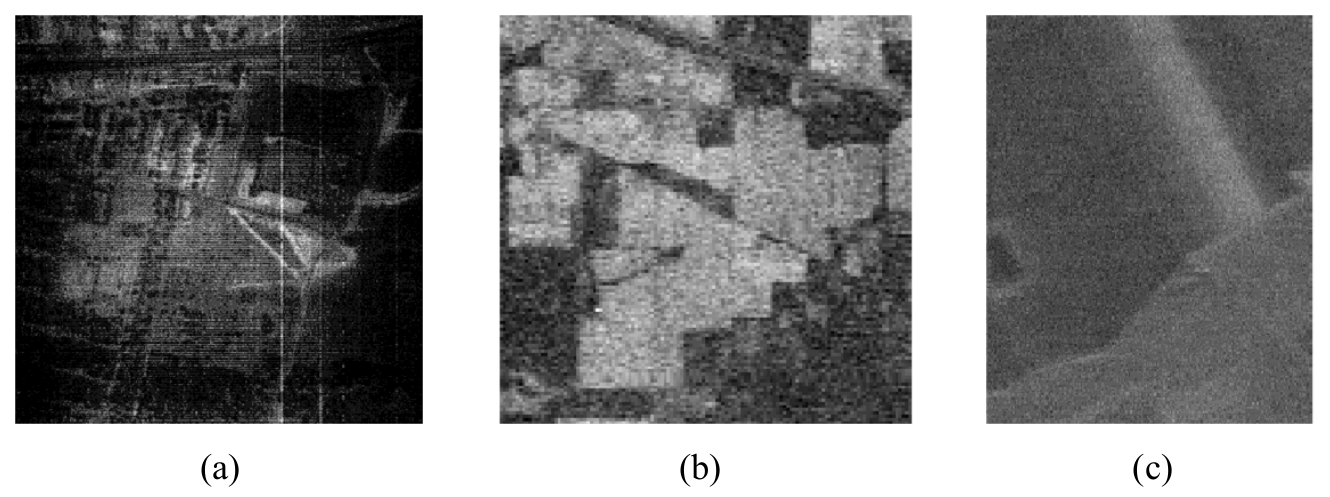}
	\end{center}
	\caption[washington]{HS data sets with different noise types: (a) the Urban data set, (b) the Indian Pines data set, (c) the Salinas data set. }
	\label{fig:Visio-noisyHSI}
\end{figure}

The observed HS images are often corrupted by mixed noise, including Gaussian noise, salt and pepper noise, and dead-line noise. Several noise types of HS images are shown in Fig. \ref{fig:Visio-noisyHSI}. 
The wealthy spatial and spectral information of HS images can be extracted by different prior constraints like LR property, sparse representation, non-local similarity, and total variation. Different LR tensor decomposition models are introduced for HS denoising. Consequently, one or two kinds of other prior constraints are combined with these tensor decomposition models.

\subsubsection{LR Tensor Decomposition}
\label{sect:LRT}
 
In this section, the LR tensor decomposition methods are divided into two categories: 1) factorization-based approaches and 2) rank minimization-based approaches. The former one needs to predefine rank values and update decomposition factors. The latter directly minimizes tensor ranks and updates LR tensors.
 
$\textbf{(1) Factorization-based approaches}$ 

Two typical representatives are used in the HS image denoising literature, namely, Tucker decomposition and CP decomposition. Renard \textit{et al}. \cite{c25} considered Gaussian noise and suggested a LR tensor approximation (LRTA) model to complete an HS image denoising task:
 \begin{equation}
	\begin{split}
		\label{eq:lrta}
&\underset{\mathcal{X}}{\textrm{min}}\;||\mathcal{T} - \mathcal{X} ||^2_F \\ 
&{\rm s.t.} \; \mathcal{X}= \mathcal{A} \times_1 \mathbf{B}_1 \times_2 \mathbf{B}_2 \times_3 \mathbf{B}_3
	\end{split}
\end{equation} 

 Nevertheless, users should manually pre-define the multiple ranks along all modes before running the Tucker decomposition-related algorithm, which is intractable in reality. In Eq. (\ref{eq:lrta}), the Tucker decomposition constraint is easily replaced by other tensor decomposition, such as CP decomposition. Liu \textit{et al}. \cite{c26} used a Parallel Factor Analysis (PARAFAC) decomposition algorithm and still assumed that HS images were corrupted by white Gaussian noise. Guo \textit{et al}. \cite{c59} presented an HS image noise-reduction model via rank-1 tensor decomposition, which was capable of extracting the signal-dominant features. However, the smallest number of rank-1 factors is served as the CP rank, which needs high computation cost to be calculated. 

 $\textbf{(2) Rank minimization approaches}$ 
 
 The tensor rank bounds are rarely available in many HS noisy scenes. To avoid the occurrence of rank estimation, another kind of methods focus on minimizing the tensor rank directly, which can be formulated as follows:
  \begin{equation}
	\begin{split}
		\label{eq:rank_mini}
&\underset{\mathcal{X}}{\textrm{min}}\;  {\rm rank}(\mathcal{X}) \\ 
&{\rm s.t.} \; \mathcal{T}= \mathcal{X} + \mathcal{S} +  \mathcal{N} \\
	\end{split}
\end{equation} 
where ${\rm rank(\mathcal{X})}$ denotes the rank of HS tensor $\mathcal{X}$ and includes different rank definitions like Tucker rank, CP rank, TT rank, and tubal rank. Due to the above rank minimizations belong to non-convex problems, these problems are NP-hard to compute. Nuclear norms are generally used as the convex surrogate of non-convex rank function. Zhang \textit{et al}. \cite{c27} proposed a tubal rank related TNN to characterize the 3-D structural complexity of multi-linear data. Based on the TNN, Fan \textit{et al}. \cite{c28} presented an LR Tensor Recovery (LRTR) model to remove Gaussian noise and sparse noise:
  \begin{equation}
	\begin{split}
		\label{eq:LRTR}
&\underset{\mathcal{X},\mathcal{S},\mathcal{N}}{\textrm{min}}\;  ||\mathcal{X}||_* + \lambda_1 || \mathcal{S}||_1 + \lambda_2 ||\mathcal{N}||_F^2 \\ 
&{\rm s.t.} \; \mathcal{T}= \mathcal{X} + \mathcal{S} +  \mathcal{N} \\
	\end{split}
\end{equation} 

Xue \textit{et al}. \cite{xue2019nonconvex} applied a non-convex logarithmic surrogate function into a TTN for tensor completion and (tensor robust principal component analysis) TRPCA tasks. 
Zheng \textit{et al}. \cite{c61} explored the LR properties of tensors along three directions and proposed two tensor models: a three-directional TNN (3DTNN) and a three-directional log-based TNN (3DLogTNN) as its convex and nonconvex relaxation. Although these pure LR tensor decomposition approaches utilize the LR prior knowledge of HS images, they are hardly effective to suppress mixed noise due to the lack of other useful information. 

\subsubsection{Other priors regularized LR Tensor Decomposition}
\label{sect:OLRT}
Various types of priors are combined with an LR tensor decomposition model to optimize the model solution including non-local similarity, spatial and spectral smoothness, spatial sparsity, subspace learning. 

$\textbf{ (1) Non-local similarity }$

\begin{figure*}[htb]
	\begin{center}
		\includegraphics[height=10cm]{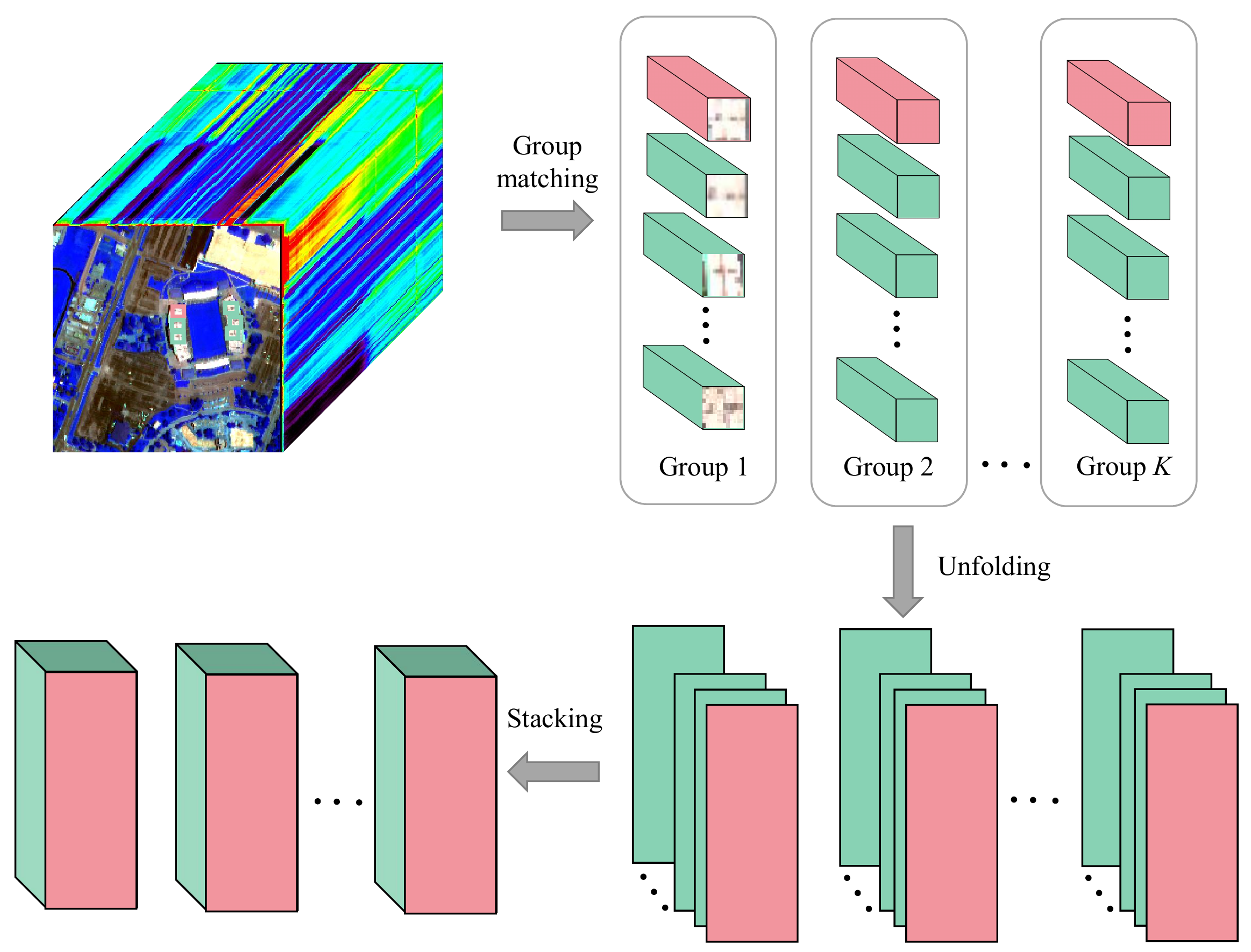}
	\end{center}
	\caption[nonlocal]{Flowchart of non-local LR tensor-based methods. }
	\label{fig:nonlocal}
\end{figure*}

An HS image often possesses many repetitive local spatial patterns, and thus a local patch always has many similar patches across this HS image \cite{c66}. Peng \textit{et al}. \cite{c62} designed a tensor dictionary learning (TDL) framework. In Fig. \ref{fig:nonlocal}, an HS image is segmented into 3-D full band patches (FBP). The similar FBPs are clustered together as a 4-D tensor group to simultaneously leverage the non-local similarity of spatial patches and the spectral correlation. TDL is the first model to exploit the non-local similarity and the LR tensor property of 4-D tensor groups, as shown in Fig. \ref{fig:nonlocal} (b). Instead of a traditional alternative least square based tucker decomposition, Bai \textit{et al}. \cite{c71} improved a hierarchical least square based nonnegative tucker decomposition method. Kong \textit{et al}. \cite{c67} incorporated the weighted tensor norm minimization into the Tucker decompositions of 4-D patches.

Differ from references \cite{c62,c71,c67}, other works \cite{c65,c63,xue2019nonlocal,chang2017hyper,he2019non,c64} obtained a 3-D tensor by stacking all non-local similar FBPs converted as matrices with a spatial mode and a spectral mode in Fig. \ref{fig:nonlocal}(d). Based on a non-local similar framework, Dong \textit{et al}. \cite{c65} proposed a Laplacian Scale Mixture (LSM) regularized LR tensor approximation method for denoising.
Xie \textit{et al}. \cite{c63} conducted a tensor sparsity regularization named intrinsic tensor sparsity (ITS) to encode the spatial and spectral correlation of the non-local similar FBP groups. With the non-local similarity of FBPs, $\mathcal{X}$ is estimated from its corruption $\mathcal{T}$ by solving the following problem 
  \begin{equation}
	\begin{split}
		\label{eq:ITS}
 \min _{\mathcal{X}} \Lambda (\mathcal{X})+\frac{\gamma}{2}\|\mathcal{T}_{i}-\mathcal{X}_i \|_{F}^{2}
	\end{split}
\end{equation} 
where the sparsity of a tensor $\mathcal{X}$ is $\Lambda (\mathcal{X})=t\|\mathcal{A}\|_{0}+(1-t) \prod_{i=1}^{N} {\rm rank}(X_{(i)})$ and $\mathcal{A}$ is the core tensor of $\mathcal{X}$ via the Tucker decomposition $\mathcal{X}= \mathcal{A} \times_1 \mathbf{B}_1 \times_2 \mathbf{B}_2 \times_3 \mathbf{B}_3$. Xue \textit{et al}. \cite{xue2019nonlocal}  presented a non-local LR regularized CP tensor decomposition (NLR-CPTD) algorithm. However, the Tucker or CP decomposition-related methods are subject to the heavy computational burden issues. 

Chang \textit{et al}. \cite{chang2017hyper} discovered the LR property of the non-local patches and used a hyper-Laplacian prior to model additional spectral information. He \textit{et al}. \cite{he2019non} developed a new paradigm, called non-local meets global (NGmeet) method, to fuse the spatial non-local similarity and the global spectral LR property. Chen \textit{et al}. \cite{c64} analyzed the advantages of a novel TR decomposition over the Tucker and CP decompositions. The proposed non-local TR decomposition method for HS image denoising is formulated as:
  \begin{equation}
	\begin{split}
		\label{eq:TR}
\min _{\mathcal{X}_{i}, \mathcal{G}_{i}} \frac{1}{2}\|\mathcal{T}_{i}-\mathcal{X}_{i}\|_{F}^{2} \quad \text { s.t. } \; \mathcal{X}_{i}=\Phi([\mathcal{G}_{i}])
	\end{split}
\end{equation} 

The non-local similarity-based tensor decomposition methods focus on removing Gaussian noise from corrupted HS images and unavoidably cause a computational burden in practice.

$\textbf{(2) Spatial and spectral smoothness }$

HS images are usually captured by airborne or space-borne platforms far from the Earth's surface. The low measurement accuracy of imaging spectrometers leads to low spatial resolutions of HS images. In general, the distribution of ground objects varies gently. Moreover, high correlations exist between different spectral bands. HS images always have relatively smoothing characteristics in the spatial and spectral domains. 

\begin{figure}[htb]
	\begin{center}
		\includegraphics[height=2.5cm]{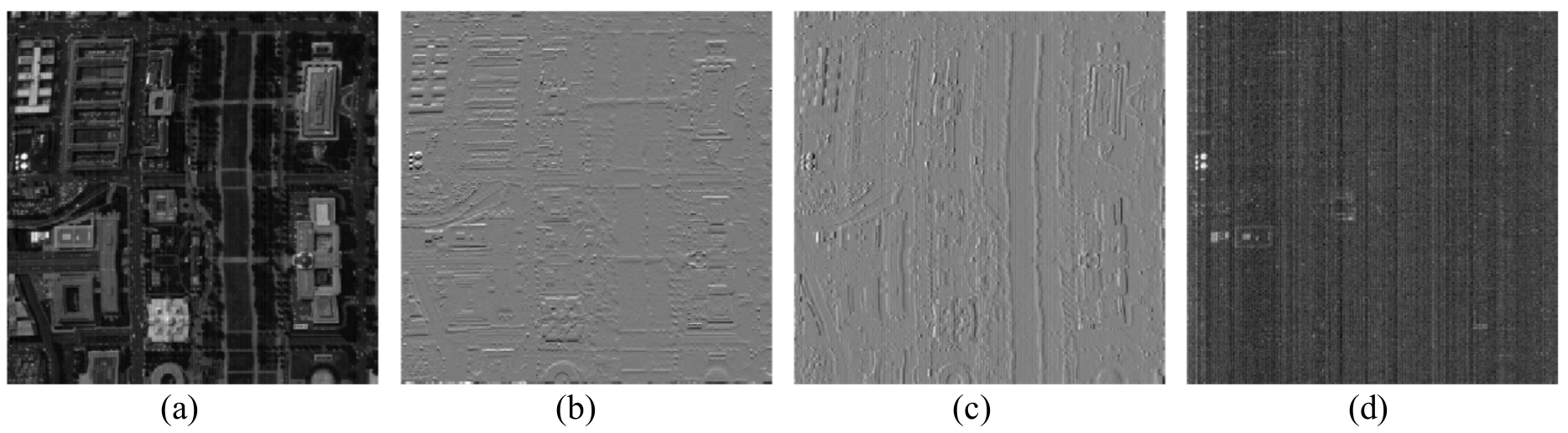}
	\end{center}
	\caption[washington]{The spatial smooth properties of Washington DC: (a) original band, (b) the gradient image along the spatial horizontal direction, (c) the gradient image along the spatial vertical direction, (d) the gradient image along the spectral direction. }
	\label{fig:sstv1}
\end{figure}

An original TV method was first proposed by Rudin \textit{et al}. \cite{c76} to remove the noise of gray-level images due to the ability to preserve edge information and promote piecewise smoothness. The HS image smoothness can be constrained by either an isotropic TV norm or an anisotropic TV norm \cite{c35}. The obvious blurring artifacts are hardly eliminated in the denoised results of the isotropic model \cite{c75}. Thus, anisotropic TV norms for HS image denoising are investigated in this paper.
We take the Washington DC (WDC) data set as a typical example to depict the gradient images along three directions in Fig. \ref{fig:sstv1}. The smoothing areas and edge information of gradient images are much clearer than the origin. 

Inspired by the TV applications to gray-level images, the 2-D spatial TV norm of $\mathcal{X}$ is easily introduced to an HS image in a band-by-band manner \cite{c30}. This simple band-by-band TV norm is defined as follows:
  \begin{equation}
	\begin{split}
		\label{eq:TV}
		|| \mathcal{X} ||_{\rm TV} = ||D_h \mathcal{X} ||_1 + ||D_v \mathcal{X} ||_1
			\end{split}
\end{equation} 
where $D_h$ and $D_v$ stand for first-order linear difference operators corresponding to the horizontal and vertical directions, respectively. These two operator are usually defined as:
  \begin{equation}
	\begin{split}
	\label{eq:dh}
||D_h \mathcal{X} ||_1 =\left\{
\begin{aligned}
\mathcal{X}(i,j+1,k)-\mathcal{X}(i,j,k) ,\quad &  \quad 1 \leq j < v \\
0, \quad &\quad j=v
\end{aligned}
\right.
\end{split}
\end{equation}

  \begin{equation}
	\begin{split}
	\label{eq:dv}
	||D_v \mathcal{X} ||_1 =\left\{
\begin{aligned}
\mathcal{X}(i+1,j,k)-\mathcal{X}(i,j,k) ,\quad &  \quad 1 \leq i < h \\
0, \quad &\quad i=h
\end{aligned}
\right.
\end{split}
\end{equation}

To enforce the spatial piecewise smoothness and the spectral consistency of HS images, a 3DTV norm \cite{c35} and a SSTV norm \cite{c19} are formulated, respectively:
  \begin{equation}
	\begin{split}
		\label{eq:3DTV}
		|| \mathcal{X} ||_{\rm 3DTV} = ||D_h \mathcal{X} ||_1 + ||D_v \mathcal{X} ||_1 + ||D_z \mathcal{X} ||_1
			\end{split}
\end{equation} 
  \begin{equation}
	\begin{split}
		\label{eq:SSTV}
||\mathcal{X}||_{\rm{SSTV}} = ||D_z(D_h \mathcal{X}) ||_1 +||D_z(D_v \mathcal{X}) ||_1
			\end{split}
\end{equation} 
where $||D_z \mathcal{X} ||_1$ is a 1-D finite-difference operator along the spectral direction and is defined as:
  \begin{equation}
	\begin{split}
	\label{eq:dz}
	||D_z \mathcal{X} ||_1 =\left\{
\begin{aligned}
\mathcal{X}(i,j,k+1)-\mathcal{X}(i,j,k) ,\quad &  \quad 1 \leq i < h \\
0, \quad &\quad k=z
\end{aligned}
\right.
\end{split}
\end{equation}

Considering the degrade model with mixed noise, Chen \textit{et al}. \cite{c77} integrated both the 2DTV and the 3DTV regularizations into the TNN. Fan \textit{et al}. \cite{c34} injected the above SSTV norm into LR tensor factorization. Wang \textit{et al}. \cite{wang2020hyperspectral} used an SSTV term in a multi-directional weighted LR tensor framework. Based on the different contributions of the three gradient terms to the 3DTV regularization, Wang \textit{et al}. \cite{c36} proposed the TV-regularized LR tensor decomposition (LRTDTV) method:
  \begin{equation}
	\begin{split}
	\label{eq:LRTDTV}
&\min _{\mathcal{X}, \mathcal{S}, \mathcal{N}} \tau\|\mathcal{X}\|_{\mathrm{3DwTV}}+\lambda\|\mathcal{S}\|_{1}+\beta\|\mathcal{N}\|_{F}^{2} \\
&\text { s.t. } \mathcal{T}=\mathcal{X}+\mathcal{S}+\mathcal{N} \\
&\mathcal{X}=\mathcal{A} \times_{1} \mathbf{B}_{1} \times_{2} \mathbf{B}_{2} \times_{3} \mathbf{B}_{3}, \mathbf{B}_{i}^{T} \mathbf{B}_{i}=\mathbf{I}(i=1,2,3)
\end{split}
\end{equation}
where the 3DwTV term is defined as:
  \begin{equation}
	\begin{split}
		\label{eq:3DwTV}
		|| \mathcal{X} ||_{\rm 3DwTV} = w_1||D_h \mathcal{X} ||_1 + w_2||D_v \mathcal{X} ||_1 + w_3||D_z \mathcal{X} ||_1
			\end{split}
\end{equation} 

Zeng \textit{et al}. \cite{c84} integrating the advantages of both a global $L_{1\--2}{\rm SSTV}$ and the local-patch TNN.
Chen \textit{et al}. \cite{c78} exploited the row sparse structure of gradient images and proposed a weighted group sparsity-regularized TV combined with LR Tucker decomposition (LRTDGS) for HS mixed noise removal. 

Due to mentioned TV norms just penalizing large gradient magnitudes and easily blurring real image edges, a new $l_0$ gradient minimization was proposed to sharpen image edges \cite{c37}. Actually, $l_1$ TV norm is a relaxation form of the $l_0$ gradient. Xiong \textit{et al}. \cite{c81} and Wang \textit{et al}. \cite{8920965} applied the $l_0$ gradient constraint in an LR BT decomposition and Tucker decomposition, respectively. However, the degrees of smoothness of this $l_0$ gradient form are controlled by a parameter, without any physical meaning. To alleviate this limitation, Ono \cite{c41} proposed a novel $l_0$ gradient projection, which directly adopts a parameter to represent the smoothing degree of the output image. Wang \textit{et al}. \cite{c82} extended the ${l_0}\text{TV}$ model into an LR tensor framework (TLR-${l_0}\text{TV}$) to preserve more information for classification tasks after HS image denoising. The optimization model of TLR-${l_0}\text{TV}$ is formulated as:
\begin{equation}
\label{eq:mlr-l0htv}
\begin{aligned}
&\underset{\mathcal{X},\mathcal{S}}{\textrm{min}}\; \sum_{k=1}^{m} \alpha _k   E_k(\mathcal{X})_{\omega} +\lambda\|\mathcal{S}\|_1 +\mu \|\mathcal{T}-\mathcal{X}-\mathcal{S}\|_F^2,\\
&s.t. \  ||{B} D \mathcal{X} ||^{\theta}_{1,0} \leq \gamma,
\end{aligned}	
\end{equation}
where the functions $E_k(\mathcal{X})_{\omega}$ are set to be $||\textbf{X}_{(k)}||_{\omega,*}$ in the WSWNN-$l_0$TV-based method and $||\mathcal{X}^k||_{\omega,*}$ in the WSWTNN-$l_0$TV-based method. Operator ${B}$ forces boundary values of gradients to be zero when $ i = h $ and $j = v $. Operator $D$ is an operator to calculate both horizontal and vertical differences. Compared with many other TV-based LR tensor decompositions, TLR-${l_0}\text{TV}$ achieves better denoising performances for mixed noise removal of HS images. In particular, HS classification accuracy is improved more effectively after denoising by TLR-l0TV.

$\textbf{(3) Subspace representation}$

\begin{figure}[htb]
	\begin{center}
		\includegraphics[height=4cm]{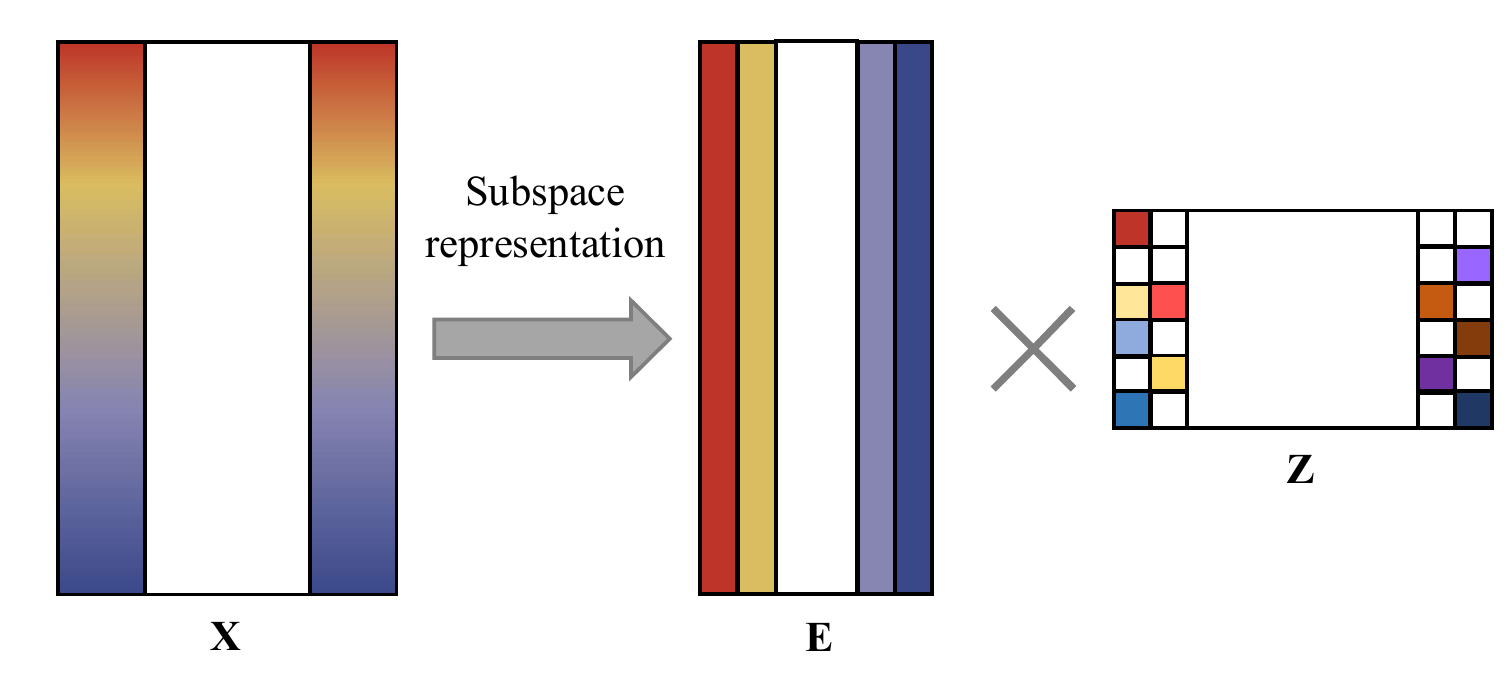}
	\end{center}
	\caption[nonlocal]{A schematic diagram of the subspace representation. }
	\label{fig:subspace}
\end{figure}

As Fig. \ref{fig:subspace} shows, an unfolding matrix $\mathbf{X}$ of a denoised HS image can be projected into a orthogonal subspace, i.e., $\mathbf{X} = \mathbf{E}\mathbf{Z}$. $\mathbf{E} \in \mathbb{R}^{z \times l}$ represents the basis of the subspace $S_l$ and $\mathbf{Z} \in \mathbb{R}^{l \times hv}$ denotes the representation coefficient of $\mathbf{X}$ with respect to $\mathbf{E}$. $\mathbf{E}$ is reasonably assumed to be orthogonal, i.e., $\mathbf{E}^{T} \mathbf{E} =\mathbf{I}$ \cite{6736073}. 

Cao \textit{et al}. \cite{c72} combined a LR and sparse factorization with the non-local tensor constraint of subspace coefficients, dubbed SNLRSF. Each spectral band of an observed HS image $\mathcal{T} \in \mathbb{R}^{h \times v \times z}$ is reshaped as each row of an HS unfolding matrix $\mathbf{T} \in \mathbb{R}^{z \times hv}$. The spectral vectors is assumed to lie in a $l$-dimensional subspace $S_l$ ($l \ll z$), and the optimization model can be written as
\begin{equation}
\label{eq:SNLRSF}
\begin{aligned}
&\underset{\mathbf{E}, \mathbf{Z}, \mathcal{L}_{i}, \mathbf{S}}{\arg \min } \frac{1}{2}\|\mathbf{T}-\mathbf{E Z}-\mathbf{S}\|_{F}^{2}+\lambda_{2}\|\mathbf{S}\|_{1} \\
&+\lambda_{1} \sum_{i}\left(\frac{1}{\delta_{i}^{2}}\left\|\Re_{i} \mathbf{Z}-\mathcal{L}_{i}\right\|_{F}^{2}+ ||\mathcal{L}_i ||_{\rm TTN} \right) \quad \text { s.t. } \quad \mathbf{E}^{T} \mathbf{E}=\mathbf{I}
\end{aligned}	
\end{equation}
where $\Re_{i} \mathbf{Z}$ is divided into three steps: 1) reshape the reduced-dimensionality coefficient image $\mathbf{Z} \in \mathbb{R}^{l \times hv}$ as a tensor $\mathbf{Z} \in \mathbb{R}^{h \times v \times l}$; 2) segment the tensor $\mathbf{Z}$ as an overlapped patch tensor $\mathbf{Z}_i \in \mathbb{R}^{p \times p \times l}$; and 3) cluster $d$ similar patches in a neighborhood area by computing Euclidean distance.

From one side, a spectral LR tensor model is explored according to the fact that spectral signatures of HS images lie in a low-dimensional subspace. From another side, a non-local LR factorization is employed to take the non-local  similarity along the spatial direction into consideration. Following the line of SNLRSF, Zheng \textit{et al}. \cite{c86}  employed LR matrix factorization to decouple spatial and spectral models. The group-sparse structure of HS images is introduced on spatial difference images (SpatDIs). A continuity constraint was applied in the spectral factor to promote the group sparsity of SpatDIs and the spectral continuity of HS images. Sun \textit{et al}. \cite{c88} projected the noisy HS images into a non-local tensor subspace spanned by a spectral difference continuous basis. The continuity of the restored HS data is significantly promoted by this difference regularization.

\subsubsection{Experimental results and analysis}
\label{sect:experiment_denoising}
\begin{figure*}[htb]
	\begin{center}
		\includegraphics[height=3.3cm]{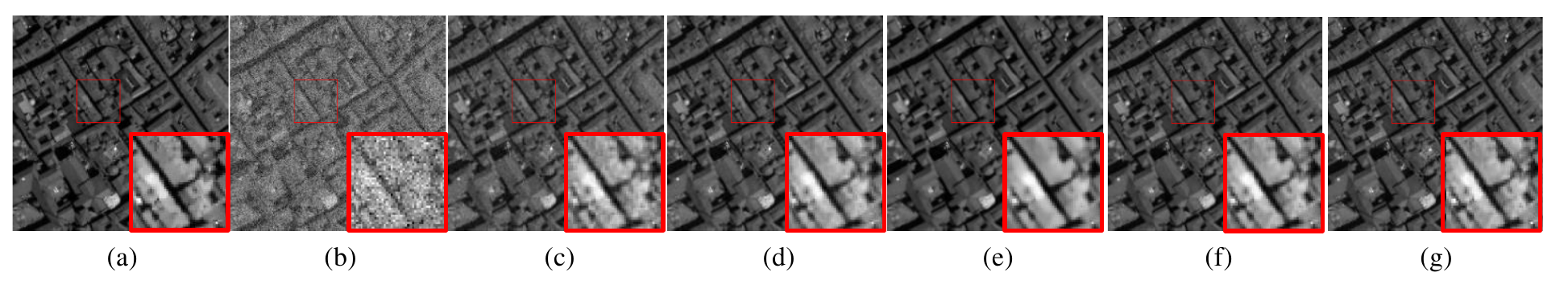}
	\end{center}
	\caption[pavia80]{The different methods for Gaussian noise removal. (a) Original HS image, (b) Gaussian noise, (c) LRTA, (d) TDL, (e) ITS, (f) LLRT, (g) NGmeet.  }
	\label{fig:Visio-denoising_gaussian}
\end{figure*}

\begin{figure*}[htb]
	\begin{center}
		\includegraphics[height=3.5cm]{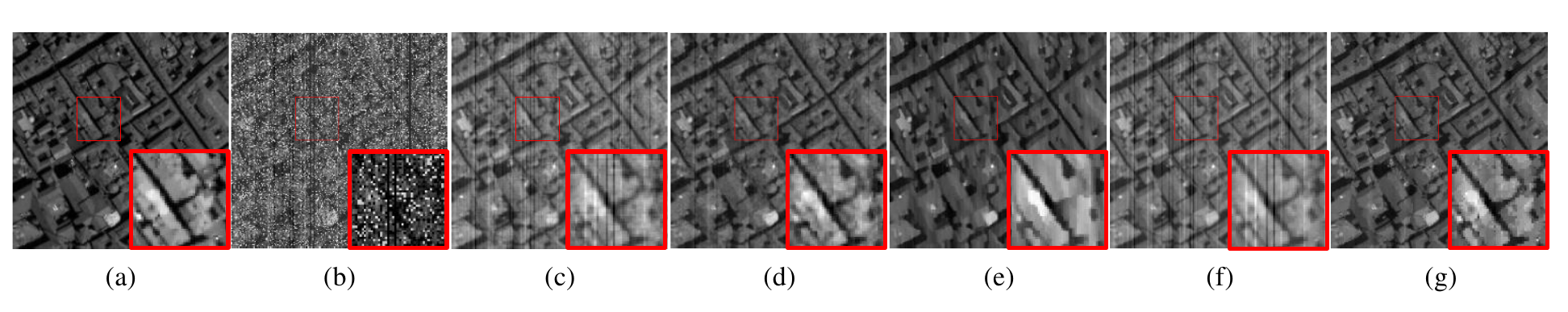}
	\end{center}
	\caption[pavia80]{The different methods for mixed noise removal. (a) Original HS image, (b) Mixed noise, (c) LRTR, (d) LRTDTV, (e) LRTDGS, (f) 3DTNN, (g) TLR-$L_0$TV  }
	\label{fig:Visio-denoising_mixed}
\end{figure*}

An HS subimage is selected from the Pavia University data set and is normalized to $[0,1]$. The zero-mean Gaussian noise of noise variance $0.12$ is added into each band and shown in Fig. \ref{fig:Visio-denoising_gaussian} (b). In a mixed noise case, the same Gaussian noise is also adopted. Each band is corrupted by the salt and pepper noise with a proportion of $0-20\%$. Dead-lines are randomly added from band $61$ to band $80$, with the width of stripes generated from $1$ to $3$, and the number of stripes randomly selected from $3$ to $10$. In addition, bands $61-70$ are corrupted by some stripes with the number randomly selected from $20$ to $40$. Four different quantitative quality indices are chosen: the mean of peak signal-to-noise ratio (MPSNR), the mean of structural similarity (MSSIM), relative dimensional global error in synthesis (ERGAS), and the mean spectral angle distance (MSAD). Larger MPSNR and MSSIM values indicate better-denoised image quality. These two indices pay attention to the restoration precision of spatial pixels. In contrast, smaller ERGAS and MSAD values illustrate better performances of denoised results. 

For Gaussian noise removal, all the competing approaches achieve good results to some degree in Fig. \ref{fig:Visio-denoising_gaussian}, in which the enlarged subregions are delineated in red boxes. But residual noise remains in the result denoised by LRTA. Compared with TDL, ITS fails to preserve detailed spatial information. LLRT provides a rather similar result with NGmeet. Consistent with the visual observation, NGmeet outperforms the other methods and obtains the highest metric values among the denoising models in Tab. \ref{tab:tab-gaussiannoise}. The non-local LR tensor methods including ITSReg, TDL, and LLRT gain better performances than LRTA, due to the formers exploiting two types of HS prior knowledge. The LRTA method is the fastest one among all the competing algorithms since LRTA just considers the spectral correlation. 

Fig. \ref{fig:Visio-denoising_mixed} shows the restoration results by five different methods under a heavy noise case. Dead-lines remaining in the images denoised by LRTR and 3DTNN are more obvious than the ones restored by LRTDTV and LRTDGS. The LR tensor-based model is employed in LRTR and 3DTNN, yet LRTDTV, LRTDGS, and TLR-$L_0$TV considered two kinds of prior knowledge: spectral correlation and spatial-spectral smoothness. LRTDTV and LRTDGS are more sensitive to dead-lines than TLR-$L_0$TV, leading to more or fewer artifacts in the denoised results. TLR-$L_0$TV removes most of the mixed noise and preserves image details like texture information and edges. To further evaluate the differences among competing denoising methods, we calculate four quality indices and show them in Tab. \ref{tab:tab-mixednoise}, with the best results in bold. TLR-$L_0$TV obtains the highest denoising performance among all the approaches. For MPSNR, LRTDTV and LRTDGS are slightly larger than 3DTNN, whereas the SSIM and ERGAS values of LRTDTV and LRTDGS are better than those of 3DTNN. LRTR and LRTDGS are the first and second faster, but they hardly handle the complex mixed noise case with some dead-lines retaining.

\begin{table}[!htbp]
\centering
\caption{Quantitative comparison of different selected algorithms for Gaussian noise removal.}
\begin{spacing}{1.0}
\scalebox{0.93}{
\begin{tabular}{c|ccccc}
\cline{1-6}
&  \multicolumn{5}{|c}{Gaussian noise removal} \\
\hline
Index &  LRTA & TDL & ITS & LLRT & NGmeet \\
\hline
PSNR    & 32.14  & 34.54  & 34.38  & \underline{35.96}  &  $\mathbf{37.06}$ \\
 
SSIM    & 0.9097 & {0.9484} & 0.9466 & \underline{0.9637} &  $\mathbf{0.9707}$\\
 
ERGAS   & 5.7044 & 4.3392 & 4.3981 & \underline{4.0462} &  $\mathbf{3.2344}$\\
 
MSAD    & 6.6720 & 5.0701 & 5.0912 & \underline{4.2402} &  $\mathbf{3.7804}$ \\
 
TIME(s) & $ \mathbf{1.48} $  & \underline{13.77}   & 650.49 & 506.84 & 29.58   \\
\hline
\end{tabular}
}
\label{tab:tab-gaussiannoise}
\end{spacing}
\end{table}

  \begin{table}[!htbp]
\centering
\caption{Quantitative comparison of different selected algorithms for mixed noise removal.}
\begin{spacing}{1.0}
\scalebox{0.8}{
\begin{tabular}{c|ccccc}
\cline{1-6}
&  \multicolumn{5}{|c}{Mixed noise removal} \\
\hline
Index & LRTR & LRTDTV & LRTDGS & 3DTNN & TLR-$L_0$TV\\
\hline
MPSNR  & 26.89   & \underline{30.76}   & \underline{30.76}   & 30.20  & $\mathbf{31.59}$\\
 
MSSIM  & 0.8157  & 0.8821  & 0.7852 & \underline{0.8945} &  $\mathbf{0.8973}$ \\
 
ERGAS & 10.8842 & 7.8154  & 9.5527 & \underline{7.4915} &  $\mathbf{7.1748}$\\
 
MSAD  & 9.9624  & \underline{7.3689}  & 10.4568 & 7.4712 &  $\mathbf{7.2055}$\\
 
TIME(s) & $\mathbf{19.67}$ & 35.29   & \underline{25.11}  & 44.25 & 325.67 \\
\hline
\end{tabular}
}
\label{tab:tab-mixednoise}
\end{spacing}
\end{table}

\begin{figure*}[htb]
	\begin{center}
		\includegraphics[height=3.1cm]{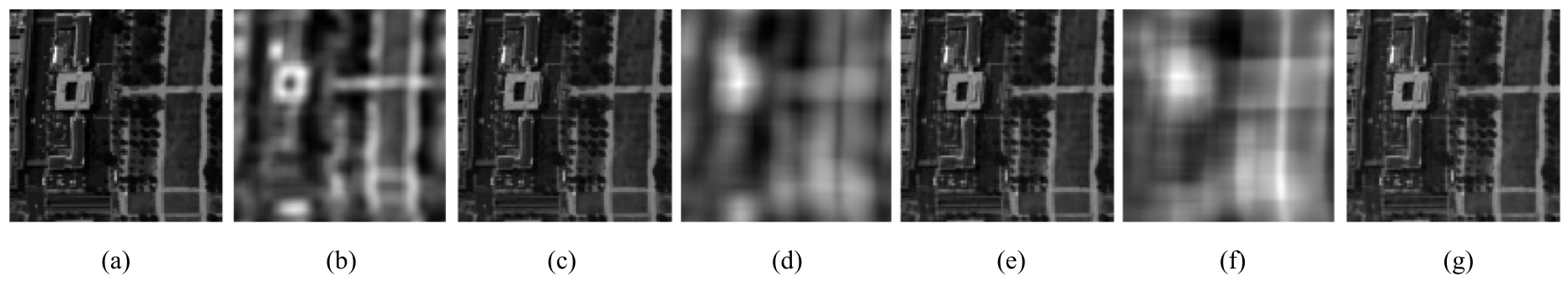}
	\end{center}
	\caption[wdc]{ OLRT for different blur cases. (a) Original WDC image, (b) the light Gaussian blur on WDC ($8 \times 8$, Sigma = 3) and corresponding deblurred image (c), (d) the heavy Gaussian blur on WDC ($17 \times 17$, Sigma = 7) and corresponding deblurred image (e), (f) the Uniform blur and corresponding deblurred image (g). }
	\label{fig:Visio-deblur}
\end{figure*}

\begin{table*}[!htbp]
\centering
\caption{A quantitative evaluation of OLRT for different blur cases.}
\begin{spacing}{1.0}
\scalebox{0.93}{
\begin{tabular}{c|ccccc}
\cline{1-6}
Blur cases &  MPSNR & MSSIM & ERGAS & MSAD & TIME(s) \\
\hline
Gaussian blur (8*8, Sigma = 3) & 43.50 & 0.9912 & 1.5910 & 1.9486 &314.50 \\
Gaussian blur (17*17, Sigma = 7) & 39.63 & 0.9807 & 2.5407 & 3.0819 &305.70 \\
 Uniform blur & 39.39 & 0.9784 & 2.9332 & 3.8355 & 314.28 \\
   
\hline
\end{tabular}
}
\label{tab:tab-deblur}
\end{spacing}
\end{table*}

\subsection{HS Deblurrring} 
\label{sect:HS_Deblurring}

The atmospheric turbulence or fundamental deviation of some imaging systems often blur HS images during the data acquisition process, which unfortunately damages the high-frequency components and the edge features of HS images. HS deblurring aims to recover sharp latent images from blurred ones. Chang \textit{et al}. \cite{c117} discussed the LR correlations along HS spatial, spectral, and non-local similarity modes and proposed a unified optimal LR tensor (OLRT) framework for multiple HS restoration tasks. But a matrix nuclear norm is used to constrain the LR property of unfolding non-local patch groups. Consequently, Chang \textit{et al}. \cite{c116} proposed a weighted LR tensor recovery (WLRTR) algorithm with a reweighted strategy. Considering spectral correlation and non-local similarity, the HS deblurring optimization problem can be formulated as follows
\begin{equation}
    \begin{aligned}
    \label{eq:WLRTR}
 &\underset{\mathcal{X}, \mathcal{A}_{i}, \mathbf{B}_{j}}{ \min }   \frac{1}{2}\|\mathcal{T}-M( \mathcal{X} ) \|_{F}^{2}+\\
 &\eta \sum_{i}\left(\left\|\mathcal{R}_{i} \mathcal{X}-\mathcal{A}_{i} \times_{1} \mathbf{B}_{1} \times_{2} \mathbf{B}_{2} \times_{3} \mathbf{B}_{3}\right\|_{F}^{2}\right.\left.+\sigma_{i}^{2}\left\|\boldsymbol{w}_{i} \circ \mathcal{A}_{i}\right\|_{1}\right) 
\end{aligned}
\end{equation}
where $w_i$ is a reweighting factor inversely proportional to singular values of $\mathcal{L}_i$ with $\mathcal{L}_i = \mathcal{A}_{i} \times_{1} \mathbf{B}_{1} \times_{2} \mathbf{B}_{2} \times_{3} \mathbf{B}_{3}$, and higher-order SVD (HOSVD) is applied to see the different sparsities of higher-order singular values, i.e., LR property. The last term $\|\mathcal{Y}-M( \mathcal{X} ) \|_{F}^{2}$ is a data fidelity item, which can be replaced by $ || \mathcal{T} - M(\mathcal{X}) - \mathcal{S} - \mathcal{N} ||^2_F$ for HS inpainting, destriping, and denoising problems.

An experimental example is given to display the deblurred performances of OLRT for the Gaussian blur with different levels and the uniform blur on the WDC data set. Fig. \ref{fig:Visio-deblur} shows the visual results under different blur cases. The specific texture information is hardly distinguished in the three blurred images shown in Fig. \ref{fig:Visio-deblur} (b), (d), and (f). The optimal LR tensor prior knowledge of OLRT reliably reflects the intrinsic structural correlation of HS images, which benefits the recovery of structural information and image edges. The quantitative results under different blur cases are reported in Tab. \ref{tab:tab-deblur}.

\subsection{HS Inpainting}
\label{sect:HS_Inpainting}

\begin{figure}[htb]
	\begin{center}
		\includegraphics[height=3.2cm]{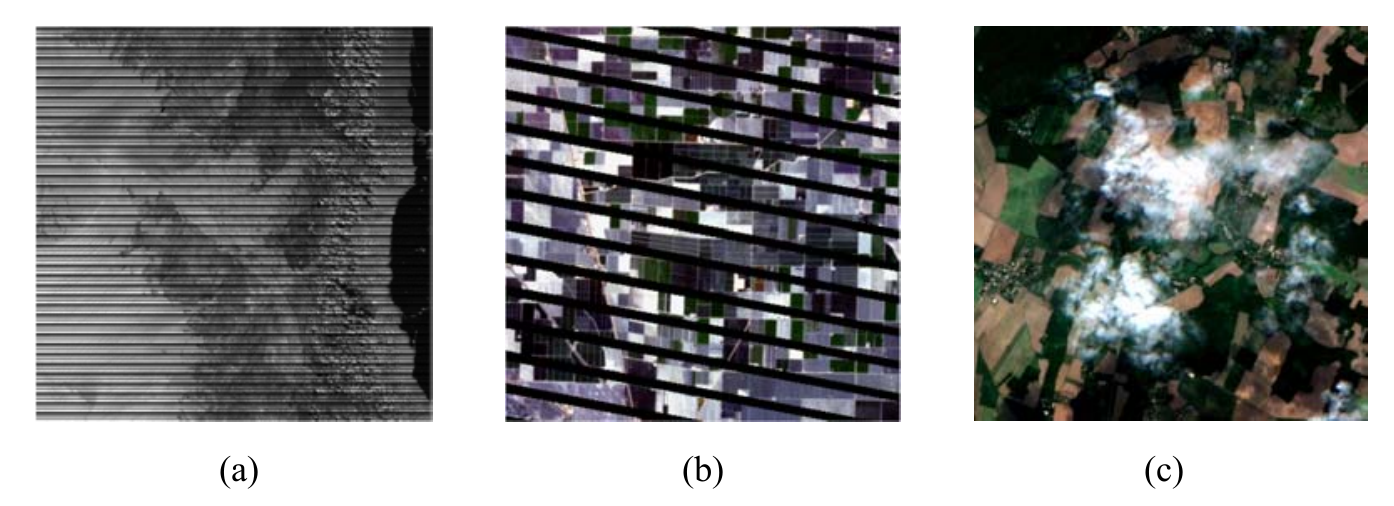}
	\end{center}
	\caption[md]{ Examples of RS data with missing information. (a) Reflectance of Aqua MODIS band 6 with sensor failure. (b) Digital number values of Landsat ETM+ with the SLC-off problem. (c) Digital number
values of a Landsat image with cloud obscuration. }
	\label{fig:Visio-missingdata}
\end{figure}

In this section, we introduce and discuss LR tensor-based methods for HS inpainting. These methods are also suitable for missing data recovery of high-dimensional RS (HDRS) images. 
RS images such as HS, MS, and multi-temporal images often from missing data problems, such as dead pixels, thick clouds, and cloud shadows, as shown in Fig. \ref{fig:Visio-missingdata}. The goal of inpainting is to estimate the missing data from observed images, which can be regarded as a tensor completion problem. 

LR tensor completion theory has been successfully applied for HS inpainting \cite{c117,c116,c118,c119,c120,c130,c131}. Liu \textit{et al}. \cite{c119} suggested a trace norm regularized CP decomposition for missing data recovery. Ng \textit{et al}. \cite{c121} learned from high-accuracy LR tensor completion (HaLRTC) \cite{c120} for recovering the missing data of HDRS and proposed an adaptive weighted TC (AWTC) method. The proposed AWTC model is expressed as
\begin{equation}
    \begin{aligned}
       \label{eq:AWTC}
       \underset{\mathcal{X}}{ \min } \frac{\eta}{2} ||  \mathcal{T} - M(\mathcal{X})   ||^2_F + \sum_{i=1}^3 w_i || \mathbf{X}_{(i)}||_*
\end{aligned}
\end{equation}    
where $w_i$ is well-designed parameter related to the singular values of $\mathbf{X}_{(i)}$.
Xie \textit{et al}. \cite{c130} proposed a LR regularization-based TC (LRRTC), fusing the logarithm of the determinant with a TTN. With the definitions of a new TNN and its t-SVD \cite{c123},  Wang \textit{et al}. \cite{c124} and Srindhuna \textit{et al}. \cite{c125} proposed new low-tuba-rank TC methods to estimate the missing values in HDRS images. Consequently, a novel TR decomposition is formulated to represent a high-dimensional tensor by circular multi-linear products on a sequence of third-order tensors \cite{c126}. Based on the TR theory, He \textit{et al}. \cite{c128} fused the spatial TV into the TR framework and developed two solving algorithms: ALM and ALS. Similarly, Wang \textit{et al}. \cite{c129} incorporated a 3DTV regularization into a novel weighted TR decomposition framework. The proposed TV-WTV model is formulated as:
\begin{equation}
\label{eq:TVWTR}
\begin{aligned}
&\underset{\mathcal{X},[\mathcal{G}]}{\textrm{min}}\; \sum^N_{n=1}\sum^3_{i=1} \theta_i ||\textbf{G}^{(n)}_{(i)}||_*+ \frac{\lambda}{2}||\mathcal{X} - \Phi([\mathcal{G}])||^2_F+\tau||\mathcal{X}||_{\rm 3DTV} \\
&s.t.\; \mathcal{X}_{\Omega}=\mathcal{T}_{\Omega}
\end{aligned}
\end{equation}

\begin{figure*}[htb]
	\begin{center}
		\includegraphics[height=6cm]{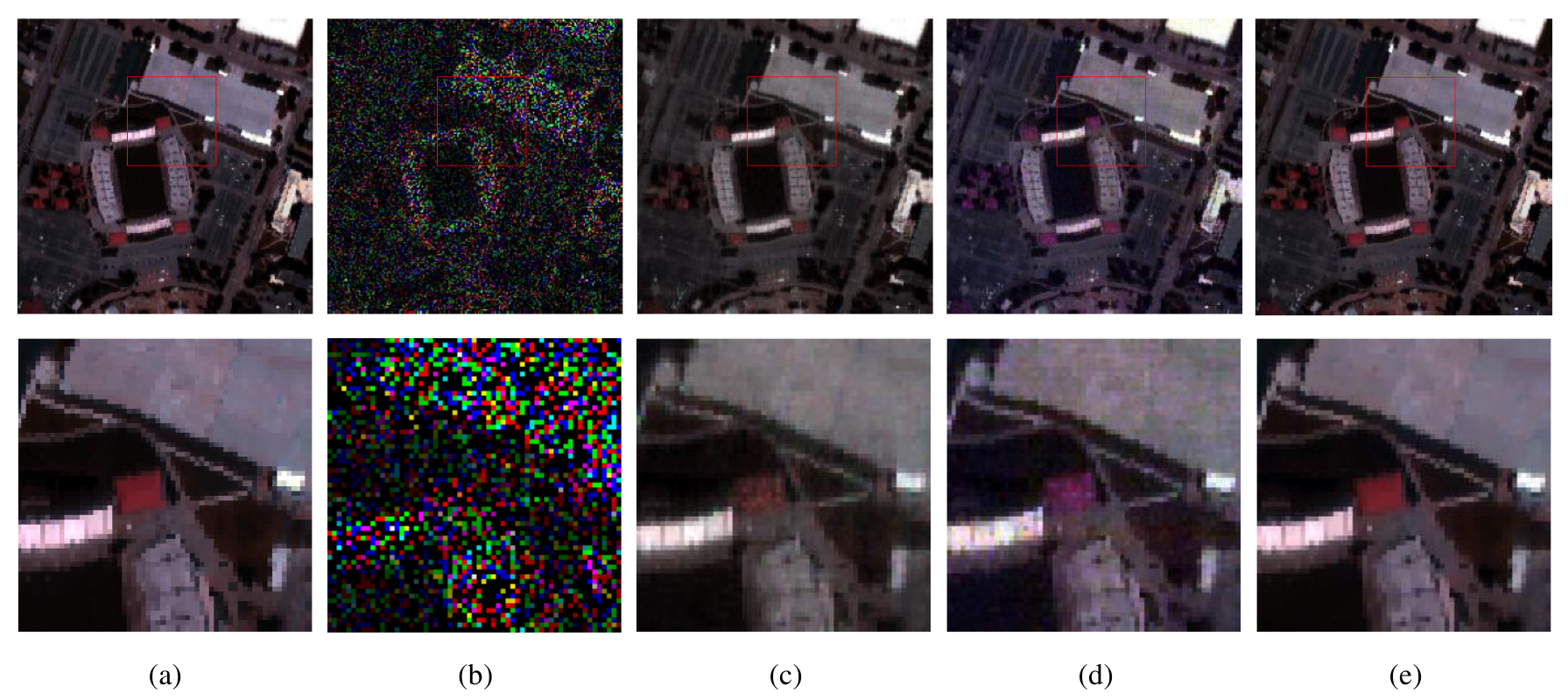}
	\end{center}
	\caption[houston]{ The inpainting results by different methods under $80\%$ missing ratio. (a) Original Houston2013 image, (b) Missing, (c)  HaLRTC, (d) LRTC, (e) TVWTR. }
	\label{fig:Visio-inpainting1}
\end{figure*}

\begin{figure*}[htb]
	\begin{center}
		\includegraphics[height=6cm]{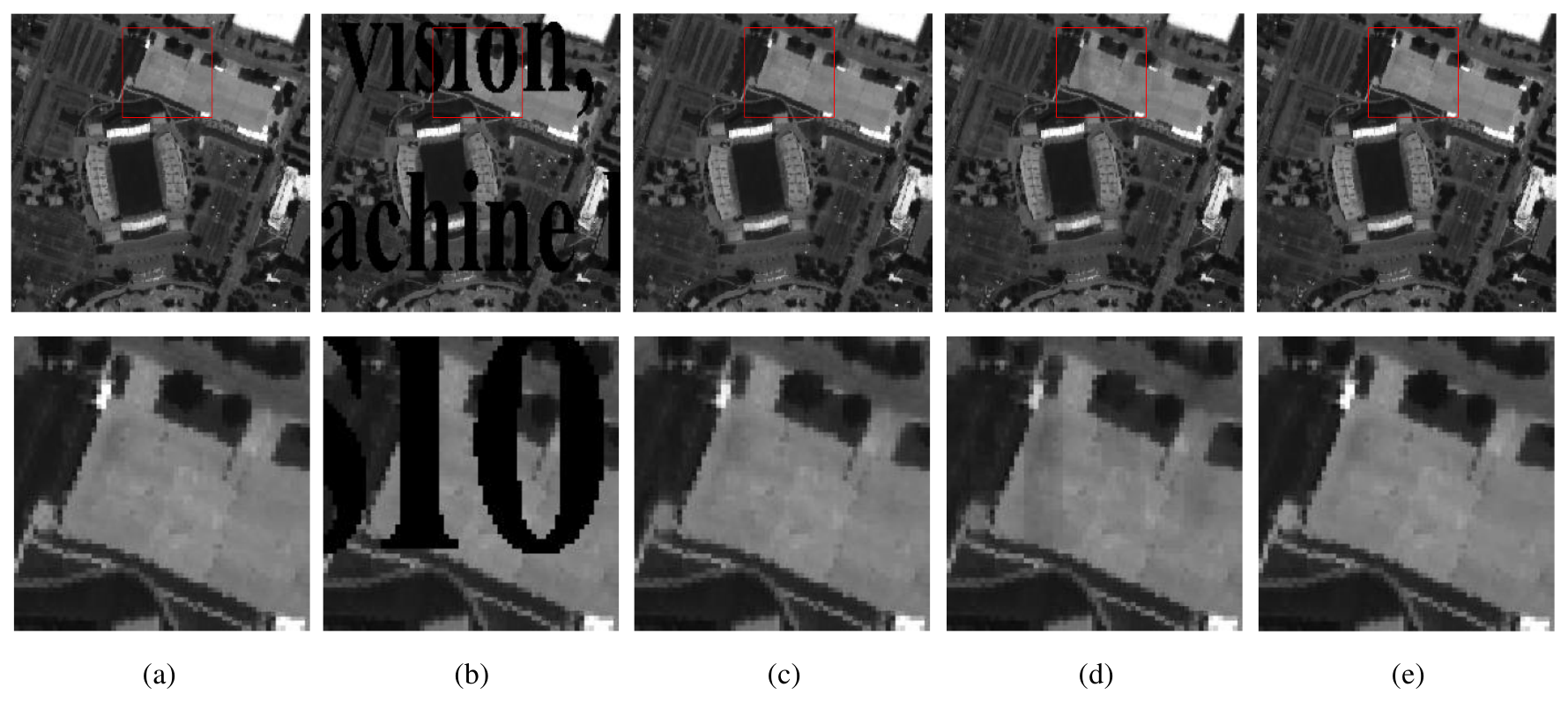}
	\end{center}
	\caption[houston]{ Inpainting results by different methods for the text removal case. (a) Original Houston2013 image, (b) Missing, (c)  HaLRTC, (d) LRTC, (e) TVWTR. }
	\label{fig:Visio-inpainting2}
\end{figure*}

For HS image inpainting tasks, we test three methods: HaLRTC, LRTC, and TVWTR on the random missing data problem and the  text removal problem. A subimage is chosen from the Houston 2013 data set for our experimental study. Fig. \ref{fig:Visio-inpainting1} shows the results of the Houston2013 data set before and after recovery under ratio = $80\%$. Although missing pixels disappear in the results of HaLRTC and LRTC, these methods produce more or fewer artifacts in the top-right corner of the zoomed area. The TVWTR method performs the best among all the compared algorithms and recovers the details like the red square center of the zoom area. In Fig. \ref{fig:Visio-inpainting2}, original HS bands are corrupted by different texts that do not appear randomly as in previous cases. The text corruption is eliminated by three tensor decomposition-based algorithms. Few text artifacts exist in the enlarged area of LRTC. Due to the consideration of the spectral correlation and the spatial-spectral smoothness, TVWTR provides the best result with reconstructing most information of the original image.

The corresponding quantitative results of two inpainting tasks are reported in Tab. \ref{tab:tab-inpainting}. Taking account of two types of prior knowledge, TVWTR gives a significantly fortified performance under two cases, as compared with the other competing methods. HaLRTC and LRTC are the fastest and second-fastest among all the comparing methods.

\begin{table*}[!htbp]
\centering
\caption{A quantitative evaluation of different methods for inpainting.}
\begin{spacing}{1.0}
\scalebox{0.93}{
\begin{tabular}{c|ccc|ccc}
\cline{1-7}
&  \multicolumn{3}{c}{Missing data ($80\%$)} &  \multicolumn{3}{|c}{Text removal} \\
\hline
Index & HaLRTC & LRTC & TVWTR&  HaLRTC & LRTC & TVWTR \\
\hline
 MPSNR & 36.54  & 38.49 & $\mathbf{49.05}$ & 50.29 & 53.39& $\mathbf{57.31}$ \\
 MSSIM & 0.9391 & 0.9555 & $\mathbf{0.9947}$ & 0.9965 & 0.9975 & $\mathbf{0.9991}$ \\
 ERGAS & 4.5703 & 3.7704 & $\mathbf{1.0211}$ & 1.0108 & 1.1345 & $\mathbf{0.4012}$\\ 
 MSAD  & 5.3845 & 4.6162 & $\mathbf{1.3792}$ & 1.2424 &1.1134 & $\mathbf{0.5497}$\\
 TIME(s) &$\mathbf{9.34}$ & 23.74 & 347.46 & $\mathbf{8.38}$ & 24.90 & 345.66\\
   
\hline
\end{tabular}
}
\label{tab:tab-inpainting}
\end{spacing}
\end{table*}

\subsection{HS Destriping}
\label{sect:HS_Destriping}
In the past three decades, plenty of airborne and space-borne imaging spectrometers have adopted a whiskbroom sensor or a pushbroom sensor commonly. The former one is built with linear charge-coupled device (CCD) detector arrays. The corresponding HS imaging systems scans the target pixel by pixel and then acquires a spatial image by track scanning with a scan mirror forward motion \cite{c134}. The latter one contains area CCD arrays. A pushbroom sensor scans the target line by line, one direction of which is utilized for spatial imaging, and the other for spectral imaging. The incoherence of the system mechanical motion and the failure of CCD arrays lead to the non-uniform response of neighboring detectors, mainly generating stripe noise. The  periodic or noneriodic stripes generally distributed along the scanning direction have a certain width and length. The values of stripes are brighter or darker than their surrounding pixels. The inherent property of stripes, i.e., $g(\mathcal{S})$ should be considered in the HS destriping model. 

Chen \textit{et al}. \cite{c132} were the first to develop a LR tensor decomposition for an MS image destriping task. The high correlation of the stripe component along the spatial domain is depicted by a LR Tucker decomposition. The final minimization model for solving the destriping problem is expressed as follows:
 \begin{equation}
    \begin{aligned}
\min _{\mathcal{X}, \mathcal{S}, \mathcal{A}, \mathbf{B}_{i}} & \frac{1}{2}\|\mathcal{Y}-\mathcal{X}-\mathcal{S}\|_{F}^{2}+\eta_{1}\left\|D_{h} \mathcal{X}\right\|_{1}+\eta_{2}\left\|D_{z} \mathcal{X}\right\|_{1} \\
&+\lambda\|\mathcal{S}\|_{2,1} \\
\text { s.t. } \mathcal{S}=& \mathcal{A} \times_{1} \mathbf{B}_{1} \times_{2} \mathbf{B}_{2} \times_{3} \mathbf{B}_{3}, \mathbf{B}_{i}^{T} \mathbf{B}_{i}=\mathbf{I}(i=1,2,3)
\end{aligned}
\end{equation}
where $\|\mathcal{S}\|_{2,1} = \sum^z_{k=1} \sum^v_{j=1} \sqrt{\sum^h_{i_1} \mathcal{S}_{i,j,k}^2 } $.

Cao \textit{et al}. \cite{c135} implemented the destriping task by the matrix nuclear norm of stripes and non-local similarity of image patches in the spatio-spectral volumes. WLRTR and OLRT \cite{c117,c116} are also effective for a HS destriping task. Chang \textit{et al}. \cite{c117} simultaneously considered the LR properties of the stripe cubics and non-local patches. The OLRT algorithm is reformulated for modeling both the recovered and stripe components as follows
\begin{equation}
\begin{aligned}
&  \min _{\mathcal{X}, \mathcal{L}_{i}^{j}, \mathcal{S}} \frac{1}{2}\|\mathcal{T}-\mathcal{X}-\mathcal{S}\|_{F}^{2}+\rho \operatorname{rank}_{1}(\mathcal{S}) \\
&+\omega_{j} \sum_{j} \sum_{i}(\frac{1}{\delta_{i}^{2}}\|\mathcal{R}_{i}^{j} 
\mathcal{X}-\mathcal{L}_{i}^{j}\|_{F}^{2}+\operatorname{rank}_{j} (\mathcal{L}_{i}^{j}) )
\end{aligned}
\end{equation}

In \cite{c133}, an HS destriping model is transformed to a tensor framework, in which the tensor-based non-convex sparse model used both $l_0$ and $l_1$ sparse priors to estimate stripes from noisy images.

\begin{figure*}[htb]
	\begin{center}
		\includegraphics[height=6cm]{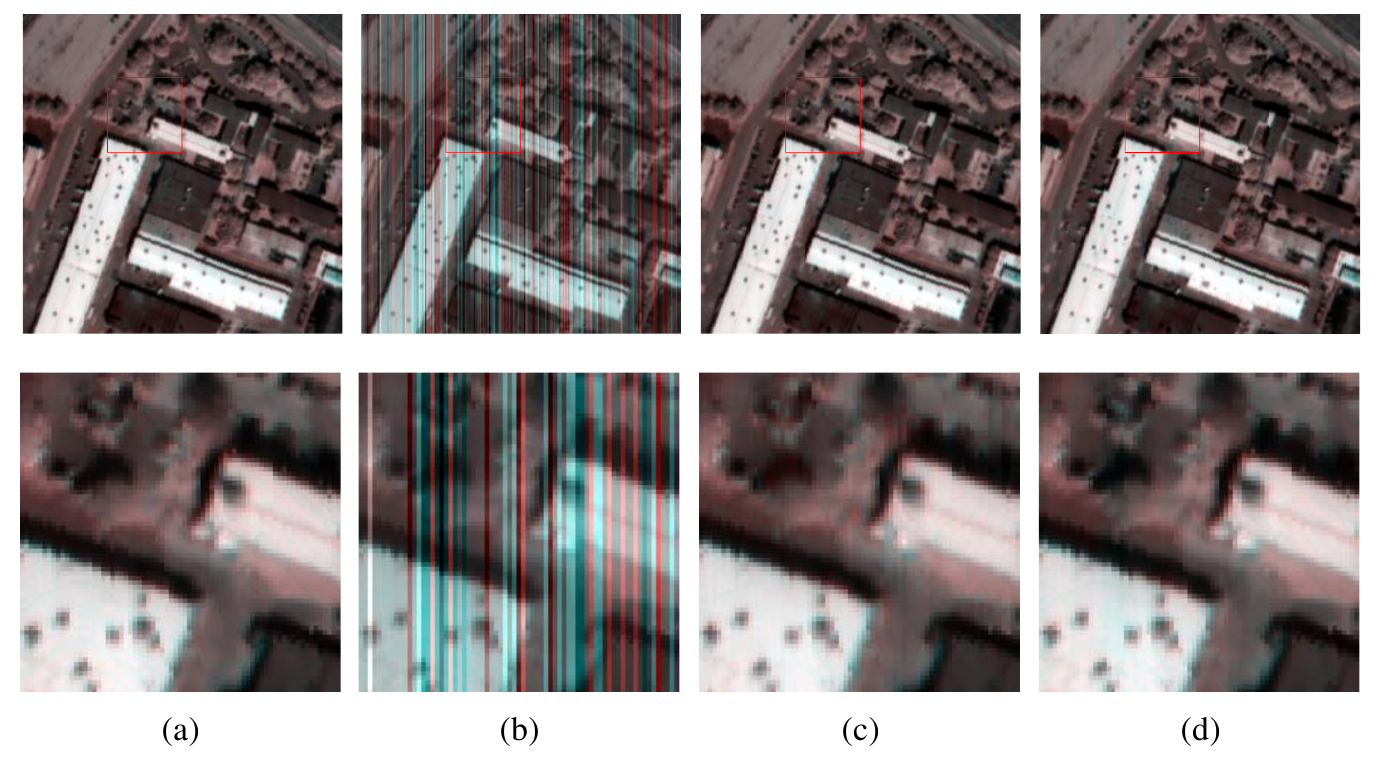}
	\end{center}
	\caption[houston2018]{The destriping results by different methods. (a) Original Houston 2018 image, (b) nonperiodic stripes, (c) WLRTR, (d) LRTD. }
	\label{fig:Visio-destriping}
\end{figure*}

We take an example with nonperiodic stripes of intensity 50 and stripe ratio 0.2, which is presented in Fig. \ref{fig:Visio-destriping} (b). Fig. \ref{fig:Visio-destriping} (c) and (d) display the destriping results of WLRTR and LRTD. The stripes are estimated and removed correctly by WLRTR and LRTD since both models consider non-local similarity and spectral correlation. Considering the third type of prior knowledge-- spatial and spectral smoothness, LRTD moderately preserves more details like clear edges than WLRTR. The quantitative comparison is in accordance with the above-mentioned visual results. Tab. \ref{tab:tab-destriping} performs destriping results with four quantitative indices. LRTD achieves higher evaluation values than WLRTR.

 \begin{table}[!htbp]
\centering
\caption{A quantitative comparison of different selected algorithms for destriping.}
\begin{spacing}{1.0}
\scalebox{0.8}{
\begin{tabular}{c|cccc}
\hline
 & MPSNR & MSSIM & ERGAS & MSAD \\
\hline
WLRTR & 39.77 & 0.9844 &7.7883 & 5.1765\\
LRTD  & $\mathbf{47.87}$ & $\mathbf{0.9912}$ & $\mathbf{3.8388}$ & $\mathbf{1.1276}$\\

\hline
\end{tabular}
}
\label{tab:tab-destriping}
\end{spacing}
\end{table}
\subsection{Future challenges}
\label{sect:challenges_restoration}

Various tensor optimization models have been developed to solve the HS restoration problem and show impressive performances. Nevertheless, these models still can be further improved for future work:

As the prior information is efficient to find the optimal solution, novel tensor-based approaches should utilize as many types of priors as possible. Therein, how best to design a unified framework to simultaneously non-local similarity, spatial and spectral smoothness, and subspace representation is a crucial challenge.

The addition of different regularizations leads to the manual adjustment of corresponding parameters. For example, a noise-adjusted parameter pre-definition strategy needs to be studied to enhance the robustness of tensor optimization models. 

 It is worth noting that we are usually blind to the location of the stripes or clouds. The locations of the stripes or mixed noise between the neighbor bands are often different and need to be estimated. How best to predict the degradation positions and design blind estimation algorithms deserves further study in following research.

Due to some HS images containing hundreds of spectral bands, the high dimensions of an HS tensor cause a time-consuming problem. The model complexity of tensorial models should be reduced with the guarantee of efficiency and accuracy of HS restoration. 

\section{HS CS}
\label{sect:HSI-CS}
Traditional HS imaging techniques are based on the Nyquist sampling theory for data acquisition. A signal must be sampled at a rate greater than twice its maximum frequency component to ensure unambiguous data \cite{c108,c109}. This signal processing needs a huge computing space and storage space. Meanwhile, the ever-increasing spectral resolution of HS images also leads to the high expense and low efficiency of transmission from airborne or space-borne platforms to ground stations. The goal of CS is to compressively sample and reconstruct signals based on sparse representation to reduce the cost of signal storage and transmission. In Fig. \ref{fig:Visio-cs}, based on the image-forming principle of a single pixel camera which uses the digital micromirror device (DMD) to accomplish the CS sampling, an HS sensor can span the necessary wavelength range and record the intensity of the light reflected by the modulator in each wavelength \cite{c215}. Since the CS rate can be far lower than the Nyquist rate, the limitation of high cost caused by the sheer volume of HS data will be alleviated. A contradiction usually exists between the massive HS data and the limited bandwidth of satellite transmission channel. HS images can be compressed first to reduce the pressure on channel transmission. Therefore, the HS CS technique is conducive to onboard burst transmission and real-time processing in RS \cite{c223}.

\begin{figure*}[htb]
	\begin{center}
		\includegraphics[height=6cm]{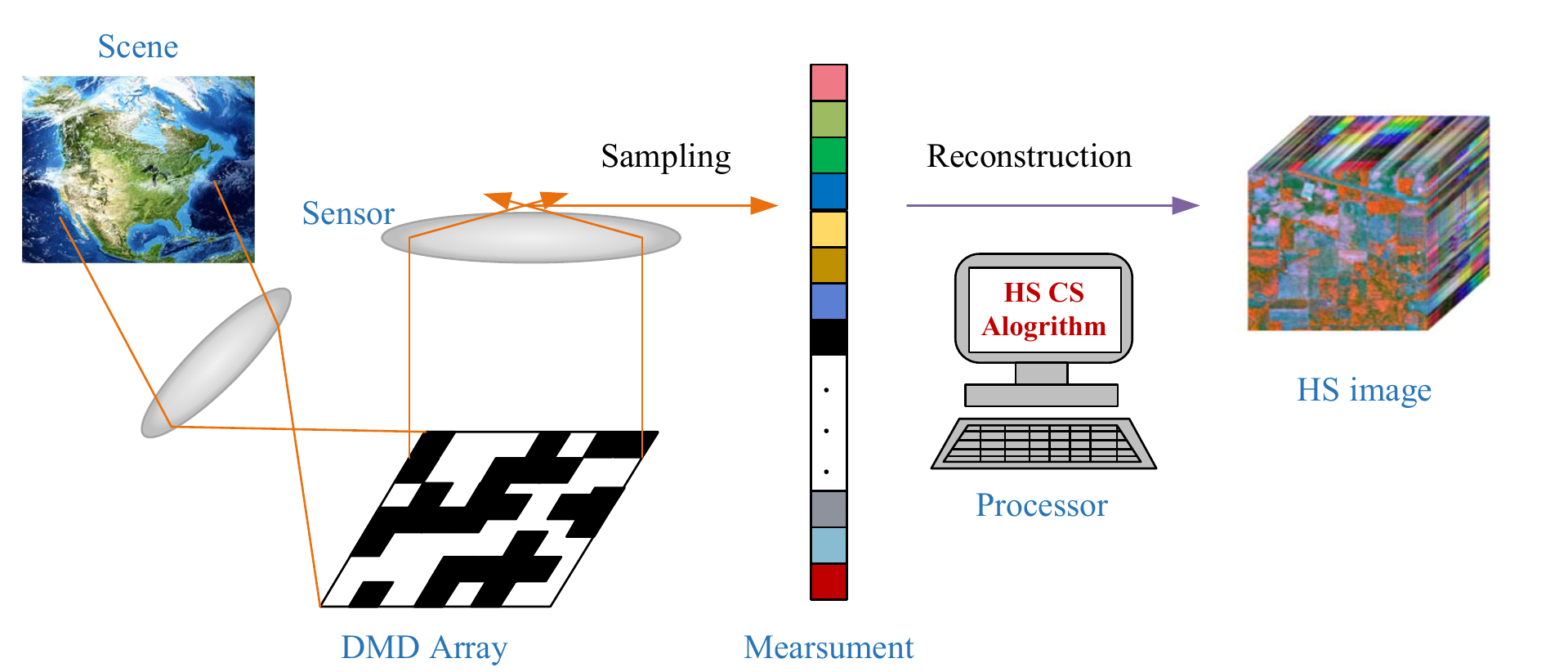}
	\end{center}
	\caption[restoration]{A schematic diagram of HS CS. }
	\label{fig:CSfr}
\end{figure*}

CS of HS images aims to preciously reconstruct an HS data $\mathcal{X} \in \mathbb{R}^{h \times v \times z}$ from a few compressive measurements $\textbf{y} \in \mathbb{R}^m$ by effective HS CS algorithms. The compressive measurements $\textbf{y}$ can be formulated by:
\begin{equation}
\label{eq:y1}
\textbf{y} = \Psi (\mathcal{X})
\end{equation}
where $\Psi$ is a measurement operator instantiated as $\Psi= \textbf{D} \cdot \textbf{H} \cdot \textbf{P}$, where $\textbf{D}$ is a random downsampling operator, $\textbf{H}$ is a random permutation matrix, $\textbf{P}$ is a WalshHadamard transform and the mapping of $\Psi$ is $ \mathbb{R}^{h \times v \times z} \rightarrow \mathbb{R}^m$ (the sampling ratio $m=hvz$). The strict reconstruction of $\mathcal{X}$ from $\mathbf{y}$ will be guaranteed by the CS theory when $\Psi$ satisfies the restricted isometry property (RIP). This compressive operator has been successfully adopted for various HS CS tasks \cite{c100,c92,c93,c94}. However, operator $\Psi$ can be replaced with real 
demands. Apparently, it is an ill-posed inverse problem to directly recover $\mathcal{X}$ from Eq. (\ref{eq:y1}). The extra prior information needs to be investigated to optimize the HS CS problem. The HS CS task can be generalized the following optimization problem:
\begin{equation}
\label{eq:hs_cs}
\begin{aligned}
\underset{\mathcal{X}}{\textrm{min}}\;&\|\textbf{y}-\Psi (\mathcal{X})\|_F^2  + \lambda F(\mathcal{X}),
\end{aligned}	
\end{equation}
where $F(\mathcal{X})$ denotes the additional regularization term to use different types of HS prior information such as spectral correlation, spatial and spectral smoothness, and non-local similarity.

\subsection{Tensor decomposition-based HS CS reconstruction methods}
\label{sect:TD-CS}

Tucker decomposition-based methods have aroused wide attention for HS CS. Tucker decomposition was first introduced into the compression of HS images to constrain the discrete wavelet transform coefficients of spectral bands \cite{c96}. Most of the following works try to study the Tucker decomposition-based variants for HS CS \cite{c100,c97,c98,c99,c95,c214}. 

$\textbf{(1) Tucker decomposition with TV}$

In one earlier work \cite{c100}, a 2-D TV norm has been penalized in an LR matrix framework, which robustly recovers a large-size HS image when the sampling ratio is only 3\%. A spectral LR model is rarely enough to depict the inherent property of HS images. 
Joint tensor Tucker decomposition with a weighted 3-D TV (JTenRe3-DTV) \cite{c97} injected a weighted 3-D TV into the LR Tucker decomposition framework to model the global spectral correlation and local spatial–spectral smoothness of an HS image. Considering the disturbance $\mathcal{E}$, the JTenRe3-DTV optimization problem for HS CS can be expressed as
\begin{equation}
\label{eq:TDTV_cs}
\begin{aligned}
&\min _{\mathcal{X}, \mathcal{E}, \mathcal{C}, \mathbf{U}_{i}} \frac{1}{2}\|\mathcal{E}\|_{F}^{2}+\lambda\|\mathcal{X}\|_{3{\rm DwTV}} \\
&\text { s.t. } \mathbf{y}=\Psi(\mathcal{X}), \mathcal{X}=\mathcal{A} \times_{1} \mathbf{B}_{1} \times_{2} \mathbf{B}_{2} \times_{3} \mathbf{B}_{3}+\mathcal{E}
\end{aligned}	
\end{equation}
In \cite{c102}, the LR tensor constraint of Eq.(\ref{eq:TDTV_cs}) was replaced by the TNN. 

$\textbf{  (2) Tucker decomposition with non-local similarity}$

The Tucker decomposition methods with non-local similarity either cluster similar patches into a 4-D group or unfold 2-D patches into a 3-D group.  
Du \textit{et al}. \cite{c101} represented each local patch of HS images as a 3-D tensor and grouped similar tensor patches to form a 4-D tensor per cluster. Each tensor group can be approximately decomposed by a sparse coefficient tensor and a few matrix dictionaries. 
Xue \textit{et al}. \cite{c99} unfolded a series of 3-D cubes into 2-D matrices along the spectral modes and stacked these matrices as a new 3-D tensor. The spatial sparsity, the non-local similarity, and the spectral correlation were simultaneously employed to obtain the proposed model
\begin{equation}
\label{eq:TDNON_cs}
\begin{aligned}
\min _{\mathbf{x}, \mathcal{A}_{p}, \mathbf{B}_{1 p}, \mathbf{B}_{2 p}, \mathbf{B}_{3 p}} &\sum_{p=1}^{P} \frac{\lambda_{1}}{2}\left\|\mathcal{X}_{p}-\mathcal{A}_{p} \times_{1} \mathbf{B}_{1 p} \times_{2} \mathbf{B}_{2 p} \times_{3} \mathbf{B}_{3 p}\right\|_{F}^{2}\\
&+\lambda_{2}\left\|\mathcal{A}_{p}\right\|_{1}+\lambda_{3} L(\mathcal{X}_{p}) \\
\text { s.t. } \mathbf{y}=\Phi \mathbf{x}, \mathcal{X}_{p}&=\mathcal{A}_{p} \times_{1} \mathbf{B}_{1 p} \times_{2} \mathbf{B}_{2 p} \times_{3} \mathbf{B}_{3 p}, \\
\mathbf{B}_{i p}^{T} \mathbf{B}_{i p}= \mathbf{I}&(i=1,2,3)
\end{aligned}	
\end{equation}
where $p = 1, . . . P$, and $P$ denotes the group number, $\mathbf{x} \in \mathbb{R}^{hvz}$ denotes the vector form of X, $L(\mathbf{X})$ is the TTN of $\mathcal{X}$,

$\textbf{(3) TR-based methods}$

Unlike the Tucker decomposition methods \cite{c101,c99} which directly captured the LR priror in the original image space at the cost of high computation, a novel subspace-based non-local TR decomposition (SNLTR) approach projected an HS image into a low-dimensional subspace \cite{c90}. The non-local similarity of the subspace coefficient tensor is constrained by a TR decomposition model. The SNLTR model is presented as
\begin{equation}
\label{eq:SNLTR-cs}
\begin{aligned}
&\min _{\mathbf{E}, \mathbf{Z}, \mathcal{L}_{i}, \mathcal{G}_{i}} \frac{1}{2}\|\mathbf{y}-\Psi(\mathbf{E} \mathbf{Z})\|_{F}^{2}+\lambda \sum_{i}\left(\frac{1}{2}\left\|\Re_{i} \mathbf{Z}-\mathcal{L}_{i}\right\|_{F}^{2}\right) \\
&\text { s.t. } \mathbf{E}^{T} \mathbf{E}=\mathbf{I}, \quad \mathcal{L}_{i}=\Phi\left(\left[\mathcal{G}_{i}\right]\right)
\end{aligned}	
\end{equation}

\subsection{HS Kronecker CS methods}
\label{sect:TD-KCS}

Unlike the current 1-D or 2-D sampling strategy, Kronecker CS (KCS) comprises Kronecker-structured sensing matrices and sparsifying bases for each HS dimension \cite{c214,8000407}. Based on multidimensional multiplexing, Yang \textit{et al}. \cite{c106} used a tensor measurement and a nonlinear sparse tensor coding to develop a self-learning tensor nonlinear CS (SLTNCS) algorithm. The sampling process and sparse representation can be represented as the model based on Tucker decomposition. Generally, an HS image $\mathcal{X} \in \mathbb{R}^{n_1 \times n_2 \times n_3} $ can be expressed as the following Tucker model:
\begin{equation}
\label{eq:KCS}
\begin{aligned}
\mathcal{X}=\mathcal{S} \times_{1} \Phi_{1} \times_{2} \Phi_{2} \times_{3} \Phi_{3}
\end{aligned}	
\end{equation}
where $\mathcal{S} \in \mathbb{R}^{m_1 \times m_2 \times m_3} $ stands for an approximate block-sparse tensor in terms of a set of three basis matrices {$\Phi_{j} \in \mathbb{R}^{ k_j \times k_j}$}, with $m_j \ll k_j, j=1,2,3 $.

In the context of KCS, three measurement or sensing matrices denoted by $\Psi_j, j=1,2,3$ of size $n_j \times k_j$ with $n_j \ll k_j$ are used to reduce the dimensionality of the measurement tensor. The compressive sampling model is given as
\begin{equation}
\label{eq:KCS-psi}
\begin{aligned}
\mathcal{Y}&=\mathcal{X} \times_{1} \Psi_{1} \times_{2} \Psi_{2}\times_{3} \Psi_{3} \\
&=\mathcal{S} \times_{1} \mathbf{Q}_{1} \times_{2} \mathbf{Q}_{2} \times_{3} \mathbf{Q}_{3}
\end{aligned}	
\end{equation}
where $\mathbf{Q}_j = \Phi_j \Psi_j, j=1,2,3$.

Zhao \textit{et al}. \cite{c106} designed a 3-D HS KCS mechanism to achieve independent samplings in three dimensions. The suitable sparsifying bases were selected and the corresponding optimized measurement matrices were generated, which adjusted the distribution of sampling ratio for each dimension of HS images. Yang \textit{et al}. \cite{c95} constrained the nonzero number of the Tucker core tensor to explore the spatial-spectral correlation. To address the issue of the computational burden on the data reconstruction of early HS KCS techniques, researchers have proposed several tensor-based methods such as the tensor-form greedy algorithm, N-way block orthogonal matching pursuit (NBOMP) \cite{6797642}, beamformed mode-based sparse estimator (BOSE) \cite{7544443} and Tensor-Based Bayesian Reconstruction (TBR) \cite{c107}. The TBR model exploited the multi-dimensional block-sparsity of tensors, which was more consistent with the sparse model in HS KCS than the conventional CS methods. A Bayesian reconstruction algorithm was developed to achieve the decoupling of hyperparameters by a low-complexity technique. 


\subsection{Experimental results and analysis}
\label{sect:experiment_cs}

An HS data experiment is employed to validate the effectiveness of tensor-based models on HS CS with four different sample ratios i.e. $1\%$, $5\%$, $10\%$, $20\%$. The Reno data set selected for HS CS experiments is size of $150 \times 150 \times 100$. The randomly permuted Hadamard transform is adopted as the compressive operator. Tab. \ref{tab:tab-cs} compares the reconstruction results by SLNTCS and JTenRe3DTV. They have quality decays with sample ratios decreasing, but SLNTCS obtains poorer results than JTenRe3DTV in lower sampling ratios. 

\begin{figure}[htb]
	\begin{center}
		\includegraphics[height=5.2cm]{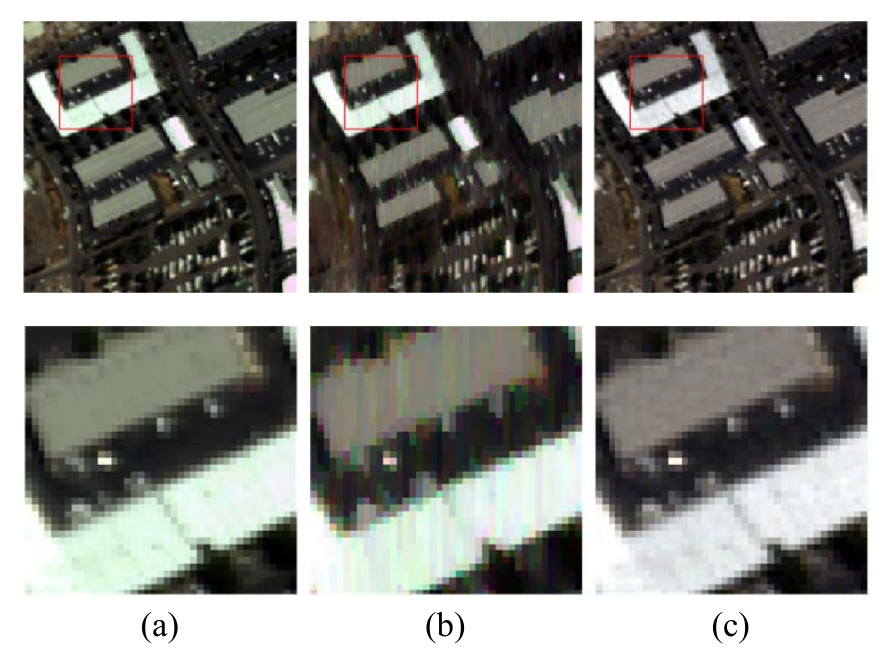}
	\end{center}
	\caption[houston]{ Inpainting results by different methods under $10\%$ sampling ratio. (a) Original Reno image, (b) SLNTCS, (c) JTenRe3DTV. }
	\label{fig:Visio-cs}
\end{figure}

In the light of visual comparison, one representative band in the sampling ratio $10\%$ is presented in Fig. \ref{fig:Visio-cs}. The basic texture information can be found in the results of two HS CS algorithms. As shown in the enlarged area, SLNTCS causes some artifacts, but JTenRe3DTV produces a more acceptable result with the smoothing white area than SLNTCS.

 \begin{table}[!htbp]
\centering
\caption{A quantitative comparison of different selected algorithms for HS CS.}
\begin{spacing}{1.0}
\scalebox{0.8}{
\begin{tabular}{c|c|cccc}
\hline
Method & Index & $1\%$ & $5\%$ & $10\%$ & $20\%$ \\
\hline
SLNTCS & MPSNR & 18.70 & 24.44 &27.72  & 32.14 \\
& MSSIM &  0.3273 & 0.6593   &  0.8047 & 0.9159\\
& ERGAS &  23.3411 & 12.1203 & 8.3119  & 5.0263\\
& MSAD  &  22.0.35 & 11.2031 & 7.6354  & 4.6003\\
\hline
JTenRe3DTV  & MPSNR & 27.91 & 34.54 & 36.28 & 37.41 \\
& MSSIM & 0.8116 & 0.9443 & 0.9638 & 0.9709\\
& ERGAS & 8.2422 & 4.0139 & 3.2990 & 2.9124\\
& MSAD  & 7.5545 & 3.5703 & 2.9233& 2.5723\\
\hline
\end{tabular}
}
\label{tab:tab-cs}
\end{spacing}
\end{table}

\subsection{Future challenges}
\label{sect:challenges_cs} 
The low acquisition rate of CS inspires a novel development potentiality for HS RS. Many tensor-based methods have been proposed to achieve remarkable HS CS reconstruction results at a lower sampling ratio. However, here we briefly point out some potential challenges.

Some novel tensor decomposition approaches need to be explored. In the past research works, Tucker decomposition has been successfully applied for HS CS. But with the development of the tensorial mathematical theory, many tensor decomposition models have been proposed and introduced in other HS applications. Therefore, how best to find more appropriate tensor decomposition for HS CS is a vital challenge. 

Noise degradation usually has a negative influence in HS CS sampling and reconstruction, which is hardly ignored in the real HS CS real imaging process. As a result, considering the noise interference and enhancing the robustness of noise in the CS process remain challenging.

\section{HS AD}
\label{sect:HSI-AD}

\begin{figure}[htb]
	\begin{center}
		\includegraphics[height=7.5cm]{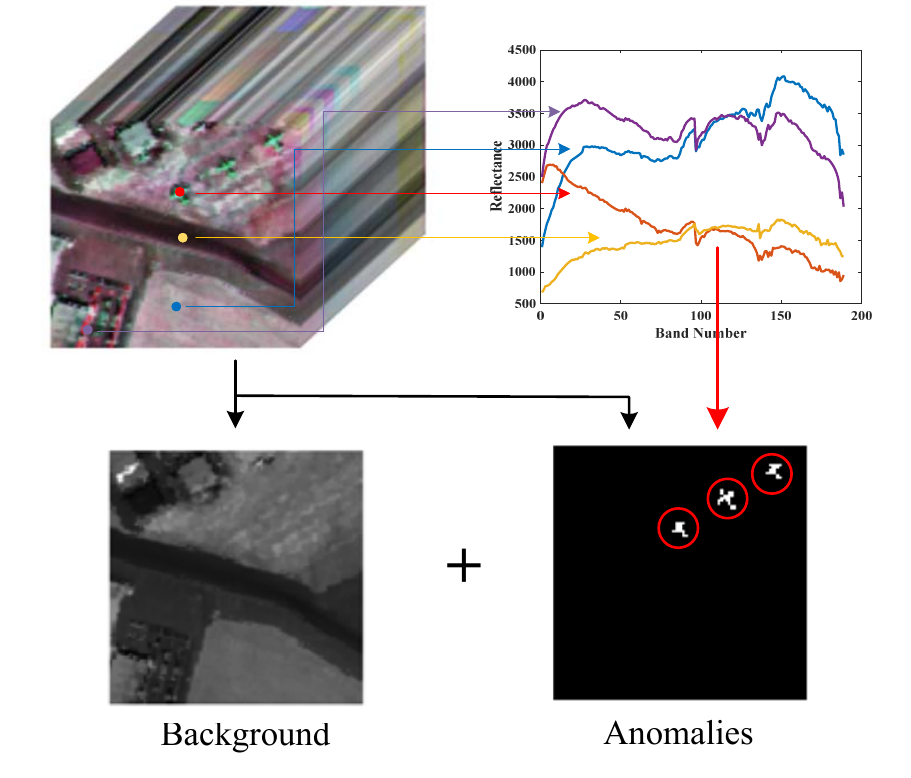}
	\end{center}
	\caption[AD]{A schematic diagram of HS image anomaly detection. }
	\label{fig:Visio-AD}
\end{figure}

 HS AD aims to discover and separate the potential man-made objects from observed image scenes, which is typically constructive for defense and surveillance developments in RS fields, such as mine exploration and military reconnaissance. For instance, aircrafts in the suburb scene and vehicles in the bridge scene are usually referred to as anomalies or outliers. In Fig. \ref{fig:Visio-AD}, AD can be regarded as an unsupervised two-class classification problem where anomalies occupy small areas compared with their surrounding background. The key to coping with this problem is to exploit the discrepancy between anomalies and their background. Anomalies commonly occur with low probabilities and their spectral signatures are quite different from neighbors.

HS images containing two spatial dimensions and one spectral dimension are intrinsically considered as a three-order tensor. Tensor-based approaches have been gradually attaching attention for HS AD in recent years. Tucker decomposition is the first and essential type of tensor-decomposition methods used for HS AD. Therefore, in the following sections, we mainly focus on the Tucker decomposition-based methods and a few other types of tensor-based methods.

\subsection{Tensor decomposition-based HS AD methods}
\label{sect:TD_ad}

$\textbf{ (1) Tucker decomposition-based methods }$

 An observed HS image $\mathcal{T}$ can be decomposed into two parts by Tucker decomposition, i.e.,
\begin{equation}
    \begin{aligned}
     \mathcal{T} = \mathcal{X} + \mathcal{S}
    \end{aligned}
\end{equation}
where $\mathcal{X}$ is LR background tensor and $\mathcal{S}$ is the sparse tensor consisting of anomalies. The Tucker decomposition for AD is formulated as the following optimization
\begin{equation}
	\begin{split}
	\label{eq:Tucker-AD}
	\left\{\begin{array}{l}
\mathcal{X}=\mathcal{A} \times_{1} \mathbf{B}_{1} \times_{2} \mathbf{B}_{2} \times_{3} \mathbf{B}_{3} \\
\mathcal{S}=\mathcal{T}-\mathcal{X}
\end{array}\right.
\end{split}    
\end{equation}

Many Tucker decomposition-based variants have been studied to improve the AD accuracy. Li \textit{et al}. \cite{c110} proposed a LR tensor decomposition based AD (LTDD) model, which employed Tucker decomposition to obtain the core tensor of the LR part. The final spectral signatures of anomalies is extracted by an unmixing approach. After the Tucker decomposition processing, Zhang \textit{et al}. \cite{c111} utilized a reconstruction-error-based method to eliminate the background pixels and remain the anomaly information. Zhu \textit{et al}. \cite{c139} advocated a weighting strategy based on tensor decomposition and cluster weighting (TDCW). In TDCW, Tucker decomposition was adopted to obtain the anomaly part. K-means clustering and segmenting, were assigned as post-processing steps to achieve a performance boost.
Song \textit{et al}. \cite{c113} proposed a tensor-based endmember extraction and LR decomposition (TEELRD) algorithm, where Tucker decomposition and k-means are employed to construct a high-quality dictionary.

Based on Tucker decomposition, Qin \textit{et al}. \cite{c115} proposed a LR and sparse tensor decomposition (LRASTD). The LRASTD can be formulated as
\begin{equation}
	\begin{split}
	\label{eq:LRASTD}
&	\min_{\mathcal{A},\mathcal{S}} || \mathcal{A} ||_* + \beta ||\mathcal{A} ||_1 +  \lambda || \mathcal{S} ||_{2,2,1}\\
 & {\rm s.t.} \quad	\mathcal{X}=\mathcal{A} \times_{1} \mathbf{B}_{1} \times_{2} \mathbf{B}_{2} \times_{3} \mathbf{B}_{3} + \mathcal{S}
\end{split}    
\end{equation}
where $|| \mathcal{S} ||_{2,2,1} = \sum^z_{k=1} || \mathcal{S}(:,:,k) ||_F$.

$\textbf{ (2) Other Tensor-based methods }$

 Chen \textit{et al}. \cite{c112} presented a TPCA-based pre-processing method to separate a principal component part and a residual part. Li \textit{et al}. \cite{c114} proposed a prior-based tensor approximation (PTA) approach, where the background was constrained by a truncated nuclear norm (TRNN) regularization and a spatial TV. The proposed PTA can be expressed as
 \begin{equation}
     \begin{aligned}
&\arg \min _{\mathcal{X}, \mathcal{S}} \frac{1}{2}\left(\left\|\mathbf{D}_{H} \mathbf{X}_{(1)}\right\|_{F}^{2}+\left\|\mathbf{D}_{v} \mathbf{X}_{(2)}\right\|_{F}^{2}\right)+\alpha\left\|\mathbf{X}_{3}\right\|_{r}+\beta\left\|\mathbf{S}_{3}\right\|_{2,1} \\
&\text { s.t. }\left\{\begin{array}{l}
\mathcal{Y}=\mathcal{X}+\mathcal{S} \\
\mathcal{X}_{1}=\operatorname{unfold}_{1}(\mathcal{X}) \\
\mathcal{X}_{2}=\operatorname{unfold}_{2}(\mathcal{X}) \\
\mathcal{X}_{3}=\operatorname{unfold}_{3}(\mathcal{X}) \\
\mathcal{S}_{3}=\operatorname{unfold}_{3}(\mathcal{S})
\end{array}\right.
\end{aligned}
 \end{equation}
where $\mathbf{D}_{H} \in \mathbb{R}^{(h-1) \times h}$ and $\mathbf{D}_{v} \in \mathbb{R}^{(v-1) \times v}$ are defined as
$\mathbf{D}_{H}=\left[\begin{array}{ccccc}
1 & -1 & & & \\
& 1 & -1 & & \\
& & \ddots & \ddots & \\
& & & 1 & -1
\end{array}\right]$\\
and
$\mathbf{D}_{v}=\left[\begin{array}{ccccc}
1 & -1 & & & \\
& 1 & -1 & & \\
& & \ddots & \ddots & \\
& & & 1 & -1
\end{array}\right]$

Wang \textit{et al}. \cite{minghuaTC} proposed a novel tensor LR and sparse representation method with a PCA pre-processing step, namely PCA-TLRSR, which was the first time to expand the concept of Tensor LR representation in HS AD and exploited the 3-D inherent structure of HS images. Assisted by the multi-subspace learning of the tensor domain and the sparsity constraint along the joint spectral-spatial dimensions, the LR background and anomalies are separated in a more accurate manner.


\subsection{Experimental results and analysis}
\label{sect:experiment_ad}

\begin{figure}[htb]
	\begin{center}
		\includegraphics[height=8.5cm, angle = 90]{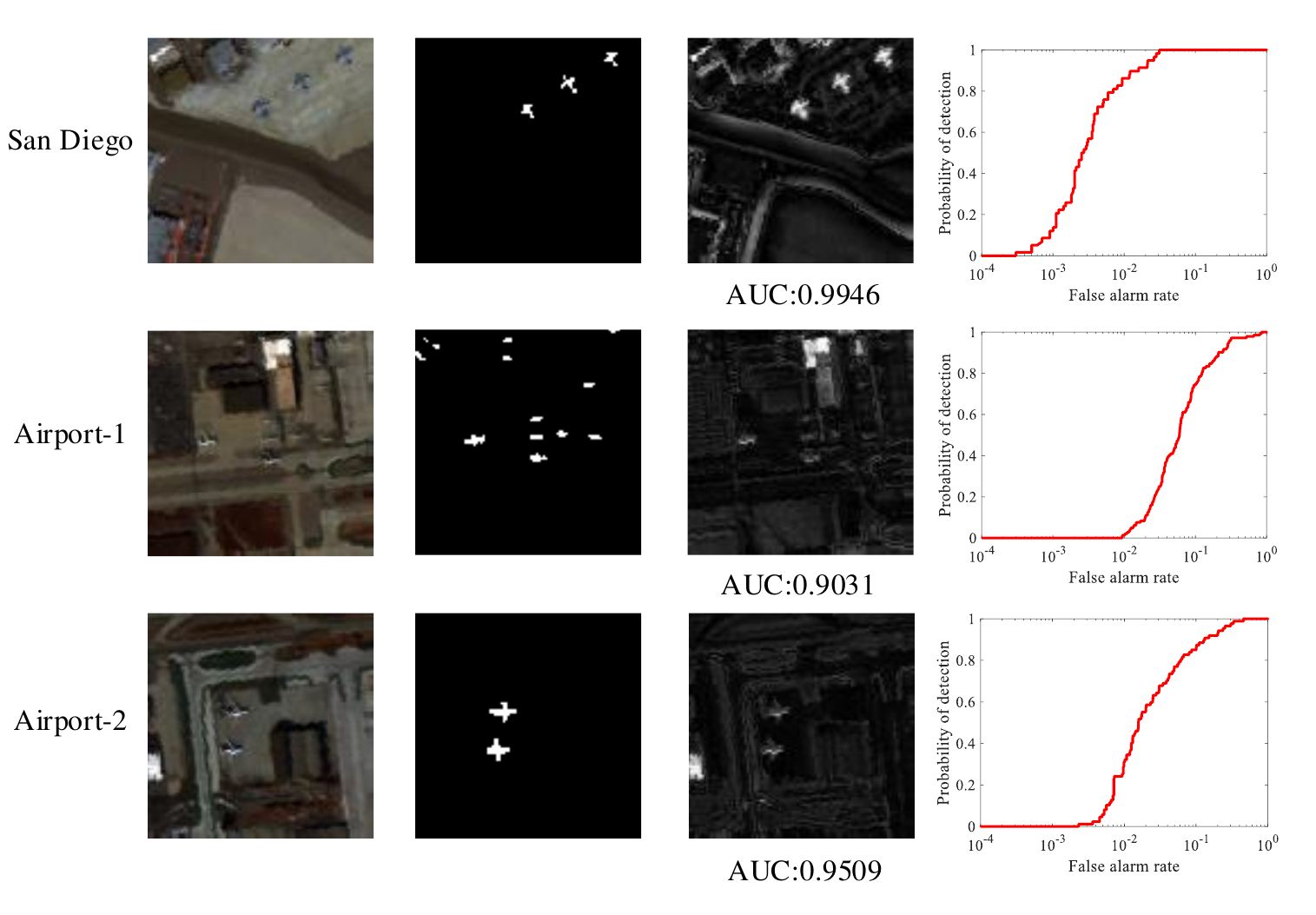}
	\end{center}
	\caption[AD]{Original HS images, ground-truth maps, detection maps, and AUC curves of PTA on different data sets. }
	\label{fig:AD-exp}
\end{figure}

Herein, we take an example of PTA on three HS data sets for AD. The San Diego data set \cite{c216} was captured by the Airborne Visible/Infrared Imaging Spectrometer (AVIRIS) sensor over the San Diego airport, CA, USA. Three flights are obviously observed in the selected region with the size $100 \times 100 \times 189$. The Airport-1 and Airport-2 \cite{c217} were also acquired by AVIRIS sensor. As shown in the second column of Fig. \ref{fig:AD-exp}, flights are regarded as anomalies in different airport scenes.  

As the detection maps of Fig. \ref{fig:AD-exp} display, most of flights are clearly detected by PTA. Except for the visual observation of resulted anomaly maps, the receiver operating characteristic (ROC) curve \cite{c218} and the area under the ROC curve (AUC) \cite{c219} are employed to quantitatively assess the detection accuracy of the tensor-based method. The ROC curve plots the varying relationship of the probability of detection (PD) and false alarm rate (FAR) for extensive possible thresholds. The area under this curve is calculated as AUC, whose ideal result is 1. PTA is capable to achieve a high detection rate and low FAR. The AUC values derived from PTA is higher than 0.9.

\subsection{Future challenges}
\label{sect:challenges_ad} 

Tucker decomposition-based models have been well developed by researchers, yet other types of tensor decompositions are rarely investigated in the HS AD community. In other words, how best to introduce other novel tensor decomposition frameworks into AD is a key challenge.

Although most anomalies are successfully detected, some background pixels like roads and roofs usually remain. The more complex background and the fewer targets make the difficulty of AD increase. To solve this problem, researchers need to explore multiple features and suitable regularizations.

The background and anomalies are often modeled as the LR part and the sparse part of HS images. The 3-D inherent structure of HS images is exploited by tensor decomposition-based methods. The spatial sparsity and the 3-D inherent structure of anomalies should be considered by a consolidated optimization strategy.

 \section{HS-MS fusion}
\label{sect:HSI-SR}

\begin{figure*}[htb]
	\begin{center}
		\includegraphics[height=8cm]{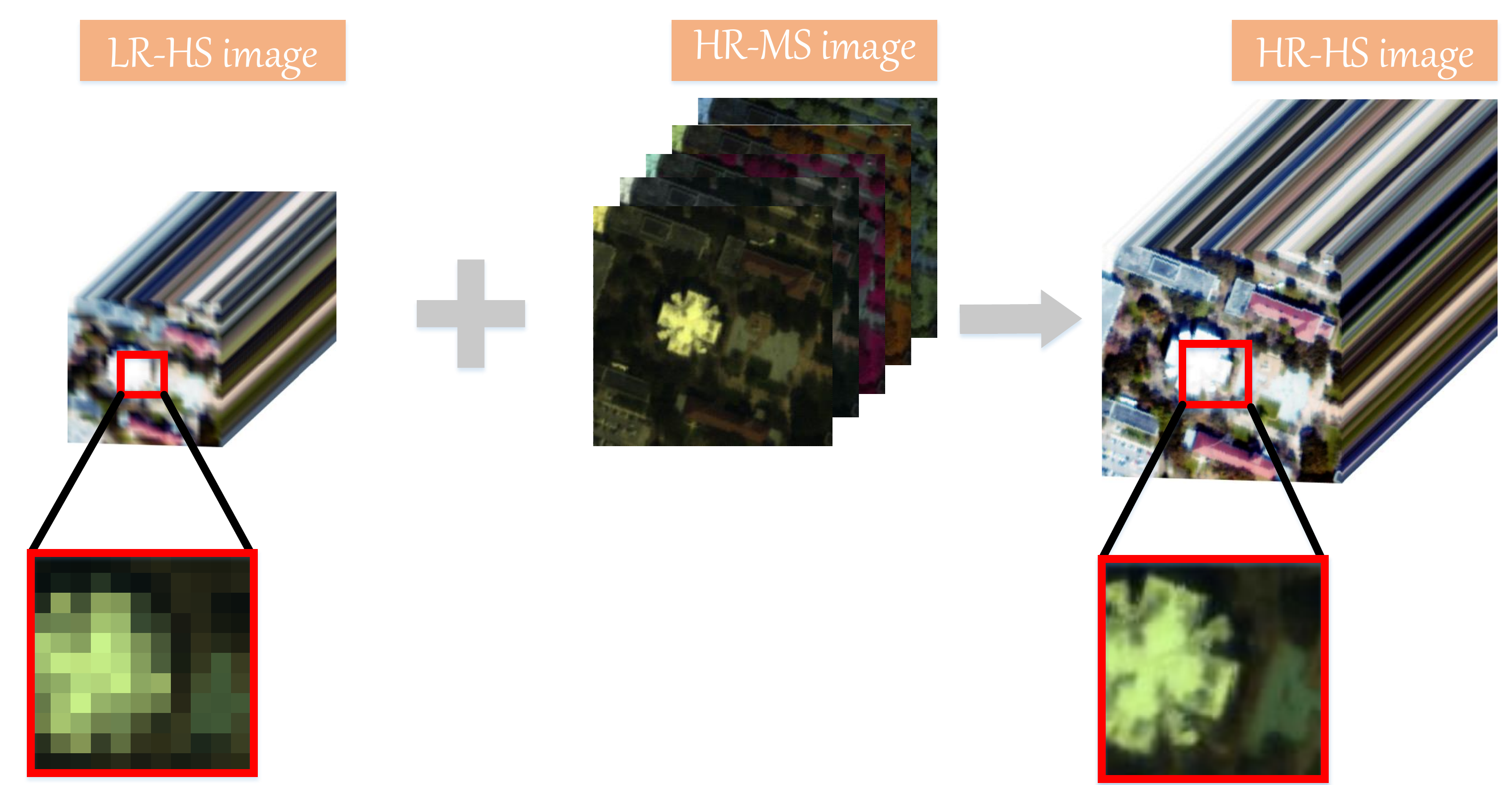}
	\end{center}
	\caption[unmixing]{Illustration of HS and MS fusion. }
	\label{fig:Visio-sr-fr}
\end{figure*}

HS images provide abundant and varied spectral information, yet hardly contain high-spatial resolution owing to the limitations of sun irradiance \cite{c140}, and imaging systems \cite{c141}. On the contrary, MS images are captured with low-spectral resolution and high-spatial resolution. HS and MS fusion aims to improve the spatial resolution of HS images with the assistance of MS images and generate final HS images with high-spatial resolution and original spectral resolution. The high-quality fused HS images benefit for the in-depth recognition and 
insight of materials, which contributes to many RS real applications, such as object classification and change detection of wetlands and farms \cite{7182258,7895167,8961105,gu2019superpixel,hong2021graph,chen2022fccdn,hong2022Spectral}.

Fig. \ref{fig:Visio-sr-fr} depicts an HS and MS fusion process to generate an HR-HS image. Suppose that a desired high-spatial-spectral resolution HS (HR-HS) image, a low-resolution HS (LR-HS) image, and a high-resolution MS (HR-MS) image are denoted by $\mathcal{X} \in \mathbb{R}^{ H \times V \times B}$, $\mathcal{Y} \in  \mathbb{R}^{ h \times v \times B}$ and $\mathcal{Z} \in \mathbb{R}^{ H \times V \times b}$ ($H \gg h$, $V \gg v$, $B \gg b$), respectively. A LR-HS image is seen as a spatially downsampled and blurring version of $\mathcal{X}$, and a HR-MS image is the spectrally downsampled version of $\mathcal{X}$. The two degradation models are expressed as follow
\begin{equation}
    \begin{aligned}
    \label{eq:HSR}
    \mathbf{Y}_{(3)} =  \mathbf{X}_{(3)} \mathbf{R} + \mathbf{N}_h
    \end{aligned}
\end{equation}
\begin{equation}
    \begin{aligned}
    \label{eq:HSR2}
    \mathbf{Z}_{(3)} = \mathbf{G}  \mathbf{X}_{(3)} + \mathbf{N}_m
    \end{aligned}
\end{equation}
where $\mathbf{R} = \mathbf{B} \mathbf{K}$, $\mathbf{B}$ denotes a convolution blurring operation. $\mathbf{K}$ is a spatial downsampling matrix, and $\mathbf{G}$ represents a spectral-response function if a MS image sensor, which can be regarded as a spectral downsampling matrix. $\mathbf{N}_h$ and $\mathbf{N}_m$ stand for noise.

According to references \cite{c142,c143,c144}, $\mathbf{R}$ and $\mathbf{G}$ are assumed to be given in advance of solving the HS SR problem 
\begin{equation}
    \begin{aligned}
    \label{eq:HSR-pro}
    \min_{\mathcal{X}} ||\mathbf{Y}_{(3)} -  \mathbf{X}_{(3)} \mathbf{R} ||^2_F +
    ||\mathbf{Z}_{(3)} - \mathbf{G}  \mathbf{X}_{(3)} ||^2_F + \tau f ({\mathcal{X}})
    \end{aligned}
\end{equation}
where the first and second F-norm are data-fidelity terms with respect with models (\ref{eq:HSR}) and (\ref{eq:HSR2}), $f ({\mathcal{X}})$ represents the prior regularization  pertinent to the desired property on the HR-HS ${\mathcal{X}}$. 
In the next section, we review currently advanced HS SR methods from two categories: tensor decomposition and prior-based tensor decomposition models.

\subsection{Tensor Factorizations for SR}
\label{sect:TDforSR} 

\subsubsection{CP Decomposition Model}

 Initially, Kanatsoulis \textit{et al}. \cite{c145} employed a coupled CP decomposition framework for HS SR. The CP decomposition of an HR-HS tensor $\mathcal{X}$ can be expressed as
 \begin{equation}
     \begin{aligned}
     \label{eq:sr-cp}
      {\mathcal { X }} &=\sum_{r=1}^{R} \mathbf{a}_{r} \circ \mathbf{b}_{r} \circ  \mathbf{c}_{r} \\
      & =\llbracket  \mathbf{A},  \mathbf{B},  \mathbf{C} \rrbracket\\
     \end{aligned}
 \end{equation}
where the latent LR factors are $\mathbf{A}= [\mathbf{a}_1,...,\mathbf{a}_r ]$, $\mathbf{B}= [\mathbf{b}_1,...,\mathbf{b}_r ]$, and $\mathbf{C}= [\mathbf{c}_1,...,\mathbf{c}_r ]$.
 In \cite{c145}, the coupled CP decomposition gave the following assumption
  \begin{equation}
     \begin{aligned}
     \label{eq:sr-cp-assum}
      {\mathcal { Y }} 
       =\llbracket  \mathbf{P}_1 \mathbf{A},  \mathbf{P}_2 \mathbf{B},  \mathbf{C} \rrbracket;
             {\mathcal { Z }} 
       =\llbracket   \mathbf{A},  \mathbf{B},  \mathbf{P}_3 \mathbf{C} \rrbracket
     \end{aligned}
 \end{equation}
 where $\mathbf{P}_1 \in \mathbb{R}^{h \times H} $, $\mathbf{P}_2 \in \mathbb{R}^{v \times V} $, and $\mathbf{P}_3 \in \mathbb{R}^{b \times B} $ are three linear degradation matrices. The identifiability of HS SR based on the algebraic properties of CP decomposition is guaranteed under relaxed conditions. However, LR properties of different dimensions are treated equally, which is rarely suitable for real HS SR. Subsequently, Kanatsoulis \textit{et al}. \cite{c151} a SR cube algorithm (SCUBA) that combined the advantages of CP decomposition and matrix factorization. Xu \textit{et al}. \cite{c152} improved CP decomposition-based method by adding a non-local tensor extraction module.


\subsubsection{Tucker Decomposition Model}

 Li \textit{et al}. \cite{c146} extended a coupled sparse tensor factorization (CSTF) approach, in which the fusion problem was transformed into the estimation of dictionaries along three modes and corresponding sparse core tensor. When a tensor $\mathcal{X}$ is decomposed by Tucker decomposition
\begin{equation}
    \begin{aligned}
    \label{eq:sr-tucker}
    \mathcal{X}=\mathcal{W} \times_{1} \mathcal{A} \times_{2} \mathcal{B} \times_{3} \mathcal{C}
    \end{aligned}
\end{equation}

The LR-HS and HR-HS degradation models are rewritten as
\begin{equation}
    \begin{aligned}
    \label{eq:sr-tucker2}
    \mathcal{Y}&=\mathcal{W} \times_{1}(\mathbf{P}_{1} \mathbf{A}) \times_{2}(\mathbf{P}_{2} \mathbf{B}) \times_{3} \mathbf{C} \\
    &=\mathcal{W} \times_{1}\mathbf{A}^* \times_{2}\mathbf{B}^*\times_{3} \mathbf{C} \\
    \end{aligned}
\end{equation}

\begin{equation}
    \begin{aligned}
    \mathcal{Z}&=\mathcal{W} \times_{1}\mathbf{A} \times_{2} \mathbf{B} \times_{3} (\mathbf{P}_{3}\mathbf{C} )\\
    &=\mathcal{W} \times_{1}\mathbf{A} \times_{2} \mathbf{B} \times_{3} \mathbf{C}^*
    \end{aligned}
\end{equation}
where $\mathbf{A}^* = \mathbf{P}_{1} \mathbf{A}$, $\mathbf{B}^* = \mathbf{P}_{2} \mathbf{B}$, and $\mathbf{C}^* = \mathbf{P}_{3}\mathbf{C}$ are the downsampled dictionaries along three modes. Taking the sparsity of core tensor $\mathcal{W}$, Li \textit{et al}. formulated the fusion problem as follows
\begin{equation}
    \begin{aligned}
    \label{eq:sr-CSTF}
    \min_{ \mathbf{A}, \mathbf{B}, \mathbf{C},  \mathcal{W}} &|| \mathcal{Y}-\mathcal{W} \times_{1} \mathbf{A}^* \times_{2}\mathbf{B}^*\times_{3} \mathbf{C} ||^2_F + \\ 
     &|| \mathcal{Z} - \mathcal{W} \times_{1}\mathbf{A} \times_{2} \mathbf{B} \times_{3} \mathbf{C}^* ||^2_F + \lambda || \mathcal{W} ||_1
    \end{aligned}
\end{equation}

The $l_1$ norm in Eq. (\ref{eq:sr-CSTF}) was replaced by a $l_2$ norm in \cite{c147}. Prévost \textit{et al}. \cite{c153} assumed HR-HS images assessed approximately low multilinear rank and developed an SR algorithm based on coupled Tucker tensor approximation (SCOTT) with HOSVD. In the Tucker decompositon framework and the BT of decompositon framework (named CT-STAR and CB-STAR) \cite{c155}, an additive variability term was admitted for the study of the general identifiability with theoretical guarantees. Zare \textit{et al}. \cite{c170} offered a coupled non-negative Tucker decomposition (CNTD) method to constrain the nonnegativity of two Tucker spectral factors.

 $\textbf{Non-local Tucker decomposition:}$ Wan \textit{et al}. \cite{c154} grouped 4-D tensor patches using the spectral correlation and similarity under Tucker decomposition. Dian \textit{et al}. \cite{c156} offered a non-local sparse tensor factorization (NLSTF) method, which induced core tensors and corresponding dictionaries from HR-MS images, and spectral dictionaries from LR-HS images. A modified NLSF\_SMBF version was developed for the semi-blind fusion of HS and MS \cite{c157}. However, the dictionary and the core tensor for each cluster are estimated separately by NLSTF and NLSF\_SMBF. 

 $\textbf{Tucker decomposition + TV:}$ Xu \textit{et al}. \cite{c160} presented a Tucker decomposition model with a unidirectional TV. Wang \textit{et al}. \cite{c158} advocated a non-local LR tensor decomposition and SU based approach to leverage spectral correlations, non-local similarity, and spatial-spectral smoothness.

 $\textbf{Tucker decomposition + Manifold:}$ Zhang \textit{et al}. \cite{c159} suggested a spatial–spectral-graph-regularized LR tensor decomposition (SSGLRTD). In SSGLRTD, the spatial and spectral manifolds between HR-MS and LR-HS images are assumed to be similar to those embedded in HR-HS images. Bu \textit{et al}. \cite{c161} presented a graph Laplacian-guided coupled tensor decomposition (gLGCTD) model that incorporated global spectral correlation and complementary submanifold structures into a unified framework.

\subsubsection{BT Decomposition Model}

Zhang \textit{et al}. \cite{c163} discovered the identifiability guarantees in \cite{c145,c146} at the cost of the lack of physical meaning for the latent factors under CP and Tucker decomposition. Therefore, they employed an alternative coupled nonnegative BT tensor decomposition (NN-CBCTD) approach for HS SR. The NN-CBTD model with rank-($L_r,L_r,1$) for HS SR is given as
\begin{equation}
    \begin{aligned}
    \label{eq:NNCBTD}
 \min _{\mathrm{A}, \mathbf{B}, \mathrm{C}} &\|\mathcal{Y}-\sum_{r=1}^{R}(\mathbf{P}_{1} \mathbf{A}_{r}\left(\mathbf{P}_{2} \mathbf{B}_{r}\right)^{\top}) \circ \mathbf{c}_{r}\|_{F}^{2}  \\
&+\|\mathcal{Z}-\sum_{r=1}^{R}(\mathbf{A}_{r} \mathbf{B}_{r}^{\top}) \circ \mathbf{P}_{3} \mathbf{c}_{r}\|_{F}^{2} \\
\text { s. t. } &\mathbf{A} \geq \mathbf{0}, \mathbf{B} \geq \mathbf{0}, \mathbf{C} \geq \mathbf{0}
    \end{aligned}
\end{equation}

Compared with a conference version \cite{c163}, the journal version \cite{c138} additionally gave more recoverability analysis and more flexible decomposition framework by using a advocated LL1 model and a block coordinate descent algorithm. Jiang \textit{et al}. \cite{c164} introduced a graph manifold named Graph Laplacian into the CBTD framework.

\subsubsection{TT Decomposition Model}

Dian \textit{et al}. \cite{c148} proposed a low tensor-train rank (LTTR)-based HS SR method. A LTTR prior was designed for learning correlations among the spatial, spectral, and non-local modes of 4-D FBP patches. The HS SR optimization can be obtained as
\begin{equation}
\begin{aligned}
\label{eq:LTTR}
    \min _{\mathbf{X}_{(3)}}\|\mathbf{Y}_{(3)}-\mathbf{X}_{(3)} \mathbf{R}\|_{F}^{2}+\|\mathbf{Z}_{(3)}-\mathbf{G X}_{(3)}\|_{F}^{2}+\tau \sum_{k=1}^{K}\|\mathcal{X}_{k}\|_{\mathrm{TT}}
\end{aligned}
\end{equation}
where $K$ denotes the number of clusters, the TT rank of tensor $\mathcal{Z}_k$ is defined
\begin{equation}
\begin{aligned}
\label{eq:LTTR-tt}
\| \mathcal{Z}_{k} \|_{\mathrm{TT}}=\sum_{t=1}^{3} \alpha_{t} \operatorname{LS}(\mathbf{Z}_{k \langle t\rangle})
\end{aligned}
\end{equation} 
and $\mathrm{LS}(\mathbf{A})=\sum_{i} \log (\sigma_{i}(\mathbf{A})+\varepsilon)$ with a small positive value $\varepsilon$.

Li \textit{et al}. \cite{c162} presented nonlocal LR tensor approximation and sparse representation (NLRSR) that formed the non-local similarity and sparial-spectral correlation by the TT rank constraint of 4-D non-local patches.

\subsubsection{TR Decomposition Model}
 The TR decomposition of an HR-HS tensor $\mathcal{X}  \in \mathbb{R}^{H \times V \times B}$ is represented as
 \begin{equation}
\begin{aligned}
\label{eq:SR-TR-x}
\mathcal{X}=\Phi [\mathcal{G}^{(1)}, \mathcal{G}^{(2)}, \mathcal{G}^{(3)}]
\end{aligned}
\end{equation}
where three TR factors are denoted by $\mathcal{G}^{(1)} \in \mathbb{R}^{r_{1} \times H \times r_{2}}$, $\mathcal{G}_{(2)} \in \mathbb{R}^{r_{2} \times V \times r_{3}}$, and  $\mathcal{G}^{(3)} \in \mathbb{R}^{r_{3} \times B \times r_{1}}$ with TR ranks $r = [r_1, r_2, r_3]$. Based on the TR theory, an LR-HS image is rewritten as
\begin{equation}
\begin{aligned}
\label{eq:SR-TR-y}
\mathcal{Y}={\Phi}[\mathcal{G}^{(1)} \times{ }_{2} \mathbf{P}_{1}, \mathcal{G}^{(2)} \times_{2} \mathbf{P}_{2}, \mathcal{G}^{(3)}]
\end{aligned}
\end{equation}
and an HR-MS image can be expressed as
\begin{equation}
\begin{aligned}
\label{eq:SR-TR-z}
\mathcal{Z}=\boldsymbol{\Phi}[\mathcal{G}^{(1)}, \mathcal{G}^{(2)}, \mathcal{G}^{(3)} \times_{2} \mathbf{P}_{3}]
\end{aligned}
\end{equation}

He \textit{et al}. \cite{c150} presented a coupled TR factorization (CTRF) model and a modified CTRF version (NCTRF) with the nuclear norm regularization of third/spectral TR factor. The NCTRF model is formulated as
\begin{equation}
\begin{aligned}
\label{eq:NCTRF}
\begin{aligned}
&\min _{\mathcal{G}^{(1)}, \mathcal{G}^{(2)}, \mathcal{G}^{(3)}}\|\mathcal{Y}-\boldsymbol{\Phi} [\mathcal{G}^{(1)} \times_{2} \mathbf{P}_{1}, \mathcal{G}^{(2)} \times_{2} \mathbf{P}_{2}, \mathcal{G}^{(3)} ] \|_{F}^{2} \\
&+\|\mathcal{Z}-\boldsymbol{\Phi} [\mathcal{G}^{(1)}, \mathcal{G}^{(2)}, \mathcal{G}^{(3)} \times_{2} \mathbf{P}_{3} ] \|_{F}^{2}+\lambda\|\mathbf{G}_{(2)}^{(3)}\|_{*}
\end{aligned}
\end{aligned}
\end{equation}

Eq. (\ref{eq:NCTRF}) becomes the CTRF model when removing the last term. In \cite{c150}, the benefit of TR decomposition for SR is elaborated via the theoretical and experimental proof related to a low-dimensional TR subspace.
The relationship between the TR spectral factors of LR-HS images and HR-MS images were explored in \cite{c149} with a high-order representation of the original HS image. The spectral structures of HR-HS images were kept to be consistent with LR-HS images by a graph-Laplacian regularization. 
Chen \textit{et al}. \cite{c169} presented a factor-smonthed TR decomposition (FSTRD) to capture the spatial-spectral continuity of HR-HS images.
Based on the basic CTRF model, Xu \textit{et al}. \cite{c171} advocated LR TR decomposition based on TNN (LRTRTNN), which exploited the LR properties of non-local similar patches and their TR factors.

\begin{figure*}[htb]
	\begin{center}
	\includegraphics[height=12cm]{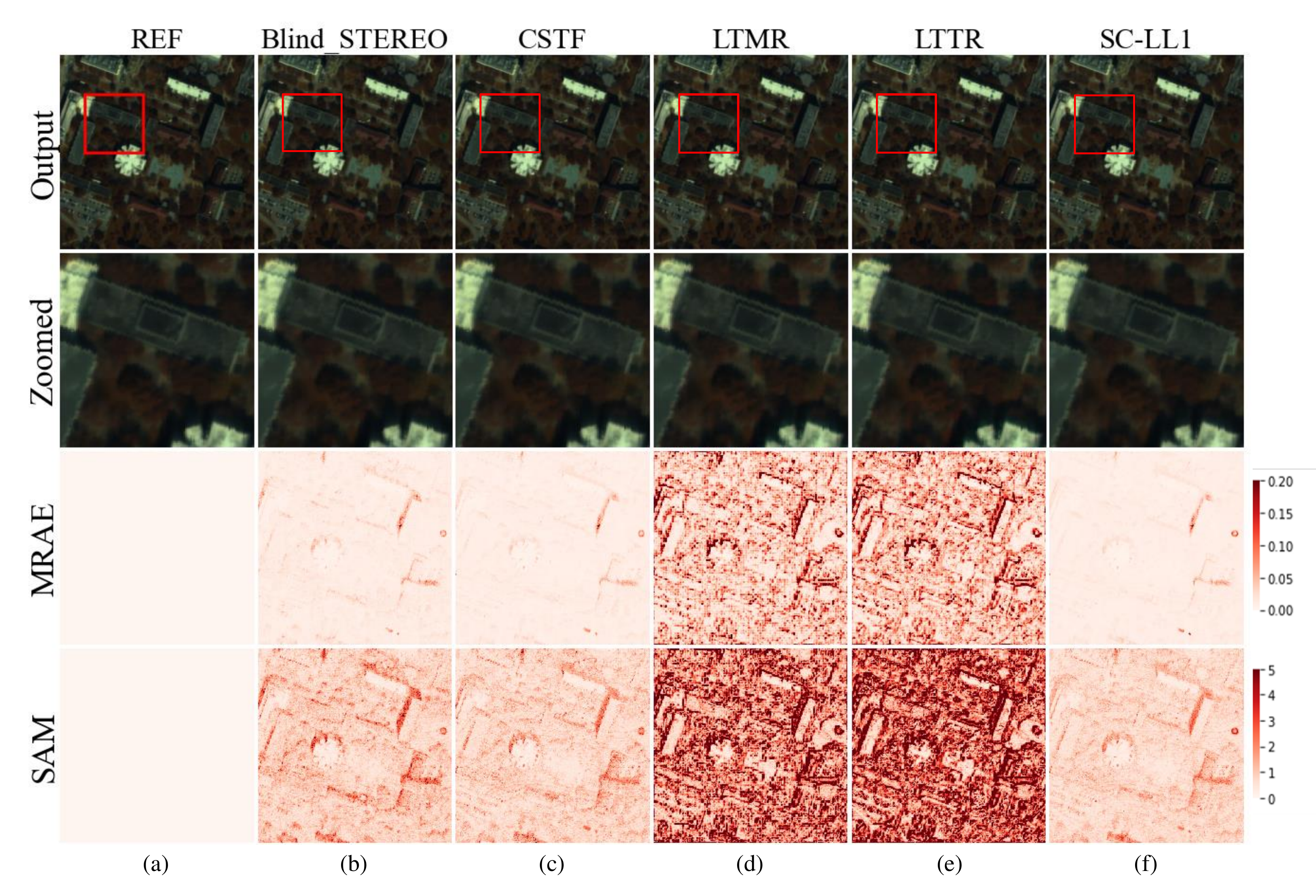}
	\end{center}
	\caption[sr]{The fusion results of five different HS-MS fusion methods. (a) REF, (b) Blind-STEREO, (c) CSTF, (d) LTMR, (e) LTTR, and (f) SC-LL1. }
	\label{fig:Visio-sr-re}
\end{figure*}
\begin{table*}[!htbp]
\centering
\caption{A quantitative comparison of different methods for HS-MS fusion.}
\begin{spacing}{1.0}
\scalebox{0.93}{
\begin{tabular}{c|ccccc}
\hline
Index & Blind-STEREO &	CSTF & LTTR & LTMR & SC-LL1 \\
\hline
 MPSNR &  53.49 & $\mathbf{54.28}$   &	38.11  & 39.58  & 54.13\\
 ERGAS &0.3584  & $\mathbf{0.3317}$  & 1.9761 & 1.5851	& 0.3076\\ 
 SAM   & 1.0132	& 0.8841  & 3.7971 & 3.0495 & $\mathbf{0.8213}$ \\
 RMSE  &0.0027	& 0.0025  & 0.0165 & 0.0132	& $\mathbf{0.0023}$\\
CC     &0.9993	& 0.9994  & 0.9819 & 0.9870	& $\mathbf{0.9995}$\\
\hline
\end{tabular}
}
\label{tab:tab-fusion}
\end{spacing}
\end{table*}

\subsubsection{Tensor Rank Minimization for SR}
\label{sect:TRMforSR} 

Based on t-SVD, Dian \textit{et al}. \cite{c165} developed a subspace based low tensor multi-rank (LTMR) that induced an HR-HS image by spectral subspace and corresponding coefficients of grouped FBPs. The specific LTMR model is expressed as
\begin{equation}
    \begin{aligned}
    \label{eq:LTMR}
    \min_{\mathcal{X}} ||\mathbf{Y}_{(3)} -  \mathbf{X}_{(3)} \mathbf{R} ||^2_F +
    ||\mathbf{Z}_{(3)} - \mathbf{G}  \mathbf{X}_{(3)} ||^2_F + \tau \sum_{k=1}^{K}\|\mathcal{X}_{k}\|_{\mathrm{TMR}}
    \end{aligned}
\end{equation}
where the multi-rank of tensor $\mathcal{X}$ is defined as $\|\mathcal{X}\|_{\mathrm{TMR}}=\frac{1}{B_{3}} \sum_{b=1}^{B_{3}} \operatorname{LS}(\hat{\mathcal{X}}(:,:,b))$, and $B$ is dimension number of the third mode of $\mathcal{X}$. To speed up the estimation of LTMR, Long \textit{et al}. \cite{c166} introduced the concept of truncation value and obtain a fast LTMR (FLTMR) algorithm. Xu \textit{et al}. \cite{c8} presented a non-local patch tensor sparse representation (NPTSR) model that characterized the the spectral and spatial similarities among non-local HS patches by the t-product based tensor sparse representation.

Considering the HS image degradation by noise, some researchers study the Noise-robust HS SR problem. Li \textit{et al}. \cite{c168} proposed a TV regularized tensor low-multilinear-rank (TV-TLMR) model to improve the performances of the mixed-noise-robust HS SR task. Liu \textit{et al}. \cite{c167} transformed the HS SR problem as a convex TTN optimization, which permitted a SR process robust to an HS image striping case.

\subsection{Experimental results and analysis}
\label{sect:experiment_sr}

In this section, we select five representative tensor decomposition-based HS-MS fusion approaches: a Tucker decomposition-based method, i.e, CSTF \cite{c146}; a CP decomposition-based method, i.e., Blind-STEREO \cite{c145}; a TT decomposition-based method, i.e., LTTR \cite{c148}; a BT decomposition-based method, i.e., SC-LL1 \cite{c138} and a tensor singular value decomposition-based method, i.e., LTMR \cite{c165}.

The quality assessment is conducted within a simulation study following Wald’s protocol \cite{c220}. One RS-HS data set is selected for the data fusion, i.e., the University of Houston campus used for the 2018 IEEE GRSS Data Fusion Contest. The original data is acquired by ITRES CASI 1500 HS camera, covering a 380-1050 nm spectral range with 48 bands at a 1-m GSD. a sub-image of $400 \times 400 \times 46 $ is chosen as the ground true after discarding some noisy bands. The input HR-MS image is generated by the reference image using the spectral response of WorldView 2, and the input LR-HS image is obtained via a Gaussian blurring kernel whose size equals five. Five quantitative metrics are used to assess the performances of the reconstructed HR-HS image, including MPSNR, ERGAS, root-mean-square error (RMSE), spectral angle mapper (SAM), and cross-correlation (CC). SAM measures the angles between the HR-HS image and the reference image, and smaller SAMs correspond to better performance. CC is a score between 0 and 1, where 1 represents the best estimation result.

Fig. \ref{fig:Visio-sr-re} presents the reconstructed false-color images, enlarged local images, SAM error heatmaps, and mean relative absolute error (MRAE) heatmaps of five HS-MS fusion methods. From Fig. \ref{fig:Visio-sr-re}, all five methods provide good spatial reconstruction results. However, LTMR and LTTR produce severe spectral distortions at the edge of the objects. In Tab. \ref{tab:tab-fusion}, the conclusion of quantitative evaluation is consistent with that of the visual one. In other words, LTTR and LTMR perform poorly in the spectral reconstruction quality. The other three methods show a competitive ability in HS-MS fusion. Especially, CSTF gains the best MPSNR and ERGAS scores, and SC-LL1 achieves the best SAM, RMSE, and CC values among the competing approaches.

\subsection{Future challenges}
\label{sect:challenges_sr} 

Though tensor decomposition-based HS-MS fusion technology has been promoted rapidly in recent years and shows a promising reconstruction ability due to its strong exploitation of spatial-spectral structure information, a number of challenges remain.

$ \textbf{Non-registered HS-MS fusion}$: Tensor decomposition-based HS-MS fusion methods focuses on the pixel-level image fusion, which implies that image registration between two input modalities is a necessary prerequisite and the fusion quality heavily depends on the registration accuracy. However, most of the current methods pay more attention to the follow-up fusion step, ignoring the importance of registration. As a challenging task, image registration handles the inputs of two modalities acquired from different platforms and times. In the future, efforts should be made to accomplish non-registered HS-MS fusion tasks.

$ \textbf{Blind HS-MS fusion}$: Existing tensor decomposition-based HS-MS fusion methods contribute to the appropriate design of handcrafted priors to derive desired reconstruction results. However, the degradation models are often given without the estimation of real PSF and spectral response function in most of tensor-based methods. It is intractable to obtain precisely the degradation functions of real cases due to the uncertainty of sensor degradation.
How to devise blind HS-MS fusion methods with unknown degradation function is a desirable challenge.

$\textbf{Inter-image variability}$: The different times or platforms of two HS and MS modalities lead to the discrepancy, referring to the inter-image variability. However, tensor decomposition-based approaches usually assume that two modalities are acquired under the same condition, and hence ignore the spectral and spatial variability that usually happens in practice. Taking the inter-image variability phenomenon into consideration when modeling the degradation process is a key challenge for future researches.

\section{HS SU}
\label{sect:HSI-unmixing}

\begin{figure*}[htb]
	\begin{center}
		\includegraphics[height=7cm]{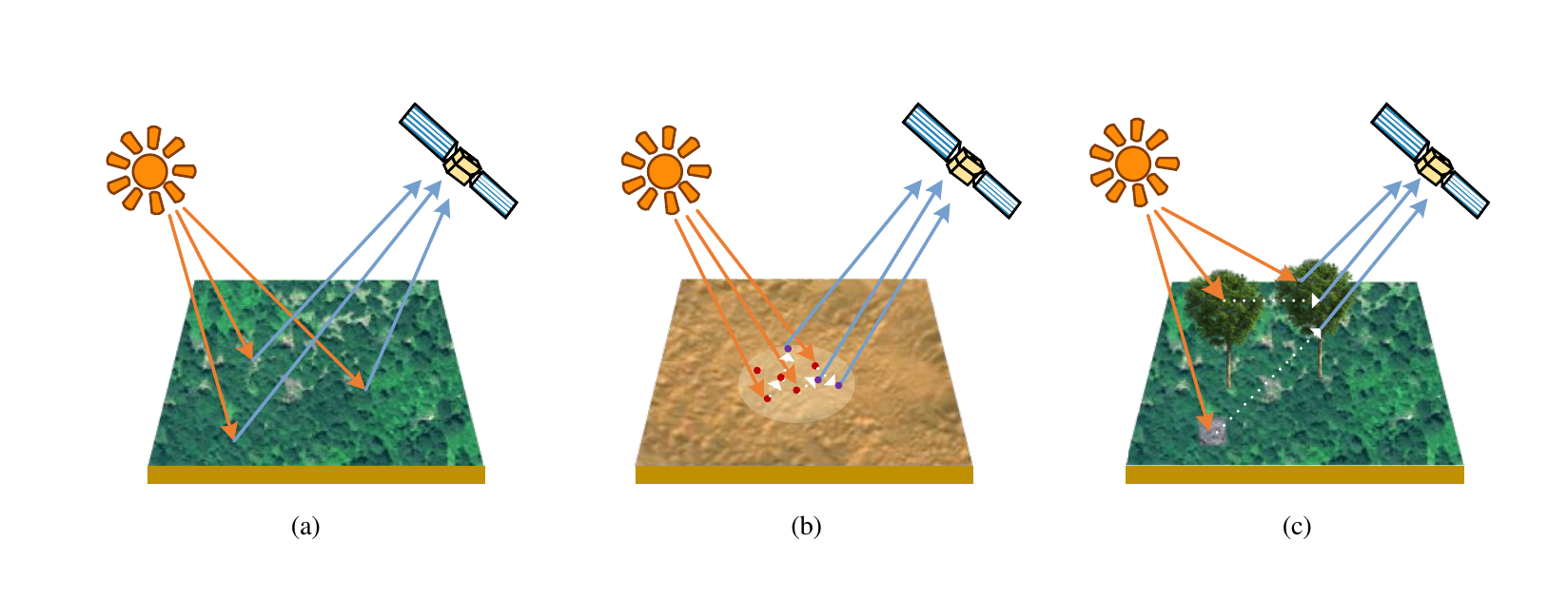}
	\end{center}
	\caption[unmixing]{Illustration of HS unmixing based on linear mixing model and nonlinear mixing model. (a) Linear mixing, (b) intimate mixture, (c) multilayered mixture. }
	\label{fig:Visio-unmixing-fr}
\end{figure*}

Owing to its acquired continuous abundance maps, SU has been widely solved the inversion problems of typical ground object parameters, such as vegetation index, surface temperature and water turbidity in several decades \cite{sonnentag2007mapping,alcantara2009improving, deng2013estimating}, and has been successfully applied in some RS applications, such as forest monitoring and land cover change detection \cite{hlavka1995unmixing}. In addition, due to the mixing phenomenon caused by heterogeneity and stratified distribution of ground objects, SU can effectively realize crop identification and monitoring \cite{lobell2004cropland, iordache2014dynamic, chi2014spectral }.

When the mixing scale is macroscopic and each incident light reaching sensors has interacted with just one material, the measured spectrum is usually regarded as a linear mixing, as shown in Fig. \ref{fig:Visio-unmixing-fr} (a). However, due to the existence of nonlinear interactions in real scenarios, several physics-based approximations of nonlinear linear mixing model (NLMM) have been proposed, mainly covering two types of mixing assumptions: intimate mixture (Fig. \ref{fig:Visio-unmixing-fr} (b)) and multilayered mixture (Fig. \ref{fig:Visio-unmixing-fr} (c)).
 The former describes the interactions suffered by the surface composed of particles at a microscopic scale. The intimate mixture usually occurs in scenes containing sand or mineral mixtures and requires a certain kind of prior knowledge of the geometric positioning of the sensor to establish the mixture model. The latter characterizes the light reflectance of various surface materials at a macroscopic scale. The multilayered mixture usually occurs in scenes composed of materials with some height differences, such as forest, grassland, or rocks, containing many nonlinear interactions between the ground and the canopy. In general, the multilayered mixture consisting of more than two orders is ignored owing to its negligible interactions. For the second-order multilayered mixture model, the family of bilinear mixing models is usually adopted to solve the NLMM.
Due to the low spatial resolution of sensors, many pixels mixed by different pure materials exist in HS imagery, which inevitably conceals useful information and hinders the high-level image processing. SU aims to separate the observed spectrum into a suite of basic components, also called endmembers, and their corresponding fractional abundances. 

\subsection{Linear Mixing Model}
With the assumption of the single interaction between the incident light and the material, representative SU methods are based on the following linear mixing model (LMM) \cite{c189,RenL2021A}:
\begin{equation}
    \begin{aligned}
    \label{eq:LMM}
    \mathbf{X} = \mathbf{E} \mathbf{A} + \mathbf{N}
    \end{aligned}
\end{equation}
where $\mathbf{X} \in \mathbb{R}^{z \times hv}$, $\mathbf{E} \in \mathbb{R}^{z \times r}$, $\mathbf{A} \in \mathbb{R}^{r \times hv}$, and $\mathbf{N} \in \mathbb{R}^{z \times hv}$ denotes the observed unfolding HS matrix, the endmember matrix, abundance matrix, and additional noise, respectively. The LMM-based methods have drawn much attention due to their model simplicity and desirable performance \cite{c205,c206,c207}. However, current LMM-based matrix factorization methods usually convert the 3-D HS cube into a 2-D matrix, leading to the loss of spatial information in the relative positions of pixels. Tensor factorization-based approaches have been dedicated to SU to overcome the limitation of LMM.

\subsubsection{CP or Tucker Decomposition Model}

Zhang \textit{et al}. \cite{c190,c191} first introduced nonnegative tensor factorization (NTF) into SU via CP decomposition. However, this NTF-SU method hardly considers the relationship between LMM and NTF, giving rise to the lack of physical interpretation. 
Imbiriba \textit{et al}. \cite{c212} considered the underlying variability of spectral signatures and developed a flexible approach, named unmixing with LR tensor regularization algorithm accounting for EM variability (ULTRA-V). The ranks of the abundance tensor and the endmember tensor were estimated with only two easily adjusted parameters.
Sun \textit{et al}. \cite{c208} first introduced Tucker decomposition for blinding unmixing and increased the sparse characteristic of abundance tensor.

\subsubsection{BT Decomposition Model}

In terms of tensor notation, an HS data tensor can be represented by sum of the outer products of an endmember (vector) and its abundance fraction (matrix). This enables a matrix-vector third-order tensor factorization that consists of $R$ component tensors:
\begin{equation}
    \begin{aligned}
    \label{eq:SU-BTD}
    \mathcal{X} & =\sum_{r=1}^{R} \mathbf{A}_{r} \cdot \mathbf{B}_{r}^{T} \circ \mathbf{c}_{r} +\mathcal{N}\\
    &=\sum_{r=1}^{R} \mathbf{E}_{r} \circ \mathbf{c}_{r} +\mathcal{N}
    \end{aligned}
\end{equation}
where $\mathbf{E}_{r}$ calculated by the product of $\mathbf{A}_{r}$ and $\mathbf{B}_{r}^{T}$ denotes the abundance matrix, $\mathbf{c}_{r}$ is the endmember vector, and $\mathcal{N}$ represented the additional noise. Apparently, this matrix-vector tensor decomposition has the same form as BT decomposition, set up a straightforward link with the previously mentioned LMM model. Qian \textit{et al}. \cite{c192} proposed a matrix-vector NTF unmixing method, called MVNTF, by combining the characteristics of CPD and Tucker decomposition to extract the complete spectral-spatial structure of HS images. The MVNTF method for SU is formulated as 
\begin{equation}
    \begin{aligned}
    \label{eq:MVNTF}
    &\min_{\mathbf{E},\mathbf{c}} || \mathcal{X}- \sum_{r=1}^{R} \mathbf{E}_{r} \circ \mathbf{c}_{r} ||^2_F \\
    & {\rm s.t.} \mathbf{A}_{r} , \mathbf{B}_{r}^{T}, \mathbf{c}_{r} \geq 0
    \end{aligned}
\end{equation}

MVNTF derived BT decomposition essentially and established a physical connection with LMM. Compared with NMF-based unmixing approaches, MVNTF can achieve better unmixing performance in most cases. Nevertheless, the abundance results extracted by MVNTF may be over-smoothing and lose detailed information due to the strict LR constraint of NTF. Various spatial and spectral structures, such as spatial-spectral smoothness and non-local similarity, are proven to tackle the problem of pure MVNTF. 

Xiong \textit{et al}. \cite{c193} presented a TV regularized NTF (NTF-TV) method to make locally smooth regions share similar abundances between neighboring pixels and suppress the effect of noises. Zheng \textit{et al}. \cite{c194} offered a sparse and LR tensor factorization (SPLRTF) method to flexibly achieve the LR and sparsity characteristics of the abundance tensor. Feng \textit{et al}. \cite{c195} installed three additional constraints, namely sparseness, volume, and nonlinearity, into the MVNTF framework to improve the accuracies in impervious surface area fraction/classification map. Li \textit{et al}. \cite{c196} integrated NMF into MVNTF by making full use of their individual merits to characterize the intrinsic structure information. Besides, a sparsity-enhanced convolutional operation (SeCoDe) method \cite{c198} incorporated a 3-D convolutional operation into MVNTF for the blind SU task.

\begin{figure*}[htb]
	\begin{center}
	\includegraphics[height=12cm]{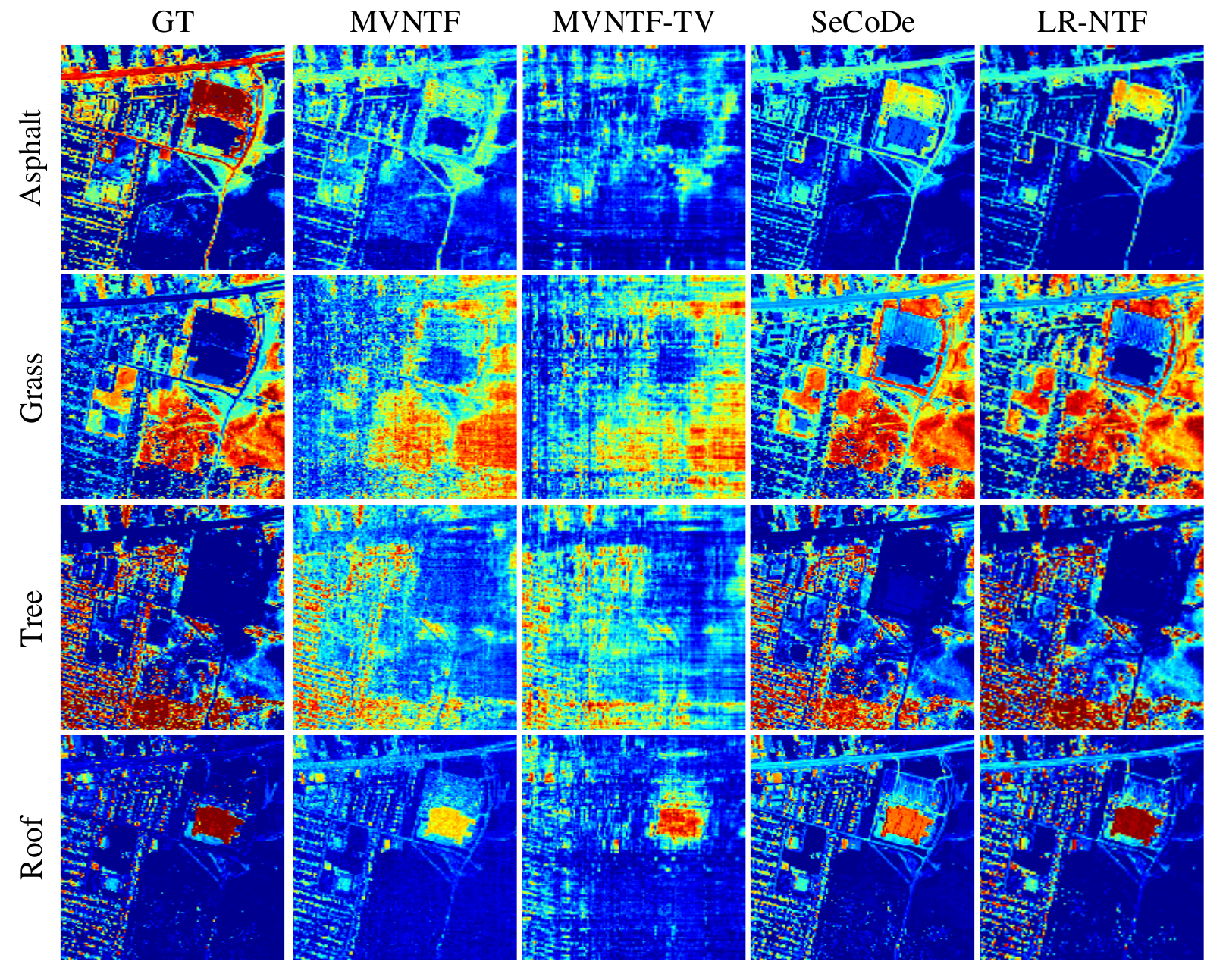}
	\end{center}
	\caption[unmixing1]{Abundance maps of different methods on the Urban data set. }
	\label{fig:Visio-unmixing1}
\end{figure*}

\begin{figure*}[htb]
	\begin{center}
	\includegraphics[height=8cm]{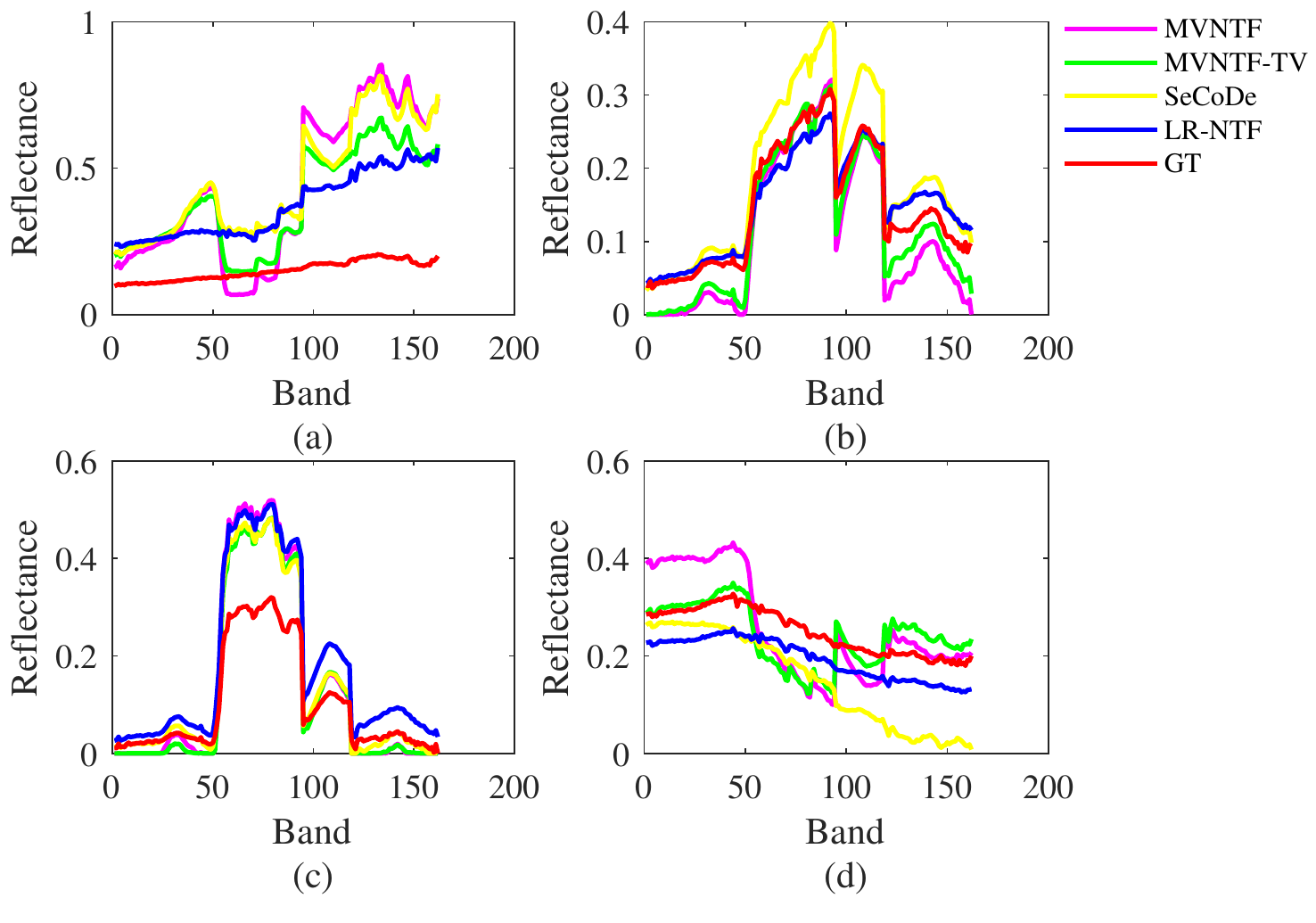}
	\end{center}
	\caption[unmixing2]{Endmember results of different methods on the Urban data set. (a) Asphalt, (b) Grass, (c) Tree, and (d) Roof. }
	\label{fig:Visio-unmixing2}
\end{figure*}

\begin{table*}[!htbp]
\centering
\caption{A quantitative comparison of different methods for HS SU.}
\begin{spacing}{1.0}
\scalebox{0.93}{
\begin{tabular}{c|c|cccc}
\hline
\multicolumn{2}{c|}{Method}&	MVNTF & MVNTF-TV & SeCoDe & LR-NTF \\
\hline
\multirow{4}{*}{SAD} & Asphalt & 0.3738	& 0.2606 & 0.2190 & $\mathbf{0.1127}$ \\
&Grass & 0.2572 &0.1722	& $\mathbf{0.0450}$ & 0.1349 \\
&Tree	&0.1474	&0.1450 & 0.0854 &$\mathbf{0.0632}$  \\
&Roof	&0.2825&	0.2273&	0.3861&	$\mathbf{0.0395}$\\
\cline{1-2}
\multicolumn{2}{c|}{MSAD} &	0.2652&	0.2013&	0.1839&
$\mathbf{0.0876}$ \\
\multicolumn{2}{c|}{RMSE}	&0.2638&	0.2588	&0.1453&$\mathbf{	0.1451}$ \\
\hline
\end{tabular}
}
\label{tab:tab-unmixing}
\end{spacing}
\end{table*}

\subsubsection{Mode-$3$ Tensor Representation Model}

Under the definition of the tensor mode-$n$ multiplication, LMM (\ref{eq:LMM}) is equivalent to
\begin{equation}
    \begin{aligned}
    \label{eq:mul-LMM}
    \mathcal{X} = \mathcal{A} \times_3 \mathbf{E} + \mathcal{N} 
    \end{aligned}
\end{equation}
where $\mathcal{A} \in \mathbb{R}^{h \times v \times R}$ denotes the abundance tensor containing $R$ endmembers. In \cite{c197}, the non-local LR tensor and 3-DTV regularization of the abundance tensor were introduced further extract the spatial contextual information of HS data. With abundance nonnegative constraint (ANC) and abundance sum-to-one constraint (ASC) \cite{c211}, the objective function of NLTR for SU is expressed as
\begin{equation}
    \begin{aligned}
    \label{eq:NLTR-SU}
&\min _{\mathcal{A}} \frac{1}{2}\left\|\mathcal{X}-\mathcal{A} \times_{3} \mathrm{E}\right\|_{F}^{2}+\lambda_{\mathrm{TV}}\|\mathcal{A}\|_{\mathrm{2DTV}}+\lambda_{\mathrm{NL}} \sum_{k=1}^{K}\left\|\mathcal{A}^{k}\right\|_{\mathrm{NL}}\\
&\text { s.t. } \mathcal{A} \geq \mathbf{0}, \quad \mathcal{A} \times_{1} \mathbf{1}_{P}=\mathbf{1}_{h \times v} 
    \end{aligned}
\end{equation}
where $\mathbf{1}_{P}$ is a $P$-dimensional vector of all 1, $\mathbf{1}_{h \times v} $ denotes a matrix of element 1, and the non-local LR regularization is defined as
\begin{equation}
    \begin{aligned}
\|\mathcal{A}^{k}\|_{\mathrm{NL}}=\sum_{i=1}^{p} \operatorname{LS}(\mathbf{A}^{(i)})
\end{aligned}                                                                               
\end{equation}    

ANC, ASC, and the sparseness of abundance are often introduced into sparse unmixing models \cite{Iordache2011Sparse,Iordache2014Collaborative}, which produces the endmembers and corresponding abundance coefficients by a known spectral library instead of extracting endmembers from HS data \cite{RenL2022A,RenL2020A}. Sun \textit{et al}. \cite{c210} developed a weighted non-local LR tensor decomposition method for HS sparse unmixing (WNLTDSU) by adding collaborative sparsity and 2DTV of the endmember tensor into a weighted non-local LR tensor framework. The LR constraint and joint sparsity in the non-local abundance tensor were imposed in a non-local tensor-based sparse unmixing (NL-TSUn) algorithm \cite{c209}.

\subsection{NonLinear Mixing Model}
To this end, numerous NLMMs have been studied in SU by modeling different order scatterings effects and producing more accurate unmixing results. To this end, numerous NLMMs have been proposed in SU by modeling different order scatterings effects and producing accurate unmixing results \cite{c199,c200,c201}. Traditional NLMMs, such as Bilinear mixture models (BMMs), usually transform an HS cube into a 2-D matrix and have the same fault as LMMs \cite{c202,c203,yao2019nonconvex}.

To effectively address the nonlinear unmixing problem, Gao et al. \cite{c204} expressed an HS cube $\mathcal{X} \in \mathbb{R}^{h \times v \times z}$ based on tensor notation in the following format
\begin{equation}
    \begin{aligned}
    \label{eq:NLMM1}
    \mathcal{X}=\mathcal{A} \times_{3} \mathbf{C}+\mathcal{B} \times_{3} \mathbf{E}+\mathcal{N}
    \end{aligned}
\end{equation}
where $\mathbf{C} \in \mathbb{R}^{z \times R}$, $\mathcal{B} \in \mathbb{R}^{h \times v \times R(R-1)/2}$, and $\mathbf{E} \in \mathbb{R}^{z \times R(R-1)/2}$ represent the mixing matrix, the nonlinear interaction abundance tensor, and the bilinear interaction endmember matrix respectively. A nonlinear unmixing method \cite{c204} was first based on NTF by taking advantage of the LR property of the abundance maps and nonlinear interaction maps, which validated the potential of tensor decomposition in nonlinear unmixing.

\subsection{Experimental results and analysis}
\label{sect:experiment_SU}

The Urban HS data set obtained by the HYDICE sensor over the urban area, Texas, USA, are selected for evaluating the performance of different unmixing methods qualitatively, including MVNTF \cite{c192}, MVNTF-TV \cite{c193}, SeCoDe \cite{c198}, and LR-NTF \cite{c204}.  For a fair comparison, HSsignal subspace identification by minimum error (HySime) \cite{c221} and vertex component analysis (VCA) \cite{c222} algorithms are adopted to determine the number of endmembers and the endmember initialization. The urban data contains $307 \times 307$ pixels and 210 bands ranging from 0.4 to 2.5 $\mu$m. Due to the water vapor and atmospheric effects, 162 bands are remained after removing the affected channels. Four main materials in this scene are investigated, that is, $\#1$ Asphalt, $\#2$ Grass, $\#3$ Tree, and $\#4$ Roof. Two quantitative metrics are utilized to evaluate the extracted abundance and endmember results, namely RMSE and SAD. 

For illustrative purposes, Fig. \ref{fig:Visio-unmixing1} and Fig. \ref{fig:Visio-unmixing2} display the extracted abundances and the corresponding endmember results of different tensor decomposition-based SU approaches. The quantitative results on the urban data are reported in Tab. \ref{tab:tab-unmixing}, where the best results are marked in bold. MVNTF yields poor unmixing performance for both endmember extraction and abundance estimation compared with other tensor-based unmixing methods since it only considers the tensor structure to represent the spectral-spatial information of HS images and ignores other useful prior regularizations. Compared with MVNTF, MVNTF-TV integrates the advantage of TV and tensor decomposition, bringing certain performance improvements in terms of SAD, MSAD, and RMSE. SeCoDe addresses the problem of spectral variabilities in a convolutional decomposition fashion effectively, thereby yielding further performance improvement of endmember and abundance results. Different from SeCoDe, LR-NTF considers the nonlinear unmixing model of tensor decomposition and the low-rankness regularization of abundances. The unmixing results of LR-NTF are superior to those of other competitive approaches on the urban data, demonstrating its superiority and effectiveness.


\subsection{Future challenges}
\label{sect:challenges_SU} 

Several advanced tensor decomposition-based methods have recently achieved effectiveness in HS SU. Nonetheless, there is still a long way to go towards the definition of statistical models and the design of algorithms. In the following, we briefly summarize some aspects that deserve further consideration:

The most commonly utilized evaluation indices for HS SU include RMSE (that measures the error between the estimated abundance map and the reference abundance map) and SAD (which assesses the similarity of the extracted endmember signatures and the true endmember signatures). However, RMSE and SAD just contribute to a quantitative comparison of SU results when the ground truth for abundances and endmembers exists. If there are no references in the real scenario, meaningful and suitable evaluation metrics should be developed in future work.

Traditional NLMMs are readily interpreted as matrix factorization problems. The tensor decomposition-based NLMM has been springing up in the recent few years.  We should consider complex interactions like the intimate and multilayered mixture for establishing general and robust tensor models.

Another important challenge is the high time consumption required by high-performance SU architectures, which hinders their applicability in real scenarios. Especially, as the number of end members and the size of the image increase, the current NTF-based unmixing methods are difficult to deal with this situation owing to a large amount of computational consumption. Therefore, the exploration of more computationally efficient tensor-based approaches will be an urgent research direction in the future.

\section{Conclusion}
\label{sect:conclusion}

HS technique accomplishes the acquisition, utilization, and analysis of nearly continuous spectral bands and permeates through a broad range of practical applications, having attached incremental attention from researchers worldwide. In HS data processing, large-scale and high-order properties are often involved in collected data. The ever-growing volume of 3-D HS data puts higher demands on the processing algorithms to replace the 2-D matrix-based methods. Tensor decomposition plays a crucial role in both problem modelings and methodological approaches, making it realizable to leverage the spectral information of each complete 1-D spectral signature and the spatial structure of each complete 2-D spatial image. In this article, we presented a comprehensive and technical review of five representative HS topics, including HS restoration, CS, AD, HS-MS fusion, and SU. Among these tasks, we reviewed current tensor decomposition-based methods with main formulations, experimental illustrations, and remaining challenges. The most important and compatible challenges related to consolidating tensor decomposition techniques for HS data processing should be emphasized and summarized in five aspects: model applicability, parameter adjustment, computational efficiency, methodological feasibility, and multi-mission applications.

$\textbf{Model applicability}$: Tensor decomposition theory and practice offer us versatile and potent weapons to solve various HS image processing problems. A high-dimensional tensor is often decomposed by different categories of tensor decomposition into several decomposition factors/cores. One sign reveals that the mathematical meaning of different factors/cores should be made connection with the physical properties of HS structure. Another sign is that each HS task contains multiple modeling problems, such as various types of HS noise (i.e., Gaussian noise, stripes, or mixed noise) caused by different kinds of sensors or external conditions. The tensor decomposition-based models should be capable of characterizing the specific HS properties and being used in different scenarios.

$\textbf{Parameter adjustment}$: In the algorithmic solution, parameter adjustment is an indispensable portion to achieve the significant performances of HS data processing. Parameters can be gradually tuned via extensive simulated experiments, while sometimes, they should be reset for various data sets due to the uncertainty of data size. In practice, users are most likely to be non-professional with little knowledge of a special algorithm, leading
to improper parameter setting and unsatisfactory processing results. Therefore, in the future, efforts should be made to design a fast proper-parameter search scheme or reduce the number of parameters to increase algorithmic practicability.

$\textbf{Computational consumption}$: Tensor decomposition-based methods have achieved satisfactory results in HS data processing, yet they sometimes cause high computational consumption. For instance, a non-local LR tensor denoising model, TDL spends more than 10 min under a data set of $200 \times 200 \times 80$. As the image size increases, the increasing number of non-local FBPs will cause a larger amount of time consumption. Thus, there still exists a vast room for promotion and innovation of improving the optimization efficiency of HS data processing.

$\textbf{Methodological feasibility}$: Unlike deep learning-based methods, designing handcrafted priors is the key to tensor decomposition-based methods. Existing methods exploit the structure information of the underlying target image by implementing various handcrafted priors, such as LR, TV, and non-local similarity. However, different priors assumptions apply to specific scenarios, making it challenging to choose suitable priors according to the characteristics of HS images to be processed. Deep learning-based methods automatically learn the prior information implicitly from data sets themselves without the trouble of manually designing a manual regularizer. As an advisable approach, deep learning can be incorporated into tensor-based methods to mine essential multi-features and enhance the methodological feasibility.

$\textbf{Multi-mission applications}$: The extremely broad field of HS imagery makes it impossible to provide an exhaustive survey on all of the promising HS RS applications. It is certainly of significant interest to develop tensor decomposition-based models for other noteworthy processing and analysis chains in future work, including classification, change detection, large-scale land cover mapping, and image quality assessment. Some HS tasks serve as the pre-processing step for high-level vision. For example, the accuracy of HS classification can be improved after an HS denoising step. How to apply tensor decomposition for high-level vision and even multi-mission frameworks may be a key challenge.

\bibliographystyle{IEEEtran}
\bibliography{refs}

\end{document}